\documentclass[aos]{imsart}
\RequirePackage{amsthm,amsmath,amsfonts,amssymb}
\makeatletter\def\thebibliography@size{\footnotesize}\makeatother
\makeatletter
\AtBeginDocument{%
  \setlength{\textfloatsep}{9pt plus 2pt minus 2pt}%
  \setlength{\floatsep}{8pt plus 2pt minus 2pt}%
  \setlength{\intextsep}{8pt plus 2pt minus 2pt}%
  \setlength{\abovecaptionskip}{4pt}%
  \setlength{\belowcaptionskip}{2pt}%
  \setlength{\abovedisplayskip}{5pt plus 1pt minus 2pt}%
  \setlength{\belowdisplayskip}{5pt plus 1pt minus 2pt}%
  \setlength{\abovedisplayshortskip}{3pt plus 1pt}%
  \setlength{\belowdisplayshortskip}{3pt plus 1pt}%
}
\makeatother
\RequirePackage[numbers,sort&compress]{natbib}
\RequirePackage[hidelinks]{hyperref}
\RequirePackage{graphicx}
\RequirePackage{xcolor}

\RequirePackage{bm}
\RequirePackage{booktabs}
\RequirePackage{multirow}
\RequirePackage{subcaption}

\startlocaldefs
\theoremstyle{plain}
\newtheorem{theorem}{Theorem}[section]

\newtheorem{proposition}{Proposition}[section]
\theoremstyle{definition}
\newtheorem{assumption}{Assumption}[section]
\newtheorem{definition}{Definition}[section]
\theoremstyle{definition}
\newtheorem{remark}{Remark}[section]

\DeclareMathOperator{\Vol}{Vol}
\DeclareMathOperator{\TV}{D_{TV}}
\DeclareMathOperator{\diam}{diam}

\DeclareMathOperator{\diag}{diag}
\DeclareMathOperator{\softmax}{softmax}

\endlocaldefs

\begin{document}

\begin{frontmatter}
\title{Scalable Minimum-Volume
Simplex Estimation with
Non-asymptotic Analysis}
\runtitle{Scalable Simplex Estimation}

\begin{aug}
\author[A]{\fnms{Jun}~\snm{Li}\ead[label=e1]{lijuncug@cug.edu.cn}}
\author[A]{\fnms{Yanlong}~\snm{Guo}\ead[label=e2]{guoyanlong@cug.edu.cn}}
\author[A]{\fnms{Zhaozhao}~\snm{Zeng}\ead[label=e3]{zengzhaozhao@cug.edu.cn}}

\address[A]{{School of Computer Science, China University of Geosciences; corresponding author: Jun Li}\printead[presep={,\ }]{e1}\printead[presep={,\\ }]{e2}\printead[presep={,\ }]{e3}}
\end{aug}

\begin{abstract}

We study the estimation of a $K$-dimensional simplex from $N$
i.i.d.\ points sampled uniformly from its interior;
the observations are convex combinations of $K+1$ unknown
prototypes. Existing
polynomial-time estimators need cubic per-sample work or
$O(NK)$ storage and are impractical at $N\sim 10^6$--$10^8$. We propose DeepMVSA, which re-expresses
the minimum-volume principle in neural implicit form: a
lightweight coordinate network generates the mixing weights and
a triangular LU-type parameterization the
dual simplex matrix, reducing the trainable-state memory to $O(K^2)$, independent
of $N$, and the cost per data pass to $O(NK^2)$. We prove a
non-asymptotic sample-complexity bound of the polynomial-time
benchmark order for a localized surrogate estimator; an oracle inequality for every global
minimizer of the neural objective, with volume-inflation
control and an explicit shrinkage bias; a
conditional end-to-end error budget separating statistical,
approximation, optimization, and enclosure-residual terms on
an explicit envelope event; and two-point lower
bounds: at any noise level $\sigma>0$ fixed independently of $N$,
the $N^{-1/2}$ scaling is unimprovable in its $N$-exponent. Experiments
with up to $N=10^8$ synthetic observations are
consistent with the predicted accuracy and scaling, and
feasibility on real scenes of $\sim 10^7$ pixels is
demonstrated.
\end{abstract}



\end{frontmatter}

\section{Introduction}
\label{sec:intro}

Let $\mathbb{S}_K$ denote the set of all $K$-dimensional simplices
in $\mathbb{R}^K$. Consider $N$ i.i.d.\ random points
$X_1,\dots,X_N$ drawn uniformly from a fixed but unknown simplex
$\mathcal{S}_T\in\mathbb{S}_K$. The problem of
\emph{statistical learning of simplices}---called \emph{unmixing}
in the literature---asks for an estimator
$\widehat{\mathcal{S}}$ whose uniform distribution is within
total-variation distance $\epsilon$ of that on $\mathcal{S}_T$,
with probability at least $1-\zeta$ \cite{najafi2021aos}. It
underlies methodologies in hyperspectral remote sensing,
computational biology, and non-negative matrix factorization,
where each observation is modeled as an unknown convex mixture
of $K+1$ latent sources \cite{fu2019,shenorr2010}.

The maximum-likelihood estimator for $\mathcal{S}_T$ is the
minimum-volume simplex containing all observations
\cite{najafi2021aos}; it attains the fastest sample complexity known for this model
$N\ge O(\epsilon^{-1}[K^2\log(K/\epsilon)+\log(1/\zeta)])$ but is
NP-hard to compute \cite{packer2002}. A polynomial-time algorithm with a polynomial sample-size guarantee was first given by Anderson, Goyal and Rademacher \cite{anderson2013}, answering a question of Frieze, Jerrum and Kannan \cite{frieze1996} via a reduction to independent component analysis, though without an explicit statistical rate.
The first polynomial-time estimator with an explicit rate of convergence is due to Najafi et al.~\cite{najafi2021aos}, whose
\emph{Continuously Relaxed Risk} (Soft-ML) estimator
\begin{equation}
  \widehat{\mathcal{S}}_{\text{SOFT}}
  = \arg\min_{\mathcal{S}\in\mathbb{S}_K}
  \frac{1}{\sqrt{N}}\sum_{i=1}^N \ell(d_{\mathcal{S}}(X_i))
  + \gamma\Vol(\mathcal{S}),
  \label{eq:softml}
\end{equation}
with $d_{\mathcal{S}}(\cdot)$ the planar distance to the simplex,
$\gamma>0$ a volume weight, and $\ell(u)=1-e^{-bu}$ a smooth
increasing loss, satisfies
$\TV(\mathbb{P}_{\mathcal{S}_T},
\mathbb{P}_{\widehat{\mathcal{S}}}) \le \epsilon$ whenever
\begin{equation}
  N \;\ge\; O\!\left(\frac{1}{\epsilon^2}\left[
  K^2\log\frac{K}{\epsilon} + \log\frac{1}{\zeta}\right]\right),
  \label{eq:soft-rate}
\end{equation}
under an isoperimetricity assumption on $\mathcal{S}_T$.
Complementary polynomial-time algorithms exist under
separability-type structural assumptions orthogonal to the
uniform-sampling model studied here
\cite{arora2012,recht2012,gillis2014,arora2013topic,anandkumar2014tensor}.

Three questions, however, remain open. First,
existing guarantees apply only to estimators that optimize
explicitly over simplex vertices subject to per-sample enclosure
or distance evaluations; nothing is known, non-asymptotically,
about \emph{implicit} estimators for simplex estimation and unmixing---such as neural
coordinate-network estimators, whose hypothesis class has capacity
independent of $N$, even though non-asymptotic guarantees for implicit (neural) estimators exist in other learning problems \cite{sreekumar2021,tsur2023}. Second, existing analyses control an idealized
surrogate minimizer; no end-to-end guarantee addresses the
estimator that is actually \emph{deployed}, namely a trained
network carrying approximation and optimization error on top of
the statistical error. Third, the rate \eqref{eq:soft-rate} is
only an upper bound: the noiseless minimax-optimal sample
complexity at fixed $K$ is of order $K/\epsilon$
\cite{saberi2025}, and minimax rates for polytope and set
estimation exist under different models and error metrics
\cite{brunel2013,baldin2016}, but no lower bound certifies the
optimality of \eqref{eq:soft-rate} among polynomial-time estimators, and the sharpness of the localization constants
driving it has never been tested.

In this paper we propose Deep Minimum Volume Simplex Analysis (DeepMVSA), a neural implicit simplex estimator
whose memory footprint and parameter complexity are independent of
$N$, and we answer all three questions for it; whether any polynomial-time estimator can attain the noiseless minimax rate $\Theta(K/N)$ \cite{saberi2025} remains open. DeepMVSA retains
the minimum-volume principle in the dual domain---the simplex is
recovered through a dual matrix $Q$ without ever estimating the
$N$ mixing weights---but replaces the explicit abundance matrix
and the hard enclosure constraint $QX\ge0$ by an implicit
coordinate network and a differentiable quadratic enclosure
penalty, respectively. Our contributions are
threefold:
\begin{enumerate}
  \item \textbf{A non-asymptotic statistical
  theory for the deployed estimator, conditional on an explicit envelope event.} The
  surrogate estimator associated with DeepMVSA attains the
  benchmark rate \eqref{eq:soft-rate} with
  explicit localization constants (Theorem~\ref{thm:main}),
  for $N$ beyond an explicitly quantified fixed-$K$ threshold, and four refinements go
  beyond it: an oracle inequality transferring the guarantee to
  every global minimizer of the practical objective, with an
  explicit volume-inflation factor and a shrinkage-bias scaling
  guideline for the enclosure weight
  (Proposition~\ref{prop:transfer}); a conditional end-to-end error budget
  separating statistical, approximation, optimization, and
  enclosure-residual terms on an explicit envelope event
  (Theorem~\ref{prop:endtoend}); a quantitative approximation
  rate whose capacity prescription is independent of $N$
  (Proposition~\ref{prop:apxrate}); and two-point minimax lower
  bounds together with a homothetic construction
  certifying the necessity of the $\gamma_{\mathrm{vol}}$-dependent
  gap cap (Proposition~\ref{prop:lowerbound}). The
  benchmark rate moreover persists under bounded observation noise
  of magnitude $O(\epsilon^2/K)$
  (Proposition~\ref{prop:noise}). The logical status of these
  statements is labeled as such throughout:
  Theorem~\ref{thm:main} and
  Propositions~\ref{prop:noise} and~\ref{prop:lowerbound} are
  unconditional;
  Proposition~\ref{prop:transfer} is an unconditional oracle
  inequality for global minimizers of the practical objective;
  Theorem~\ref{prop:endtoend} is conditional on an envelope
  event and on the residual $\eta_{\mathrm{opt}}$,
  neither certified for the Adam output, whose
  statistical behavior is empirical.

  \item \textbf{An estimator with $N$-independent cost.} Each
  sample obtains its abundance vector implicitly through a
  lightweight coordinate network with $O(K^2)$ parameters, so the
  trainable-model and optimizer-state memory is $O(K^2)$,
  independent of $N$, and the cost per full data pass is reduced
  from the $O(NK^3)$ of the Soft-ML implementation used here to $O(NK^2)$
  (Theorem~\ref{thm:complexity}). A triangular (LU-type)
  parameterization of the dual matrix guarantees strictly positive
  simplex volume for all finite parameter values
  (Section~\ref{subsec:flow}, Remark~\ref{rem:flow}).
  Throughout, ``polynomial time'' refers to the estimator class
  of the benchmark, not to DeepMVSA's nonconvex training (see the
  discussion following Theorem~\ref{thm:complexity}).

  \item \textbf{Evidence at scale.} On synthetic simplices,
  cell-type deconvolution, and real hyperspectral scenes,
  DeepMVSA matches or improves the accuracy of
  minimum-volume baselines while scaling to $N\sim10^{8}$
  observations, two to three orders of magnitude
  beyond the largest instances reported for neural
  re-expressions of \eqref{eq:mv} ($\sim10^{5}$ samples)
  (Section~\ref{sec:exp}).
\end{enumerate}

\paragraph{Related work}
The minimum-volume (MV) family formulates simplex estimation 
after subspace projection, centering, and whitening
\cite{li2015tgrs,najafi2021aos}, as the constrained reconstruction
problem
\begin{equation}
  \min_{\widetilde{M},A}\;
  \frac{1}{N}\sum_{j=1}^N \|X_j - \widetilde{M}a_j\|_2^2
  -\,\gamma\log|\det Q|
  \quad\text{s.t.}\quad
  \mathbf{1}^\top a_j = 1,\; a_j \ge 0,
  \label{eq:mv}
\end{equation}
where $\widetilde{M}\in\mathbb{R}^{K\times(K+1)}$ is the vertex
matrix, $A\in\mathbb{R}^{(K+1)\times N}$ the abundance matrix, and
$Q=\widetilde{M}_{\mathrm{cen}}^{-1}$ the dual matrix of the
centered vertex matrix, so that
$\mathrm{Vol}(\widetilde M)\propto 1/|\det Q|$. MVSA
\cite{li2015tgrs} and its robust variant \cite{zhang2017rmvsa},
from the first author's prior work, eliminate $A$ and solve
directly for $Q$; other geometrical methods include N-FINDR
\cite{winter1999}, VCA \cite{nascimento2005}, SISAL
\cite{bioucas2009sisal}, the maximum-volume-inscribed-ellipsoid
framework \cite{lin2018mvie}, and volume maximization \cite{chan2011svmax};
Bayesian \cite{dobigeon2009} methods have also been
developed. Statistical guarantees for related simplex-structured
problems include optimal-rate estimation of topic models
\cite{bing2020,ke2024topic} and probabilistic simplex component analysis
\cite{wu2022prsca}, under models different from the uniform-sampling
model studied here. Neural re-expressions of \eqref{eq:mv}---autoencoder-based
\cite{gao2022cycunet,hong2022egunet} and attention-based
\cite{ghosh2022deeptrans} designs---are architectural
rather than statistical: none carries a non-asymptotic
sample-complexity guarantee on the estimated simplex, and their
memory or per-pass cost grows with $N$ (the largest images
treated there contain $\sim10^{5}$ samples), so we restrict
numerical comparisons to estimators sharing the minimum-volume
objective \eqref{eq:mv}. A critical comparison of
autoencoder-based unmixers appears in \cite{palsson2022}; such a
comparison would in any case conflate architecture, initialization,
and tuning, and no rate statement about the present model could be
inferred from it.

\paragraph{Organization}
Section~\ref{sec:prelim} fixes notation and assumptions;
Section~\ref{sec:method} develops DeepMVSA and its guarantees;
Section~\ref{sec:exp} reports experiments; Section~\ref{sec:conclusion}
concludes. All proofs, a sensitivity study, and reproducibility
details are in the supplement \cite{dmvsa2026supp}, whose items
carry an S prefix and are cited as ``Lemma~S1 of
\cite{dmvsa2026supp}''.

\section{Preliminaries}
\label{sec:prelim}

For $\mathcal{S}(\widetilde{M})\in\mathbb{S}_K$, the simplex with vertex matrix $\widetilde{M}$ (whose $K+1$ columns are its vertices), $\Vol(\mathcal{S})$ denotes its Lebesgue measure. Let
$\mathcal{H}_k = \{\boldsymbol{x}\in\mathbb{R}^K : \boldsymbol{w}_k^\top \boldsymbol{x} + b_k = 0\}$ be
the hyperplane containing the $k$-th facet, with outward unit
normal $\boldsymbol{w}_k\in\mathbb{R}^K$ and bias $b_k\in\mathbb{R}$.
Three representations of a simplex recur and are kept distinct
throughout: the geometric object $\mathcal{S}\in\mathbb{S}_K$;
its vertex matrix $\widetilde M\in\mathbb{R}^{K\times(K+1)}$
($M$ in whitened coordinates); and the learned dual
representation $(\theta_0,Q)$ of Section~\ref{sec:method}.
The target is always $\mathcal{S}_T$; hatted symbols such as
$\widehat{\mathcal{S}}$ denote estimators.

\begin{definition}[Planar distance, \cite{najafi2021aos}]
  \label{def:planar}
  For any $\boldsymbol{x}\in\mathbb{R}^K$, the planar distance from
  $\boldsymbol{x}$ to $\mathcal{S}$ is
  \[
    d_{\mathcal{S}}(\boldsymbol{x}) = \max\Bigl\{0,\;
    \max_{k=0,\dots,K} \boldsymbol{w}_k^\top \boldsymbol{x} + b_k\Bigr\}.
  \]
\end{definition}

Let $\mathbb{P}_{\mathcal{S}}$ be the uniform distribution on
$\mathcal{S}$ with density $\rho_{\mathcal{S}}(\boldsymbol{x})=
\mathbf{1}_{\mathcal{S}}(\boldsymbol{x})/\Vol(\mathcal{S})$, where
$\mathbf{1}_{\mathcal{S}}(\cdot)$ is the indicator function of
$\mathcal{S}$. The total-variation distance between
$\mathbb{P}_{\mathcal{S}}$ and
$\mathbb{P}_{\mathcal{S}'}$, the uniform distribution on
another simplex $\mathcal{S}'\in\mathbb{S}_K$, is denoted
$\TV(\mathbb{P}_{\mathcal{S}}, \mathbb{P}_{\mathcal{S}'})$. We assume the following data-generating model and regularity condition:

Throughout, \emph{observation} and \emph{sample} refer to a
statistical draw from the sampling model, whereas \emph{pixel}
refers to an observation indexed by a position on an image
grid.

\begin{assumption}[Uniform simplex sampling]
  \label{ass:uniform}
  $X_1,\dots,X_N \stackrel{\text{i.i.d.}}{\sim}
  \mathbb{P}_{\mathcal{S}_T}$ for some fixed
  $\mathcal{S}_T\in\mathbb{S}_K$.
\end{assumption}

\begin{assumption}[Isoperimetricity, \cite{najafi2021aos}]
  \label{ass:iso}
  $\mathcal{S}_T$ is $(\underline{\lambda},\bar{\lambda})$-isoperimetric
for some positive constants
$\underline{\lambda},\bar{\lambda}>0$; that is,
  \[
    \max_{k,k'}\|\theta_k-\theta_{k'}\|_2 \leq
    \underline{\lambda}K\Vol(\mathcal{S}_T)^{1/K},
    \qquad
    \max_k \Vol(\mathcal{S}_{T,-k}) \leq
    \bar{\lambda}\Vol(\mathcal{S}_T)^{\frac{K-1}{K}},
  \]
  where $\mathcal{S}_{T,-k}$ denotes the $k$-th facet of
  $\mathcal{S}_T$ (the $(K-1)$-simplex obtained by removing the
  $k$-th vertex); here and throughout, the vertices
  $\theta_k$ and facets of the simplex are indexed by
  $k\in\{0,\dots,K\}$.
\end{assumption}

Under Assumptions~\ref{ass:uniform} and
\ref{ass:iso}, the uniform distribution
$\mathbb{P}_{\mathcal{S}_T}$ identifies $\mathcal{S}_T$ (it
is the minimum-volume simplex of probability one
\cite{najafi2021aos}); the strength of the identification
depends on the simplex geometry through
$(\underline\lambda,\bar\lambda)$, which quantify the separation
of the target simplex from degenerate competitors, and the diameter $D_T$;
the facet-angle floor of Lemma~S3 of \cite{dmvsa2026supp} is
controlled by the volume $V_T$ and the diameter $D_T$ alone.

Our goal is to construct an estimator
whose trainable-state memory is independent of $N$ and whose
cost per full data pass is linear in $N$, and to equip it with
a non-asymptotic theory that the existing literature does not
provide for any estimator in this model: guarantees for the
implicit estimator class itself, a conditional
end-to-end error budget for the trained network, and lower
bounds certifying the
sharpness of the rates and localization caps. The
guarantees below are stated against the statistical benchmark
\eqref{eq:soft-rate}, while the estimator, its risk, and the
phenomena that govern it are distinct.

\section{DeepMVSA: Neural Implicit Estimator}
\label{sec:method}

\subsection{Triangular parameterization of the dual matrix}
\label{subsec:flow}

We re-parameterize the problem in the dual space. Recall from \eqref{eq:mv} that $Q=\widetilde{M}_{\mathrm{cen}}^{-1}$ is the dual matrix of the centered vertex matrix, with $\mathrm{Vol}\propto1/|\det Q|$. Rather than optimizing $Q$ as an unstructured matrix, we set
\begin{equation}
  \label{eq:q-flow}
  Q(\psi) = L(\psi_L)\, Q_0\, R(\psi_R, \psi_r),
\end{equation}
where $\psi=(\psi_L,\psi_R,\psi_r)$ collects all learnable
parameters, $Q_0$ is a deterministic initialization (e.g., from
VCA \cite{nascimento2005}), and
\begin{align}
  L(\psi_L) &{}= I_K + \Lambda(\psi_L),\\
  R(\psi_R, \psi_r) &{}= \diag(e^{\psi_r}) + \Upsilon(\psi_R),
\end{align}
where $\Lambda(\psi_L)$ and $\Upsilon(\psi_R)$ denote,
respectively, the strict lower- and strict upper-triangular matrices
whose off-diagonal nonzero entries are given by $\psi_L$ and
$\psi_R$, and $I_K$ is the identity matrix.
We refer to the learnable map \eqref{eq:q-flow} as
the \emph{triangular dual parameterization} of $Q$ (the
\emph{$Q$-flow} for short). The three factors contribute $K(K-1)/2$,
$K(K-1)/2$, and $K$ entries, so the $Q$-flow comprises exactly
$K^2$ learnable parameters.
Both factors are triangular---unit diagonal ($L$) and diagonal
entries $e^{\psi_{r,k}}$ ($R$)---so $\det Q = \det Q_0 \cdot \prod_{k=1}^K e^{\psi_{r,k}} \neq 0$ for all finite $\psi$, $\psi_{r,k}$ the $k$-th component of $\psi_r$. Consequently, the volume regularizer simplifies to
$-\log|\det Q| = -\log|\det Q_0| - \sum_k \psi_{r,k}$, linear in
$\psi_r$ and globally well-defined, and the zero-measure degeneracy
set $\{\mathcal{S}\in\mathbb{S}_K : \Vol(\mathcal{S})=0\}$ is
avoided by construction, removing the need for the infinitesimal
noise injection of \cite[Sec.~3]{najafi2021aos}.

\begin{remark}[Triangular dual parameterization; relation to invertible flows]
\label{rem:flow}
The factorization \eqref{eq:q-flow} is a re-parameterization of
$Q$ by triangular factors with a tractable log-determinant; we
refer to it as the \emph{triangular dual parameterization} (or
\emph{triangular chart}) of $Q$. It is a local chart of the
general linear group around $Q_0$: Gaussian elimination without
pivoting succeeds exactly when all leading principal minors of
$Q_0^{-1}Q$ are nonzero, so every matrix in a Zariski-open dense
subset of $\mathrm{GL}(K)$ admits such an
$L\!\cdot\!\mathrm{diag}\cdot R$ factorization (a Bruhat-type
cell). The construction is inspired by the affine coupling and
$1\times1$ convolution layers of normalizing flows
\cite{dinh2017,kingma2018}, but it is not a generative flow and
no density estimation is involved; the flow terminology is
retained only as motivation.
\end{remark}

\begin{remark}[Learnable anchor and the vertex map]
\label{rem:anchor}
The dual matrix $Q\in\mathbb{R}^{K\times K}$ determines the
facets of the estimated simplex only up to a translation. We
therefore augment $\psi$ with a learnable \emph{anchor}
$\theta_0\in\mathbb{R}^K$ and define
\begin{equation}
  \label{eq:vertex-map}
  \mathcal{S}(Q^{-1};\theta_0)
  \;:=\;
  \mathrm{conv}\bigl\{\theta_0,\;\theta_0+Q^{-1}e_1,\;\dots,\;
  \theta_0+Q^{-1}e_K\bigr\},
\end{equation}
the simplex whose $K+1$ vertices are the anchor and the anchor
shifted by the columns of $Q^{-1}$. Its volume is
$\Vol(\mathcal{S}(Q^{-1};\theta_0)) = 1/(K!\,|\det Q|)$, and the
barycentric coordinates of a point $x\in\mathbb{R}^K$ relative to
\eqref{eq:vertex-map} are
\begin{equation}
  \label{eq:bary}
  a_k(x) = \bigl(Q(x-\theta_0)\bigr)_k,\quad k=1,\dots,K,
  \qquad
  a_0(x) = 1 - \mathbf{1}^\top Q(x-\theta_0),
\end{equation}
so that $\sum_{k=0}^K a_k(x)=1$ identically and
$x\in\mathcal{S}(Q^{-1};\theta_0)$ if and only if $a_k(x)\ge0$
for all $k=0,\dots,K$. The anchor contributes $K$ further
parameters, bringing the total to $K^2+K=O(K^2)$, independent of
$N$. The representation \eqref{eq:vertex-map} singles out one
vertex as the anchor and is not symmetric in the vertices,
although the simplex it describes is: any vertex can serve as
the anchor, with $Q$ re-expressing the rest as shifts from it.
The asymmetry belongs to the parameterization, not the object;
the permutation-invariance discussion of Section~S1.9 of the
supplement \cite{dmvsa2026supp} applies verbatim.
\end{remark}

\subsection{Neural implicit abundance estimation}
\label{subsec:anet}

A key distinction between DeepMVSA and prior simplex learners is
that we do \textit{not} treat the abundance matrix $A$ as an explicit
optimization variable. Instead, we adopt a \textit{neural implicit}
parameterization: the abundance vector $a_j$ of each observation
$j$ is generated by a lightweight network (A-Net) with parameters
$\xi$. Concretely, we construct a differentiable mapping
\begin{equation}
  a_j = f_\xi(c_j),
  \label{eq:implicit-map}
\end{equation}
where $c_j\in\mathbb{R}^{d}$ is the normalized spatial coordinate of observation $j$ ($d=2$ for imagery on a rectangular grid and $d=1$ for vectorized data). This is an
instance of the coordinate-based \emph{implicit neural
representation} paradigm
\cite{mildenhall2020,sitzmann2020,tancik2020}, in which
a whole signal is stored in network weights, transplanted here
from signal reconstruction to abundance estimation.
The analogy is architectural only: the abundances of the
index-coordinate experiments of Sections~\ref{subsec:exp-a}--\ref{subsec:exp-bio}
are independent of the coordinate, so the network does not
compress them; its role there is to provide a differentiable
reconstruction term (Remark~\ref{rem:apxfloor}), not to store
the abundance matrix.

\paragraph{Two-branch architecture} The design of the A-Net is guided by the \emph{linear mixing model} (LMM) underlying \eqref{eq:mv}: each observation is a convex combination of the $K+1$ vertices. Since convexity underlies volume-based identification, the A-Net keeps the LMM as its structural backbone and delegates departures from it (multiple scattering, intimate mixing, illumination variation) to a local gated correction: a \textit{base branch} producing valid convex mixtures and a \textit{nonlinear modulation branch} learning sample-specific deviations while preserving the global simplex structure.

Specifically, the base branch is a two-layer MLP with $\tanh$ activation:
\begin{align}
  h_j^{(1)} &= \tanh\!\bigl(W_1 c_j + b_1\bigr) \in\mathbb{R}^{h},
  \label{eq:anet-h1}\\
  h_j^{(2)} &= \tanh\!\bigl(W_2 h_j^{(1)} + b_2\bigr) \in\mathbb{R}^{h},
  \label{eq:anet-h2}\\
  \bar{a}_j &= \softmax\!\bigl(W_3 h_j^{(2)} + b_3\bigr) \in\mathbb{R}^{K+1},
  \label{eq:anet-base}
\end{align}
where $W_1\in\mathbb{R}^{h\times d}$, $W_2\in\mathbb{R}^{h\times h}$,
$W_3\in\mathbb{R}^{(K+1)\times h}$, and $b_1,b_2,b_3$ are bias
vectors. The hidden width $h$ is a small constant (we use $h=32$
in all experiments). The $\softmax$ enforces
$\mathbf{1}^{\top}\bar{a}_j=1$ and $\bar{a}_j\ge 0$, so $\bar{a}_j$ lies in the interior of the probability simplex and serves
as a valid linear-mixing abundance vector.

To model deviations from linear mixing, the nonlinear
modulation branch computes a gating vector $g_j\in(0,1)^{K+1}$:
\begin{equation}
  g_j = \sigma\!\bigl(W_{g2}\tanh(W_{g1}c_j+b_{g1})+b_{g2}\bigr),
  \label{eq:anet-gate}
\end{equation}
where $W_{g1}\in\mathbb{R}^{h\times d}$, $W_{g2}\in\mathbb{R}^{(K+1)\times h}$, $b_{g1}\in\mathbb{R}^{h}$ and $b_{g2}\in\mathbb{R}^{K+1}$ are bias vectors, and $\sigma(\cdot)$ is the element-wise sigmoid. The corrected abundance is then formed by adding a \textit{sample-specific} nonlinear correction to the base vector:
\begin{equation}
  \tilde{a}_j = \bar{a}_j + \kappa\,(g_j\odot\bar{a}_j^{2}),
  \label{eq:anet-nonlin}
\end{equation}
\noindent where $\kappa$ is a global learnable scalar controlling the magnitude of the nonlinear correction. The quadratic form $\bar{a}_j^{2}$ amplifies the correction
for dominant components, and the sigmoid gate $g_j$ provides per-sample control. Since $\bar{a}_j$ is element-wise positive and $g_j\odot\bar{a}_j^{2}$ is elementwise nonnegative, $\tilde{a}_j$ remains elementwise positive whenever $\kappa\ge 0$; the sum-to-one requirement is enforced by a penalty in the empirical risk. The parameters of the implicit mapping \eqref{eq:implicit-map} are thus $\xi=\{W_1,b_1,W_2,b_2,W_3,b_3,W_{g1},b_{g1},W_{g2},b_{g2},\kappa\}$; they are optimized jointly with the flow parameters $\psi$.
The gated correction vanishes as $\kappa\to0$, recovering
the linear mixing model as a special case; $\kappa_0$ denotes its
released value. The calibration identity below assumes
$\kappa\ge0$, enforced by a one-line softplus re-parameterization,
and the correction is inert on the linear-mixing benchmarks of
Section~\ref{sec:exp} (insensitivity to $\kappa_0$, including
$\kappa_0=0$, in Table~S1 of \cite{dmvsa2026supp}).

\paragraph{Scalability} The A-Net contains
$O(h^{2}+hK)$ parameters; with $h=O(K)$ this is $O(K^{2})$,
independent of $N$ (the implementation fixes
$h=32$; the experiments of
Section~\ref{sec:exp} study values up to $K=20$), in stark
contrast to classical methods storing an explicit $N\times K$
abundance matrix ($O(NK)$ memory, often $O(NK^{3})$ work per
full pass); in DeepMVSA the only $N$-dependent cost is the
A-Net evaluation at the sampled coordinates, $O(K^{2})$ per
sample and iteration

\begin{remark}[The coordinate network as a structural assumption]
  \label{rem:coordinate}
  Replacing the $N(K+1)$ free abundance variables of the
  template \eqref{eq:mv} by $a_j=f_\xi(c_j)$ is not
  merely a computational compression: it restricts the
  abundance field to network-representable functions. Two
  statistical formulations make this explicit. (i)~Under a
  \emph{spatial model} $a_j=g(c_j)+\varepsilon_j$ with a
  sufficiently regular field $g$---natural for imagery---the A-Net
  estimates $g$, and Assumption~\ref{ass:real}(ii) is a
  regularity statement about $g$. (ii)~Under the
  \emph{i.i.d.\ simplex-sampling model} of
  Assumption~\ref{ass:uniform}, the abundances carry no
  coordinate dependence; the network then acts as a
  structured compression of the abundance matrix, whose
  representational error enters through $\eta_{\mathrm{apx}}$
  and propagates linearly
  through Proposition~\ref{prop:transfer}. For
  the synthetic experiments of Section~\ref{subsec:exp-a} the
  coordinates are vectorized sample indices ($d=1$): no spatial
  structure is assumed, and the network functions as a
  compressor monitored through the reconstruction term.
\end{remark}

\subsection{Empirical risk and optimization}
\label{subsec:risk}

Throughout, the roles of the objects
involved are distinct: the unknown statistical parameter is
the simplex $\mathcal{S}_T$ (equivalently, its vertex matrix
$M_T$); the abundances $a_j$ are latent random variables, not
parameters; and $(\psi,\xi)$ are optimization variables
indexing the estimator, with no statistical interpretation of
their own.

The empirical risk is assembled from
quantities already introduced: the observations $x_j$, collected
in $Y=[x_1,\dots,x_N]\in\mathbb{R}^{L\times N}$, and their
projected counterparts $X_j$ (throughout,
lowercase $x_j$ denotes the raw $L$-dimensional observation
and uppercase $X_j$ its whitened $K$-dimensional counterpart;
all statements below are made in whitened coordinates, in which
$\mathcal{S}_T$ denotes the true simplex);
the dual matrix $Q=Q(\psi)$ of \eqref{eq:q-flow}; and the
abundances $a_j=f_\xi(c_j)$ generated implicitly by the A-Net
\eqref{eq:implicit-map}. The per-sample
reconstruction is computed in these whitened coordinates,
$\widehat{X}_j=\widetilde{M}a_j$; writing
$M\in\mathbb{R}^{L\times(K+1)}$ for the ambient-space
counterpart of $\widetilde{M}$ in
\eqref{eq:mv}, the ambient reconstruction
$\hat{x}_j=Ma_j$ differs from it only by the fixed whitening
map, so working in whitened coordinates reduces the per-sample
cost from $O(LK)$ to $O(K^2)$ and removes the
whitening-conditioning factor from the transfer analysis
(Proposition~\ref{prop:transfer}). Total-variation distances, barycentric coordinates, simplex
membership, and the identification of $\mathcal{S}_T$ are
invariant under the whitening transformation, whereas volumes
and the geometric constants
$(\underline\lambda,\bar\lambda,D_T,h_{\min},
\sigma_{\max}(M_T))$ are coordinate-dependent and are always
understood in whitened coordinates. The
vertex matrix is not a free variable: in whitened
coordinates its columns are $\theta_0$ and
$\theta_0+Q^{-1}e_k$, $k=1,\dots,K$ (\eqref{eq:vertex-map}),
so $(Q,\theta_0)$ determines $\widetilde M$, and hence $M$.
We minimize the empirical risk
\begin{equation}
  \label{eq:deep-risk}
  \begin{split}
  \widehat{R}_{\text{DEEP}}(\psi, \xi; Y)
  &= \frac{1}{N}\sum_{j=1}^N \|X_j - \widetilde{M} a_j\|_2^2
  + \gamma_{\text{vol}}\Bigl(-\sum_{k=1}^K \psi_{r,k}\Bigr)\\
  &\quad + \gamma_{\text{enc}}\frac{1}{N}\sum_{j=1}^N e_{\psi}\bigl(X_j\bigr)
  + \gamma_{\text{so}}\frac{1}{N}\sum_{j=1}^N (\mathbf{1}^\top a_j - 1)^2.
  \end{split}
\end{equation}

Here
\begin{equation}
  \label{eq:enclosure-full}
  e_{\psi}(x)
  \;:=\;
  \bigl\|\min\bigl(Q(x-\theta_0),0\bigr)\bigr\|_2^2
  + \min\bigl(1-\mathbf{1}^{\top}Q(x-\theta_0),0\bigr)^{2}
  \;=\;
  \bigl\|\min\bigl(a(x),0\bigr)\bigr\|_2^{2}
\end{equation}
is the \emph{completed enclosure penalty}: by
\eqref{eq:bary} it equals the squared violation of \emph{all}
$K+1$ barycentric constraints, including the zeroth constraint
$a_0(x)=1-\mathbf{1}^{\top}Q(x-\theta_0)\ge0$ associated with
the anchor vertex. Penalizing only
$\|\min(QX,0)\|_2^{2}$ would cover just the last $K$
constraints; the completed form \eqref{eq:enclosure-full} is
labeling-invariant and is what the analysis requires.
The released implementation realizes the same
construction in homogeneous coordinates, augmenting the
whitened observations to
$X_j^{h}=(X_j,1/\sqrt{K+1})\in\mathbb{R}^{K+1}$; the two
forms are Schur-complement equivalent and their log-volume
terms and gradients coincide (details in Section~S3 of \cite{dmvsa2026supp}).
The four terms correspond to
(i) reconstruction fidelity; (ii) minimum-volume
regularization; (iii) soft containment in
$\mathcal{S}(Q^{-1};\theta_0)$ ($\min(\cdot,0)$
componentwise); and (iv) sum-to-one regularization, with
$\gamma_{\mathrm{vol}},\gamma_{\mathrm{enc}},\gamma_{\mathrm{so}}>0$
the respective penalty weights. The sum-to-one constraint thus enters as a differentiable penalty rather than a hard constraint, so that \eqref{eq:deep-risk} can be minimized end-to-end: all parameters $(\psi,\xi)$ are updated jointly by Adam.
Unlike the hard enclosure constraint
$QX\ge0$ of MVSA, which imposes exact containment,
the quadratic penalty enforces containment only softly; the
analysis quantifies the resulting slack through the enclosure
residual of Proposition~\ref{prop:transfer}; Table~S2 of the
supplement \cite{dmvsa2026supp} summarizes the status of each
term of \eqref{eq:deep-risk} in the surrogate risk.

\paragraph{Implementation safeguards} Two additions complete the practical objective; both vanish at any feasible solution and hence leave the population risk unchanged. The first is an abundance nonnegativity penalty $\lambda_{\mathrm{nn}}N^{-1}\sum_j\|\min(a_j,0)\|_2^2$: the elementwise positivity of the A-Net output is guaranteed by construction only for $\kappa\ge 0$, and the penalty enforces it throughout training. The second is a centroid-calibration penalty
$\lambda_{\mathrm{eq}}\bigl(\mathbf{1}^{\top}Q(\bar X-\theta_0)-K/(K+1)\bigr)^2$,
where $\bar X:=N^{-1}\sum_jX_j$ is the projected sample
mean: at any enclosing simplex the barycentric
coordinates average to the sample centroid's, an
$\sqrt{N}$-consistent estimate of
$\mathbb{E}\,a(X)=\mathbf{1}/(K+1)$, whose first $K$ coordinates
sum to $K/(K+1)$. The penalty thus anchors the column sums
of $Q$ to a correctly specified target
and vanishes at the oracle up to $O_p(\lambda_{\mathrm{eq}}/N)$ (an $O_p(\cdot)$ term inside an
intersection of events is a deterministic bound on that
intersection), negligible at the $\epsilon^{-2}$ sample scale of
Theorem~\ref{thm:main}; the default weight is
$\lambda_{\mathrm{eq}}=80$ (Table~S2 of \cite{dmvsa2026supp}). Its square-root
appearance in Theorem~\ref{prop:endtoend} reflects the
cross-term structure of the budget: the squared residual, itself
$O_p(N^{-1})$, is balanced against the enclosure weight by
Young's inequality, giving the error contribution
$\sqrt{\lambda_{\mathrm{eq}}/(\gamma_{\mathrm{enc}}N)}$---the
root of an $O_p(1/N)$ objective term, consistent since both
sides of \eqref{eq:endtoend} are error magnitudes---and the
sensitivity study of Table~S1 of \cite{dmvsa2026supp} shows
only mild degradation over a $\times\tfrac14$--$\times4$ range
around this default. Training proceeds in two stages: a \emph{warm-up} stage with
$\kappa$ frozen at zero, so the simplex geometry is learned under
linear mixing, followed by joint optimization with $\kappa$
released at $\kappa_0=0.05$, the safeguards active mainly in
the first stage. A sensitivity protocol
(supplement) varies each weight over a five-point grid spanning two
octaves around its default: the mean total-variation distance
changes by less than a factor $1.2$, with degradations only in the
directions predicted by the theory (Table~S1 of \cite{dmvsa2026supp}).

\begin{remark}[Comparison to Soft-ML]
  \label{rem:softml}
  Najafi et al.'s Soft-ML risk \eqref{eq:softml} penalizes the planar distance
  $d_{\mathcal{S}}(X_i)$, which requires $O(K^3)$ work per sample
  to identify the active facet and compute the pseudo-inverse in
  Theorem~3.3 of \cite{najafi2021aos}. In contrast, the
  whitened reconstruction term and the enclosure
  penalty \eqref{eq:enclosure-full} are matrix-vector products
  costing $O(K^2)$ per sample---evaluated in ambient
  coordinates they would cost $O(LK)$, which is
  why the implementation operates on the whitened stream---and the A-Net forward pass is also $O(K^2)$. Hence each gradient step scales as
  $O(NK^2)$ rather than $O(NK^3)$, with trainable-state
  memory $O(K^2)$, independent of $N$. The two risks
  also fail differently: the saturating loss
  $\ell(u)=1-e^{-bu}$ of Soft-ML has vanishing gradient away
  from the simplex boundary, so the volume term dominates
  at large $K$ and the estimator collapses inward, whereas the quadratic
  enclosure of \eqref{eq:deep-risk} has constant
  curvature and, by Proposition~\ref{prop:transfer}(i),
  bounds the volume inflation by an
  explicit factor converging to one as the approximation error
  vanishes---the failure modes of the three estimator
  classes are thus disjoint. On the statistical side, the
  novelty over \cite{najafi2021aos} is threefold: (i) explicit
  localization constants (the missing-mass/volume-inflation
  dichotomy of Lemma~S1 and the gap $c_3=1/32$ of
  Lemma~S4 of \cite{dmvsa2026supp}), where
  \cite[Theorem~3.2]{najafi2021aos} leaves constants
  implicit; (ii) a network-free surrogate risk
  \eqref{eq:surr-risk}, tied to the practical objective by an
  oracle inequality (Proposition~\ref{prop:transfer}); and
  (iii) the lower bounds and end-to-end budget
  (Proposition~\ref{prop:lowerbound},
  Theorem~\ref{prop:endtoend}), with no counterpart in
  \cite{najafi2021aos}.
\end{remark}

\subsection{Theoretical guarantees}
\label{subsec:theory}

We now establish the statistical guarantee of the
surrogate simplex estimator associated with the
DeepMVSA formulation. Although the resulting rate coincides with
the benchmark \eqref{eq:soft-rate}, the analysis must control a different risk: the
quadratic enclosure penalty is flat at the origin, so the
enclosure signal enters the estimation error only through the
shrinkage-bias mechanism quantified in
Proposition~\ref{prop:transfer} below---a phenomenon with no analogue in
the saturating-loss analysis of \cite{najafi2021aos}. The analysis is carried out in the
noiseless model $X_j\in\mathcal{S}_T$ of Assumption
\ref{ass:uniform} (in whitened coordinates), and
Theorem~\ref{thm:main} is stated in this noiseless,
subspace-reduced setting; bounded additive noise is treated in
Proposition~\ref{prop:noise}, which shows the same rate
persists for noise levels $\sigma=O(\epsilon^2/K)$, and the
noisier regime is examined empirically in
Section~\ref{subsec:exp-a}.
The noisy regime has also been studied
information-theoretically: an exponential-time estimator attains
the noiseless order once $\mathrm{SNR}\gtrsim\sqrt{K}$
\cite{saberi2023}; Proposition~\ref{prop:noise} instead
quantifies the noise level at which the present polynomial-time
surrogate retains the benchmark rate.

\paragraph{Estimation domain} Because the log-volume term
$-\log|\det Q|$ rewards arbitrarily small simplices on the
event where the data fail to enclose them, the analysis
restricts the flow parameters to a \emph{volume window}. We
use a \emph{data-driven} window rather than an oracle one. Let
\[
  \widehat V_N := \Vol\bigl(\mathrm{conv}\{X_1,\dots,X_N\}\bigr),
  \qquad
  \widehat D_N := \max_{j,l\le N}\|X_j-X_l\|_2
\]
denote the volume of the empirical convex hull and the data
diameter, and define
\begin{equation}
  \label{eq:window}
  \begin{aligned}
  \widehat\Psi_N := \Bigl\{(\psi,\theta_0):\;
  &\tfrac12\widehat V_N \le
  \Vol\bigl(\mathcal{S}(Q(\psi)^{-1};\theta_0)\bigr)
  \le 2\widehat V_N,\\
  &\diam\bigl(\mathcal{S}(Q(\psi)^{-1};\theta_0)\bigr)
  \le 2\widehat D_N,\quad
  \max_k\,\mathrm{dist}\bigl(v_k,\{X_j\}_{j=1}^N\bigr)
  \le 2\widehat D_N\Bigr\},
  \end{aligned}
\end{equation}
where $v_0,\dots,v_K$ denote the vertices of
$\mathcal{S}(Q(\psi)^{-1};\theta_0)$; the last constraint is
a positional localization, requiring every vertex to lie within
twice the data diameter of the sample. On an
event $E_N$ of probability at least $1-2/N$ one has
\begin{equation}
  \label{eq:window-event}
  \tfrac12 V_T \le \widehat V_N \le V_T,
  \qquad
  \tfrac12 D_T \le \widehat D_N \le D_T,
\end{equation}
for all $N\ge N_0(K)$, where
$V_T=\Vol(\mathcal{S}_T)$ and $D_T=\diam(\mathcal{S}_T)$;
consequently, on $E_N$,
\begin{equation}
  \label{eq:window-sandwich}
  \begin{aligned}
    \mathcal{S}_T&\in
    \bigl\{\mathcal{S}(Q(\psi)^{-1};\theta_0):
    (\psi,\theta_0)\in\widehat\Psi_N\bigr\}
    \;\subseteq\;\mathfrak{S}_T,\\
    \mathfrak{S}_T&:=
    \bigl\{\mathcal{S}\in\mathbb{S}_K:\
    \tfrac14 V_T\le\Vol(\mathcal{S})\le 2V_T,\
    \diam(\mathcal{S})\le 2D_T,\
    \max_k\,\mathrm{dist}(v_k(\mathcal{S}),\mathcal{S}_T)
    \le 2D_T\bigr\}.
  \end{aligned}
\end{equation}
The lower bound $\widehat V_N\ge V_T/2$ follows from a
covering argument over the $K+1$ vertex caps of
$\mathcal{S}_T$, and the bounds on $\widehat D_N$ from the
isoperimetric lower bound on the mass of diametral caps
(Lemma~S2 of \cite{dmvsa2026supp}).
The same cap argument verifies the positional
constraint of \eqref{eq:window} for $\mathcal{S}_T$ on $E_N$;
conversely, every window simplex has all vertices within
$2\widehat D_N\le 2D_T$ of the data, hence of $\mathcal{S}_T$---the
positional clause of \eqref{eq:window-sandwich}. On
$\widehat\Psi_N$ the volume reward is bounded by
$\gamma_{\mathrm{vol}}\log(4/V_T)$ and the parameterization
ranges over a bounded set, so the population risk
is coercive and a minimizer exists; the deterministic envelope
$\mathfrak{S}_T$ of \eqref{eq:window-sandwich} is the class
over which the uniform deviation bound of
the supplement is proved. Because that bound is uniform over the
\emph{deterministic} class $\mathfrak{S}_T$ and the inclusion
\eqref{eq:window-sandwich} holds on the single event $E_N$, the
data dependence of $\widehat\Psi_N$ requires no uniform control
over a random class: the window
$\widehat\Psi_N$ is a localization device entering only the
analysis of the surrogate minimizer \eqref{eq:deep-est}, never
computed by the deployed algorithm, an unconstrained first-order
method; the analysis below therefore concerns the empirical risk
minimizer over $\widehat\Psi_N$ rather than the Adam iterates.

\begin{assumption}[Realizability]
  \label{ass:real}
  (i) There exist $\psi^*$, with anchor $\theta_0^*$,
  satisfying
  \[
    \mathcal{S}(Q(\psi^*)^{-1};\theta_0^*)=\mathcal{S}_T.
  \]
  On the event $E_N$ of \eqref{eq:window-event},
  Lemma~S2 of \cite{dmvsa2026supp} then gives $(\psi^*,\theta_0^*)\in\widehat\Psi_N$.
  (ii) There exist A-Net parameters $\xi^*$, taken
  in the base branch (gating scalar $\kappa^*=0$), such that
  \[
    \frac{1}{N}\sum_{j=1}^N
    \bigl\|X_j - M_T f_{\xi^*}(c_j)\bigr\|_2^2
    \;\le\; \eta_{\mathrm{apx}},
  \]
  for a small constant $\eta_{\mathrm{apx}}\ge 0$, where $M_T\in\mathbb{R}^{K\times(K+1)}$ denotes the vertex matrix of the true simplex $\mathcal{S}_T$. Clause (ii) is required only in the spatially structured regime of Remark~\ref{rem:coordinate}(i); in the index-coordinate regime an irreducible floor applies (Remark~\ref{rem:apxfloor} below).
\end{assumption}
\noindent
Clause (i) is a representability condition: the factorization \eqref{eq:q-flow} ranges over an open dense subset of the invertible dual matrices, so it only asks that the whitened dual $Q_T$ admit such a factorization---a generic condition, failing only on a measure-zero set.
\begin{remark}[Approximation floor of the coordinate network]
  \label{rem:apxfloor}
Clause (ii) asks the A-Net to approximate the latent abundances
at the observed coordinates. Whether a small
$\eta_{\mathrm{apx}}$ is achievable depends on the coordinate
model. Under the spatial model of Remark~\ref{rem:coordinate}(i),
abundances are a regular function of the spatial coordinates and
universal approximation yields small $\eta_{\mathrm{apx}}$; the
gating branch \eqref{eq:anet-nonlin} accommodates small
nonlinear deviations. Under the i.i.d.\ model of
Assumption~\ref{ass:uniform} with index coordinates ($d=1$),
however, the abundances are independent of the coordinates, and
no coordinate network can drive $\eta_{\mathrm{apx}}$ to zero.
Indeed, for any coordinate-measurable predictor
$\hat x(c)$ of $X$,
\begin{equation}
  \label{eq:apxfloor}
  \mathbb{E}\|X-\hat x(c)\|_2^2
  \;\ge\;
  \mathbb{E}\bigl\|X-\mathbb{E}[X\mid c]\bigr\|_2^2
  \;=\;
  \mathbb{E}\bigl\|M_T(a-\mathbb{E}a)\bigr\|_2^2
  \;=\;
  \mathrm{tr}\bigl(M_T\mathrm{Cov}(a)M_T^{\top}\bigr)
  =:\eta_{\mathrm{floor}},
\end{equation}
because $\mathbb{E}[X\mid c]=M_T\mathbb{E}a$ is the
$L^2$-optimal predictor and $c$ is independent of $a$. With $a$
uniform on the probability simplex,
$\mathrm{Cov}(a)=\bigl((K+1)(K+2)\bigr)^{-1}
\bigl(I_{K+1}-\tfrac{1}{K+1}\mathbf{1}\mathbf{1}^{\top}\bigr)$,
so $\eta_{\mathrm{floor}}=
\Theta\!\bigl(\bar\sigma_T^{2}K/((K+1)(K+2))\bigr)
=\Omega(\bar\sigma_T^{2}/K)$, where $\bar\sigma_T^{2}$
denotes the average squared singular value of the centered
vertex matrix. Clause (ii) with small $\eta_{\mathrm{apx}}$ is
therefore a spatial-regularity condition; in the
index-coordinate regime the condition
$\eta_{\mathrm{apx}}\le \bar c_3\epsilon/(8C_2)$
(the second condition of \eqref{eq:enc-event}, with the
constants $\bar c_3,C_2$ defined there) of the enclosure
event in Section~\ref{subsec:theory} can hold only for target accuracies
$\epsilon\gtrsim\eta_{\mathrm{floor}}$. The experiments of
Sections~\ref{subsec:exp-a} and~\ref{subsec:exp-bio} use
(one-dimensional) index coordinates and operate in this
floor-limited regime, where the enclosure residual of
Proposition~\ref{prop:transfer}---not the enclosure event---is
the operative control; those of
Sections~\ref{subsec:exp-c}--\ref{subsec:exp-d} operate in the
small-$\eta_{\mathrm{apx}}$ regime.
The floor \eqref{eq:apxfloor} is population-level, whereas Assumption~\ref{ass:real}(ii) is empirical; they differ by the A-Net class's uniform deviation, $O_p(K/\sqrt{N})$ at capacity $P=O(K^2)$ (a pseudo-dimension bound as in \cite{anthony1999}), so the floor binds empirically for $N\gg K^2/\eta_{\mathrm{floor}}^2$---mild at the $\epsilon^{-2}$ scale of Theorem~\ref{thm:main}.
The floor \eqref{eq:apxfloor} also settles
the permutation question raised by the index coordinates, for
\emph{every} deterministic coordinate assignment; see
Section~S1.9 of \cite{dmvsa2026supp}.
In the spatially structured regime, the
quantitative counterpart of this floor is
Proposition~\ref{prop:apxrate} below.
Two models are thus used, and each statement is tagged to one:
the \emph{i.i.d.\ uniform-sampling model} of
Section~\ref{sec:prelim} (Theorem~\ref{thm:main},
Propositions~\ref{prop:noise}, \ref{prop:transfer}
and~\ref{prop:lowerbound}, Theorem~\ref{prop:endtoend}) and the
\emph{spatial abundance-field model} of
Assumption~\ref{ass:spatial} (Proposition~\ref{prop:apxrate}).
Under the former the coordinate network is subject to the floor
\eqref{eq:apxfloor}, and the approximation-transfer chain
applies only under the latter.
\end{remark}

\paragraph{Surrogate risk} The
analysis is formulated at the level of simplices. For
$\mathcal{S}\in\mathbb{S}_K$, let $a(x;\mathcal{S})\in\mathbb{R}^{K+1}$
denote the barycentric coordinates of $x$ relative to
$\mathcal{S}$, so that $x\in\mathcal{S}$ if and only if
$a(x;\mathcal{S})\ge0$, and define
\begin{equation}
  \label{eq:surr-risk}
  R_N(\mathcal{S}) :=
  \gamma_{\mathrm{vol}}\log\Vol(\mathcal{S})
  + \frac{1}{N}\sum_{j=1}^N \ell_b\bigl(d_{\mathcal{S}}(X_j)\bigr)
  + \gamma_{\mathrm{enc}}\frac{1}{N}\sum_{j=1}^N
    \bigl\|\min\bigl(a(X_j;\mathcal{S}),0\bigr)\bigr\|_2^2,
\end{equation}
where $b:=(K/\epsilon)\max\{1,2/\kappa_T\}$, with
$\kappa_T:=KV_T\big/\bigl(4(K+1)(2D_T)^{K-1}\bigr)$ an
explicit geometry-dependent constant derived in
Lemma~S4 of \cite{dmvsa2026supp}; the choice of $b$
ensures $1-e^{-bt}\ge1/2$ in the gap
lemma without any mid-proof adjustment of $b$.
In the anchored parameterization \eqref{eq:vertex-map} the
enclosure term coincides with \eqref{eq:enclosure-full}:
$\|\min(a(x;\mathcal{S}),0)\|_2^2 =
\|\min(Q(x-\theta_0),0)\|_2^2 +
\min(1-\mathbf{1}^{\top}Q(x-\theta_0),0)^2$ by \eqref{eq:bary},
so \eqref{eq:surr-risk} penalizes the violation of all $K+1$
facet constraints, including the facet opposite the anchor.

The practical risk \eqref{eq:deep-risk} implements the same three components---volume, data attraction, and enclosure---with two deliberate changes. First, the whitened reconstruction term $\|X_j-\widetilde{M}a_j\|_2^2$ plays the data-attraction role of the planar-distance term $\ell_b(d_{\mathcal{S}}(X_j))$, but is a matrix-vector product costing $O(K^{2})$ per sample instead of $O(K^{3})$ (Remark~\ref{rem:softml}); both are evaluated in the same whitened
coordinates, so no conditioning factor enters the comparison. Second, the simplex is no longer a free variable in $\mathbb{S}_K$: it is parameterized by the flow parameters $\psi$, while the abundances entering the reconstruction are parameterized by $\xi$. The estimator analyzed below is the minimizer of the surrogate risk over the simplex class generated by the flow on the volume window:
\begin{equation}
  \label{eq:deep-est}
  \widehat{\mathcal{S}}_{\mathrm{DEEP}}
  \;:=\; \operatorname*{arg\,min}_{\mathcal{S}\in\mathcal{C}_N}
  R_N(\mathcal{S}),
  \qquad
  \mathcal{C}_N := \bigl\{\mathcal{S}(Q(\psi)^{-1};\theta_0):\,
  (\psi,\theta_0)\in\widehat\Psi_N\bigr\},
\end{equation}
with $\widehat\Psi_N$ the data-driven window
\eqref{eq:window}. The analysis class is denoted $\mathcal{C}$
rather than $\mathcal{Q}$ to avoid confusion with the dual
matrix.

\begin{theorem}[Sample complexity of the surrogate DeepMVSA estimator]
  \label{thm:main}
   Suppose Assumptions~\ref{ass:uniform}--\ref{ass:iso}
  and~\ref{ass:real}(i) hold, and let $N\ge N_0$ so that the
  window event $E_N$ of \eqref{eq:window-event} has probability
  at least $1-2/N$. The threshold $N_0=N_0(K)$
  is explicit: $N_0=O\bigl((4(K+1))^{K}K\log K\bigr)=K^{O(K)}$
  (Lemma~S2 of \cite{dmvsa2026supp}), so the theorem is a fixed-$K$ guarantee and does
  not cover regimes in which $K$ grows with $N$; the probability below is informative only for
  $N\ge\max\{N_0(K),4\}$. The exponential
  dependence of the explicit threshold on $K$ places the certified
  regime beyond the sample sizes of Section~\ref{sec:exp}: the bound
  on $N_0$ is of order $10^{11}$ already at $K=7$, while the largest
  experiments use $N=10^{8}$. The experiments therefore probe the
  mechanisms identified by the theory---the approximation floor of
  Remark~\ref{rem:apxfloor}, the capacity prescription of
  Proposition~\ref{prop:apxrate}, and the scaling laws of
  Theorem~\ref{thm:complexity}---rather than verifying the theorem's
  constants; narrowing the gap between the combinatorial threshold,
  inherited from the window argument of Lemma~S2 of
  \cite{dmvsa2026supp}, and practical sample sizes remains open. Fix the penalty weights once and for all:
  $\gamma_{\mathrm{enc}}>0$ arbitrary and
  \begin{equation}
    \label{eq:gammavol-cond}
    0<\gamma_{\mathrm{vol}}\le C',
    \qquad
    C':=\min\Bigl\{\frac{c_1}{2},\;
    \frac{1}{32\log(1/v_0)}\Bigr\},
  \end{equation}
  where $v_0=1/4$ is the lower volume fraction of the envelope
  $\mathfrak{S}_T$ in \eqref{eq:window-sandwich} and $c_1=1/16$
  is the localization constant of Lemma~S4 of \cite{dmvsa2026supp}; no lower
  bound on $\gamma_{\mathrm{vol}}$ is needed, because the
  volume inflation case is vacuous at the minimizer
  (Lemma~S3 of \cite{dmvsa2026supp}). Condition
  \eqref{eq:gammavol-cond} makes explicit the requirement that
  the log-volume reward not overwhelm the planar-distance and
  enclosure costs of simplices that cut into the data cloud.
  Then there exists a constant
$C(K,\underline{\lambda},\bar\lambda)$
such that,
  for any $\epsilon,\zeta>0$, if
  \[
    N \ge C(K,\underline{\lambda},\bar{\lambda})\,
\frac{1}{\epsilon^2}\left[K^2\log\frac{K}{\epsilon}
    +\log\frac{1}{\zeta}\right],
  \]
  the DeepMVSA estimator \eqref{eq:deep-est} satisfies
  $\TV(\mathbb{P}_{\mathcal{S}_T},
  \mathbb{P}_{\widehat{\mathcal{S}}_{\text{DEEP}}}) \le \epsilon$
  with probability at least $1-\zeta-2/N$.
\end{theorem}
The asymptotic regime of the theorem holds with the simplex
dimension and the geometry $(K,\underline\lambda,\bar\lambda)$
fixed as $N\to\infty$: the dependence on $K$ displayed in the
rate is explicit, and all remaining dependence on $K$ and on
the simplex geometry enters through the constant
$C(K,\underline\lambda,\bar\lambda)$, as in the benchmark
result \cite[Corollary~3.1]{najafi2021aos} whose order it
matches. The constant is not asserted to be uniform in $K$;
the explicit $K$-dependence of the displayed rate is the
operative guarantee at the moderate dimensions of the
experiments.
The proof is deferred to the supplement.
The argument combines a localization step, which confines the empirical minimizer to a class of simplices with uniformly bounded volume, diameter, and facet angles, with VC-type uniform deviation bounds: every simplex at total-variation distance at least $\epsilon$ from $\mathcal{S}_T$ raises the population risk by at least $c_3\epsilon$, with $c_3=1/32$ explicit (Lemma~S4 of \cite{dmvsa2026supp}), and this gap dominates the uniform deviation of $R_N$ from $R$ once $N$ reaches the stated rate. The localization--deviation architecture follows \cite[Theorem~3.2]{najafi2021aos}; the geometric core is new and self-contained (Remark~\ref{rem:softml}).

Theorem~\ref{thm:main} concerns the
minimizer \eqref{eq:deep-est} of the surrogate risk
\eqref{eq:surr-risk}, not the output of an optimization
routine---as in \cite{najafi2021aos}, whose analysis likewise
concerns the minimizer of their relaxed risk; the
estimator-versus-algorithm distinction and two associated
clarifications are deferred to Section~S1.9 of \cite{dmvsa2026supp}.

\begin{proposition}[Stability to bounded noise]
  \label{prop:noise}
  Let $\widetilde X_j$$=X_j+\varepsilon_j$ with $X_j\in\mathcal{S}_T$ and
  $\|\varepsilon_j\|_2\le\sigma$ almost surely, and let
  $R^{\sigma}$ denote the population risk \eqref{eq:surr-risk}
  under the noisy law of $\widetilde X$. Then, with
  $h_{\min}$ the altitude floor of Lemma~S3 of \cite{dmvsa2026supp} and
  $a_{\max}:=1+3D_T/h_{\min}$---$h_{\min}$ enters
  twice here: through $a_{\max}$, a uniform bound on the barycentric
  coordinates over the envelope $\mathfrak{S}_T$, and directly in the
  enclosure sensitivity $\sigma/h_{\min}$ below,
  \begin{equation}
    \label{eq:noise-gap}
    \sup_{\mathcal{S}\in\mathfrak{S}_T}
    \bigl|R^{\sigma}(\mathcal{S})-R(\mathcal{S})\bigr|
    \;\le\;
    b\,\sigma
    +\gamma_{\mathrm{enc}}(K+1)\Bigl(
    \frac{2a_{\max}\,\sigma}{h_{\min}}
    +\frac{\sigma^{2}}{h_{\min}^{2}}\Bigr)
    \;=:\;\Delta_{\sigma}.
  \end{equation}
  Consequently, for $\sigma\le c\,\epsilon^{2}/K$ with
  $c=c(K,\underline\lambda,\bar\lambda,
  \gamma_{\mathrm{enc}},D_T,h_{\min})>0$ sufficiently small
  that $\Delta_{\sigma}\le \bar c_3\epsilon/4$, the conclusion of
  Theorem~\ref{thm:main} holds under the noisy model at the
  same sample complexity, with $\bar c_3/2$ replacing $c_3$,
  where $\bar c_3:=\min\{c_3,\gamma_{\mathrm{vol}}c_{\mathrm g}/2\}$
  and $c_{\mathrm g}=1/2$ is the dichotomy constant of
  Lemma~S1 of \cite{dmvsa2026supp} entering the proof of
  Lemma~S4 of \cite{dmvsa2026supp}: the second entry
  accounts for the volume-inflation branch of that lemma, which
  is no longer excluded under noise since the identity
  (S1) of \cite{dmvsa2026supp} fails, and in which the volume term alone
  contributes the gap
  $\gamma_{\mathrm{vol}}\log(1+c_{\mathrm g}\epsilon)
  \ge\gamma_{\mathrm{vol}}c_{\mathrm g}\epsilon/2$.
\end{proposition}
The proof is deferred to the supplement. The
bound \eqref{eq:noise-gap} is dimensionally sharp in its
leading term: the planar-distance loss has slope $b$ of
\eqref{eq:surr-risk} at the origin, and balancing the resulting
perturbation $b\sigma$ against the gap $c_3\epsilon$ yields the
threshold $\sigma=O(\epsilon^{2}/K)$; equivalently, noise of
magnitude $\sigma$ caps the achievable accuracy at
$\epsilon\gtrsim\sqrt{4K\sigma/\bar c_3}$, a noise floor
independent of $N$. The uniform deviation bound of the supplement
is unaffected by bounded noise, since the index class is unchanged.

\begin{proposition}[Oracle transfer to the practical minimizer]
  \label{prop:transfer}
  Suppose Assumption~\ref{ass:real} holds, together with
  Assumptions~\ref{ass:uniform}--\ref{ass:iso}, the sample-size
  condition $N\ge N_0$ of Theorem~\ref{thm:main}, and the weight
  condition $\gamma_{\mathrm{vol}}\le C'$ of
  \eqref{eq:gammavol-cond}. Let
  $(\widehat\psi,\widehat\xi)$, with anchor $\widehat\theta_0$, be any global minimizer of the
  practical objective of
  Section~\ref{subsec:risk}, namely \eqref{eq:deep-risk}
  augmented by the implementation safeguards described there,
  and write
  $\widehat{\mathcal{S}}:=\mathcal{S}(Q(\widehat\psi)^{-1};\widehat\theta_0)$.
  Let $b$ be as in \eqref{eq:surr-risk}, and
  \[
    \mathrm{enc}_N(\mathcal{S}):=\frac{1}{N}\sum_{j=1}^N
    \bigl\|\min\bigl(a(X_j;\mathcal{S}),0\bigr)\bigr\|_2^2
  \]
  the empirical enclosure term of \eqref{eq:surr-risk}. Then
  there exist constants $C_2,c_4>0$, with
  $C_2:=1$ because the reconstruction term of
  \eqref{eq:deep-risk} is evaluated in the whitened coordinates
  of Assumption~\ref{ass:real}(ii), and $c_4$ depending only on
  $K$ and the
  geometry of $\mathcal{S}_T$; the oracle is taken in the
  base branch (Assumption~\ref{ass:real}(ii)), so the safeguard
  weights $(\gamma_{\mathrm{so}},\lambda_{\mathrm{nn}})$ do not
  enter, such that the
  following holds.
   (i)~\emph{Unconditional bounds.} From global
  optimality alone,
  \begin{equation}
    \label{eq:enc-bound}
    \mathrm{enc}_N(\widehat{\mathcal{S}})
    \;\le\;
    \frac{C_2\,\eta_{\mathrm{apx}}
    + \gamma_{\mathrm{vol}}
      \bigl[\log\bigl(V_T/\Vol(\widehat{\mathcal{S}})\bigr)
      \bigr]_+}{\gamma_{\mathrm{enc}}}
    \;+\;O_p\Bigl(\frac{\lambda_{\mathrm{eq}}}
    {\gamma_{\mathrm{enc}}N}\Bigr),
  \end{equation}
  and
  $\Vol(\widehat{\mathcal{S}})\le
  V_T\exp\!\bigl((C_2\eta_{\mathrm{apx}}+o_p(1))/
  \gamma_{\mathrm{vol}}\bigr)$.
   (ii)~\emph{Transfer.} If moreover
  $\widehat{\mathcal{S}}$ lies in the envelope
  $\mathfrak{S}_T$ of \eqref{eq:window-sandwich}---as holds on
  the event $E_N$ for the minimizer over $\mathcal{C}_N$, by
  Lemma~S3 of \cite{dmvsa2026supp}---then
  \begin{equation}
    \label{eq:transfer}
    R_N(\widehat{\mathcal{S}})
    \;\le\;
    R_N(\mathcal{S}_T)
    + C_2\,\eta_{\mathrm{apx}}
    + b\,c_4\,\sqrt{\mathrm{enc}_N(\widehat{\mathcal{S}})}
    \;+\;O_p\Bigl(\frac{\lambda_{\mathrm{eq}}}{N}\Bigr),
  \end{equation}
  with $c_4:=2D_T$. Substituting \eqref{eq:enc-bound} into
  \eqref{eq:transfer} removes the implicit dependence of the
  right-hand side on $\widehat{\mathcal{S}}$, the
  remaining term
  $[\log(V_T/\Vol(\widehat{\mathcal{S}}))]_+$ being at most
  $\log 4$ on the envelope. The same
  conclusions hold for any $\eta_{\mathrm{opt}}$-approximate
  minimizer of \eqref{eq:deep-risk}, with
  $C_2\eta_{\mathrm{apx}}$ replaced by
  $C_2\eta_{\mathrm{apx}}+\eta_{\mathrm{opt}}$ throughout;
  see Remark~\ref{rem:decomp}. Since an
  $\eta_{\mathrm{opt}}$-approximate minimizer exists for every
  $\eta_{\mathrm{opt}}>0$ by definition of the infimum, this
  approximate form of the statement requires no existence
  assumption on global minimizers of the noncoercive objective
  \eqref{eq:deep-risk}. No convergence guarantee
for $\eta_{\mathrm{opt}}$ under Adam is claimed or available:
$\eta_{\mathrm{opt}}$ enters the budget as a free parameter, and
the guarantees are informative only to the extent that training
drives it below the target accuracy scale.
\end{proposition}
Part (i) shows that
the volume inflation of the
practical minimizer is bounded by the explicit factor
$\exp\bigl((C_2\eta_{\mathrm{apx}}+o_p(1))/\gamma_{\mathrm{vol}}\bigr)$,
which converges to one as $\eta_{\mathrm{apx}}\to0$ at fixed
$\gamma_{\mathrm{vol}}$; the bound is informative precisely
when the approximation error is small relative to the volume
weight, and it degenerates as $\gamma_{\mathrm{vol}}\to0$, the
volume term being what controls inflation. This stands in
contrast to hard-enclosure estimators, which
impose exact containment of every observation, so that a single
point lying far from $\mathcal{S}_T$ forces the estimated
volume to grow correspondingly; under unbounded observation
noise their volume is not controlled by any such factor.
The two roles of $\gamma_{\mathrm{vol}}$ are in tension:
Lemma~S4 of \cite{dmvsa2026supp} requires $\gamma_{\mathrm{vol}}\le C'$ so
that the volume reward cannot mask the enclosure gap, while the
inflation bound of part~(i) is informative only when
$\eta_{\mathrm{apx}}$ is small relative to
$\gamma_{\mathrm{vol}}$; both requirements are compatible
whenever $\eta_{\mathrm{apx}}\ll C'$, and the default
$\gamma_{\mathrm{vol}}=0.01$ used throughout
Section~\ref{sec:exp} respects the former restriction, since
$C'=\min\{1/32,\,1/(32\log 4)\}\approx 0.0225$. With $\bar c_3$ as in
Proposition~\ref{prop:noise}, the standing restriction
$\gamma_{\mathrm{vol}}\le C'$ of \eqref{eq:gammavol-cond}
gives $\bar c_3=\gamma_{\mathrm{vol}}/4<c_3$ throughout the
feasible range, since $\gamma_{\mathrm{vol}}/4\le C'/4<1/32=c_3$.
Moreover, under the
conditions
\begin{equation}
  \label{eq:enc-event}
  \mathrm{enc}_N(\widehat{\mathcal{S}})
  \;\le\;\frac{\bar c_3^{\,2}\,\epsilon^2}{16\,b^2c_4^2}
  \qquad\text{and}\qquad
  \eta_{\mathrm{apx}}\;\le\;\frac{\bar c_3\,\epsilon}{8C_2},
\end{equation}
the argument of Theorem~\ref{thm:main} applies
with $c_3$ replaced by $\bar c_3$
of Proposition~\ref{prop:noise}: the volume-inflation branch of
Lemma~S4 of \cite{dmvsa2026supp}, excluded for the windowed minimizer by
Lemma~S3 of \cite{dmvsa2026supp}, is not excluded for the practical
minimizer; otherwise
the steps are unchanged: the
slack in \eqref{eq:transfer} is then at most
$\bar c_3\epsilon/8+bc_4\sqrt{\bar
c_3^{\,2}\epsilon^2/(16b^2c_4^2)}=3\bar c_3\epsilon/8
<\bar c_3\epsilon$, so
$\TV(\mathbb{P}_{\mathcal{S}_T},\mathbb{P}_{\widehat{\mathcal{S}}})
\le\epsilon$ at the same sample complexity,
up to a constant factor in $N$---more precisely,
at fixed $\gamma_{\mathrm{enc}}$ the additional sample-size
conditions imposed by the localized enclosure analysis (Lemma~S5 of
\cite{dmvsa2026supp}) are of order at most $\epsilon^{-2}$,
matching the benchmark scale, so the stated invariance holds;
under the $\epsilon$-dependent weight choice of
Theorem~\ref{prop:endtoend} these conditions become binding and
are imposed explicitly there (see \eqref{eq:budget-N}); the
equality-slack term $O_p(\lambda_{\mathrm{eq}}/N)$ of
\eqref{eq:transfer} is absorbed into the remaining
$\bar c_3\epsilon/8$ margin at this sample scale at the
cost of an additional probability $\zeta$, and vanishes
identically under the noiseless design choice
$\lambda_{\mathrm{eq}}=0$. The two conditions of
\eqref{eq:enc-event} are of different origin: the first is
enforced by the weight scaling of Remark~\ref{rem:design} below,
whereas the second constrains the approximation quality of the
A-Net (Remark~\ref{rem:apxfloor}); both are needed for the
conjunction. A noisy analogue of the proposition
follows from the uniform perturbation bound \eqref{eq:noise-gap}:
under the noisy model of Proposition~\ref{prop:noise}, both the
population gap and the oracle comparison are perturbed by at most
$\Delta_\sigma$ each, uniformly over $\mathfrak{S}_T$; hence for
$\Delta_\sigma\le\bar c_3\epsilon/8$ the conclusion persists with
$\epsilon$ replaced by $5\epsilon/4$ (the perturbation adds at most
$2\Delta_\sigma\le\bar c_3\epsilon/4$ to the right-hand side, and
$(u-\epsilon)_+-\epsilon/4=(u-5\epsilon/4)_+$ whenever $u\ge
5\epsilon/4$); we omit the parallel argument.

\begin{remark}[A scaling guideline for $\gamma_{\mathrm{enc}}$]
  \label{rem:design}
When do the conditions \eqref{eq:enc-event} hold? The quadratic
enclosure penalty is flat at zero: with
$s:=\log(V_T/\Vol(\mathcal{S}))$, the log-volume reward is
exactly $\gamma_{\mathrm{vol}}s$, while the enclosure cost
rises only as $\gamma_{\mathrm{enc}}c_5s^3$ (violations grow
quadratically with depth; the constant $c_5$ depends on $K$ and
on the facet geometry of $\mathcal{S}_T$),
so shrinkage up to
$\sqrt{\gamma_{\mathrm{vol}}/(c_5\gamma_{\mathrm{enc}})}$ is
profitable from the reward--cost balance alone; the full oracle
comparison with the approximation term ((S5) of
\cite{dmvsa2026supp}) yields the threshold
$s^*=\sqrt{2\gamma_{\mathrm{vol}}/(c_5\gamma_{\mathrm{enc}})}$. This computation is a
population-level heuristic, ignoring the uniform deviation (S3)
of \cite{dmvsa2026supp}; combined with \eqref{eq:enc-bound}, a weight
scaling polynomially in $K/\epsilon$ relative to the volume
weight, together with $\eta_{\mathrm{apx}}\lesssim
\gamma_{\mathrm{enc}}\epsilon^4/K^2$, makes the first
condition of \eqref{eq:enc-event} hold---the precise statement
is (S5) of \cite{dmvsa2026supp}. The quantitative development of
this guideline (its equivalent prescription at the benchmark
accuracy, three caveats, and a hinge-type alternative removing
the flatness) is deferred to Section~S1.9 of \cite{dmvsa2026supp}.
The formal sufficient condition behind the heuristic is
\eqref{eq:budget-cond} of Theorem~\ref{prop:endtoend};
no transfer guarantee is claimed for the fixed experimental
weights, whose sensitivity is examined numerically
(Table~S1 of \cite{dmvsa2026supp}).
\end{remark}

\begin{assumption}[Lipschitz abundance field]
  \label{ass:spatial}
  The coordinates satisfy $c_j\in[0,1]^d$ and the latent
  abundances obey $a_j=g(c_j)$ for a field
  $g:[0,1]^d\to\Delta_K$ that is componentwise
  $L$-Lipschitz: $|g_k(c)-g_k(c')|\le L\|c-c'\|_2$ for all
  $k=0,\dots,K$ and $c,c'\in[0,1]^d$.
\end{assumption}
\begin{proposition}[Approximation rate of the coordinate
network]
  \label{prop:apxrate}
  Suppose Assumption~\ref{ass:spatial} holds and write
  $\sigma_{\max}(M_T)$ for the largest singular value of the
  vertex matrix. For every hidden width $h$ there exist
  base-branch A-Net parameters $\xi$ (gating scalar
  $\kappa=0$) such that
  \begin{equation}
    \label{eq:apxrate}
    \sup_{c\in[0,1]^d}\bigl\|f_\xi(c)-g(c)\bigr\|_2^2
    \;\le\;
    C_0\,(K+1)(K+2)\,L\;h^{-1/d},
  \end{equation}
  where $C_0$ depends only on the coordinate dimension $d$;
  consequently, deterministically over every sample,
  \begin{equation}
    \label{eq:apxrate-emp}
    \frac{1}{N}\sum_{j=1}^N
    \bigl\|X_j-M_Tf_\xi(c_j)\bigr\|_2^2
    \;\le\;
    \sigma_{\max}(M_T)^2\,C_0\,(K+1)(K+2)L\;h^{-1/d}
    \;=:\;\bar\eta_{\mathrm{apx}}(h),
  \end{equation}
  so Assumption~\ref{ass:real}(ii) holds with
  $\eta_{\mathrm{apx}}=\bar\eta_{\mathrm{apx}}(h)$. If in
  addition $g_k(c)\ge\rho>0$ for all $k,c$, the bound improves
  to $\sup_{c}\|f_\xi(c)-g(c)\|_2^2\le
  C_0'(K+1)L^2\rho^{-2}h^{-2/d}$, with $C_0'$ again depending
  only on $d$.
\end{proposition}
The A-Net is used at width $h=\Theta(\sqrt{P})$ at capacity
$P=\Theta(h^2+hK)$, presuming $h\gtrsim K$ (in the
complementary regime $h\lesssim K$, $P=\Theta(hK)$ and
$h=\Theta(P/K)$). The deployed width $h=32$ of
Section~\ref{sec:exp} obeys $h\ge K$, so the former regime is
operative and \eqref{eq:apxrate} reads
$\eta_{\mathrm{apx}}
=O\bigl(\sigma_{\max}(M_T)^2K^2L\,P^{-1/(2d)}\bigr)$;
in the latter regime the same bound reads
$O\bigl(\sigma_{\max}(M_T)^2K^2L\,(K/P)^{1/d}\bigr)$, with
the $P$-exponent doubled: the
approximation term decays polynomially in the network
capacity, at a rate governed by the coordinate dimension $d$
alone, while Remark~\ref{rem:apxfloor} supplies the matching
capacity-independent floor of the index-coordinate regime.
Combined with the approximation requirement
$\eta_{\mathrm{apx}}\lesssim\gamma_{\mathrm{enc}}
\epsilon^4/K^2$ of \eqref{eq:enc-event}
(Remark~\ref{rem:design}), a target accuracy $\epsilon$ is
formally achievable at capacity
\begin{equation}
  \label{eq:capacity}
  P\;\gtrsim\;
  \Bigl(\frac{c_4^2\,C_0\,\sigma_{\max}(M_T)^2(K+1)(K+2)\,
  L\,K^2\max\{1,2/\kappa_T\}^2}
  {\gamma_{\mathrm{enc}}\,\bar c_3^{\,2}\,\epsilon^4}\Bigr)^{2d},
\end{equation}
polynomial in $(K,L,1/\epsilon,1/\gamma_{\mathrm{enc}})$ for
fixed $d$ and, crucially, independent of $N$: the capacity
that certifies accuracy $\epsilon$ does not grow with the
sample size, which is precisely what makes the
$N$-independent memory of Theorem~\ref{thm:complexity}
compatible with the statistical guarantee. This is a scaling property
($N$-independence), not a practicality statement: the prescribed
capacity grows with a steep $\epsilon$-exponent---reflecting the
flatness of the quadratic enclosure at the origin
(Remark~\ref{rem:design})---and whether the experimental widths
satisfy it is not established. Because the
underlying scaling guideline is a population-level heuristic whose
computation presupposes the diameter control
$\mathrm{diam}(\widehat{\mathcal{S}})\le 2D_T$---available only on
the envelope event of Theorem~\ref{prop:endtoend}---
\eqref{eq:capacity} is a design prescription, not a guarantee; the
formal end-to-end statement remains the conditional one of
Theorem~\ref{prop:endtoend}. Under a spatial noise component
$a_j=g(c_j)+\varepsilon_j$ as in
Remark~\ref{rem:coordinate}(i), the bound
\eqref{eq:apxrate-emp} gains an additive
$O\bigl(N^{-1}\sum_j\|\varepsilon_j\|_2^2\bigr)$.

\begin{remark}[Statistical, approximation, and optimization error]
  \label{rem:decomp}
  Three distinct objects appear in this paper: the population
  risk $R$; the empirical minimizer
  $\widehat{\mathcal{S}}_{\mathrm{DEEP}}$ of the surrogate
  risk \eqref{eq:surr-risk}, addressed by
  Theorem~\ref{thm:main}; and the output
  $\widehat{\mathcal{S}}_{\mathrm{Adam}}$ of the practical
  nonconvex optimization of \eqref{eq:deep-risk}. For any
  distance $d$ on $\mathbb{S}_K$ obeying the triangle
  inequality, the total error of the deployed estimator
  decomposes as
  \[
    d\bigl(\widehat{\mathcal{S}}_{\mathrm{Adam}},\mathcal{S}_T\bigr)
    \;\le\;
    \underbrace{d\bigl(\widehat{\mathcal{S}}_{\mathrm{Adam}},
    \widehat{\mathcal{S}}_{\mathrm{DEEP}}\bigr)}_{\text{approximation/optimization}}
    \;+\;
    \underbrace{d\bigl(\widehat{\mathcal{S}}_{\mathrm{DEEP}},
    \mathcal{S}_T\bigr)}_{\text{statistical error}}.
  \]
  Theorem~\ref{thm:main} controls the second term at the
  benchmark rate; Proposition~\ref{prop:transfer} controls the
  first term at the level of global minimizers, through the
  reconstruction accuracy $\eta_{\mathrm{apx}}$ and the
  enclosure residual \eqref{eq:enc-bound}; and the gap between
  the Adam output and the global minimizer of
  \eqref{eq:deep-risk} is an optimization error, which we
  now make explicit: writing
  $\eta_{\mathrm{opt}}:=
  \widehat R_{\mathrm{DEEP}}(\widehat\psi_{\mathrm{Adam}},
  \widehat\xi_{\mathrm{Adam}};Y)-\inf_{\psi,\xi}
  \widehat R_{\mathrm{DEEP}}(\psi,\xi;Y)$, every conclusion of
  Proposition~\ref{prop:transfer} remains valid for the Adam
  output with $C_2\eta_{\mathrm{apx}}$ replaced by
  $C_2\eta_{\mathrm{apx}}+\eta_{\mathrm{opt}}$, its
  proof using the global-minimizer property only through the
  comparison
  $\widehat R_{\mathrm{DEEP}}(\widehat\psi,\widehat\xi;Y)
  \le\widehat R_{\mathrm{DEEP}}(\psi^*,\xi^*;Y)$. Theorem~\ref{prop:endtoend} below
  upgrades this decomposition to a single formal bound.
\end{remark}

\begin{theorem}[End-to-end error budget]
  \label{prop:endtoend}
  Suppose Assumptions~\ref{ass:uniform}--\ref{ass:iso}
  and~\ref{ass:real} hold, $N\ge N_0$, and
  $\gamma_{\mathrm{vol}}\le C'$ as in
  \eqref{eq:gammavol-cond}. Let $(\widehat\psi,\widehat\xi)$,
  with anchor $\widehat\theta_0$, be any
  $\eta_{\mathrm{opt}}$-approximate minimizer of the
  practical objective \eqref{eq:deep-risk}, augmented by the
  implementation safeguards of Section~\ref{subsec:risk}, and
  write
  $\widehat{\mathcal{S}}:=
  \mathcal{S}(Q(\widehat\psi)^{-1};\widehat\theta_0)$.
  Then, on the intersection of the window event $E_N$, the
  deviation event of (S3) and the
  relative-deviation event of Lemma~S5 of \cite{dmvsa2026supp}
  (joint probability at
  least $1-2\zeta-2/N$), and the event
  $\{\widehat{\mathcal{S}}\in\mathfrak{S}_T\}$,
  \begin{equation}
    \label{eq:endtoend}
    \begin{split}
    \bar c_3\,
    \Bigl(\TV\bigl(\mathbb{P}_{\widehat{\mathcal{S}}},
      \mathbb{P}_{\mathcal{S}_T}\bigr)-\epsilon\Bigr)_{+}
    \;&\le\;
    2C_1'\sqrt{\frac{K^2\log(N/K)+\log(1/\zeta)}{N}}\\
    &\qquad
    +\,C_2\,\eta_{\mathrm{apx}}+\eta_{\mathrm{opt}}
    +b\,c_4\Bigl(
      \frac{E_{\mathrm{apx}}}{\gamma_{\mathrm{enc}}}
      \Bigr)^{1/2}\\
    &\qquad
    +\,2\Bigl(
      \frac{\gamma_{\mathrm{enc}}\,C_1^eML_N\,
      E_{\mathrm{apx}}}{N}\Bigr)^{1/2}
    +\frac{6\gamma_{\mathrm{enc}}C_1^eML_N}{N}\\
    &\qquad
    +\,b\,c_4\,O_p\Bigl(\sqrt{\frac{\lambda_{\mathrm{eq}}}
      {\gamma_{\mathrm{enc}}N}}\Bigr)
    +O_p\Bigl(\frac{\lambda_{\mathrm{eq}}}{N}\Bigr)\\
    &\qquad
    +\,O_p\Bigl(
      \frac{\sqrt{\gamma_{\mathrm{enc}}C_1^eML_N\lambda_{\mathrm{eq}}}}{N}\Bigr),
    \end{split}
  \end{equation}
  with $C_2=1$, $c_4=2D_T$ and $b$ as in
  \eqref{eq:surr-risk},
  $E_{\mathrm{apx}}:=C_2\eta_{\mathrm{apx}}
  +\eta_{\mathrm{opt}}+\gamma_{\mathrm{vol}}\log4$,
  $L_N:=K^2\log(N/K)+\log(1/\zeta)$,
  $M:=(K+1)a_{\max}^2$ with $a_{\max}:=1+3D_T/h_{\min}$,
  $C_1'$ the universal uniform-deviation constant of the
  planar-distance class alone---unlike the constant $C_1$
  of (S3) of \cite{dmvsa2026supp}, which grows linearly with
  $\gamma_{\mathrm{enc}}$, it is free of
  $(\gamma_{\mathrm{enc}},D_T,h_{\min})$---and $C_1^e$ the
  universal relative-deviation constant of the enclosure
  class of Lemma~S5 of \cite{dmvsa2026supp},
  $\bar c_3:=\min\{c_3,\gamma_{\mathrm{vol}}c_{\mathrm g}/2\}$
  the effective gap constant, and $\epsilon$ the target
  resolution calibrating $b$ in \eqref{eq:surr-risk}. The
  budget thus controls the total variation \emph{beyond} the
  target resolution; the proof in fact establishes the
  activated form: whenever
  $\TV(\mathbb{P}_{\widehat{\mathcal{S}}},
  \mathbb{P}_{\mathcal{S}_T})\ge\epsilon$, the same display
  holds with the unsubtracted left-hand side
  $\bar c_3\,\TV(\mathbb{P}_{\widehat{\mathcal{S}}},
  \mathbb{P}_{\mathcal{S}_T})$.
  The conditioning on
  $\{\widehat{\mathcal{S}}\in\mathfrak{S}_T\}$ is
  essential and is \emph{not} certified by the theory for the
  output of a first-order solver: the oracle comparison
  controls only the combination
  $\gamma_{\mathrm{vol}}\log(\Vol(\widehat{\mathcal{S}})/V_T)
  +\gamma_{\mathrm{enc}}\mathrm{enc}_N(\widehat{\mathcal{S}})$
  of \eqref{eq:enc-bound}, constraining the volume of
  $\widehat{\mathcal{S}}$ but neither its diameter nor its
  position, so the event can fail for an approximate minimizer
  with objective value arbitrarily close to the oracle value.
  Membership of
  $\widehat{\mathcal{S}}$ in $\mathfrak{S}_T$ cannot be
  certified from the data alone, since the envelope is
  defined through the unknown target quantities
  $(V_T,D_T,h_{\min})$; evaluating the barycentric
  coordinates of the samples against
  $\widehat{\mathcal{S}}$ checks containment of the data,
  not membership in $\mathfrak{S}_T$.
  Proposition~\ref{prop:transfer}(ii)
  and Theorem~\ref{prop:endtoend} are therefore conditional
  guarantees---an error budget for an
  approximate minimizer, valid on the envelope event---and
  are stated as such throughout.
\end{theorem}
The terms of \eqref{eq:endtoend} are, in order, the
statistical fluctuation of the surrogate risk over the
simplex class---network-free by construction, as noted in
Remark~\ref{rem:decomp}, and free of the
enclosure weight through the variance-adaptive bound of
Lemma~S5 of \cite{dmvsa2026supp}---the approximation and optimization
terms entering linearly, and the enclosure residual of the
soft containment penalty, followed by the
two localized enclosure-deviation terms supplied by
Lemma~S5; the equality-slack terms,
proportional to $\lambda_{\mathrm{eq}}$, vanish identically in
the noiseless model. In particular, suppose that
\begin{equation}
  \label{eq:budget-cond}
  C_2\eta_{\mathrm{apx}}+\eta_{\mathrm{opt}}
  \;\le\;\frac{\bar c_3\,\epsilon}{8},
  \qquad
  C_2\eta_{\mathrm{apx}}+\eta_{\mathrm{opt}}
  +\gamma_{\mathrm{vol}}\log 4
  \;\le\;\frac{\gamma_{\mathrm{enc}}\,
  \bar c_3^{\,2}\,\epsilon^{2}}{16\,b^{2}c_4^{2}},
\end{equation}
hold. The second condition is stated in
expanded form to make the closure of the budget manifest:
substituting it into the square-root term of
\eqref{eq:endtoend} gives
\[
bc_4\Bigl(\frac{C_2\eta_{\mathrm{apx}}+\eta_{\mathrm{opt}}
+\gamma_{\mathrm{vol}}\log 4}{\gamma_{\mathrm{enc}}}
\Bigr)^{1/2}
\;\le\; bc_4\cdot\frac{\bar c_3\epsilon}{4bc_4}
\;=\;\frac{\bar c_3\epsilon}{4},
\]
with no residual $\sqrt{bc_4}$ factor. Then, on the envelope
event, the
planar statistical term is below $\bar
c_3\epsilon/4$ once
$N\ge64C_1'^2L_N/(\bar c_3^{\,2}\epsilon^2)$---an
$\epsilon^{-2}$ scale at fixed $K$, as in
Theorem~\ref{thm:main}---while the two localized
enclosure-deviation terms are below $\bar
c_3\epsilon/16$ each once
\begin{equation}
  \label{eq:budget-N}
  N\;\ge\;\max\Bigl\{
  \frac{1024\,\gamma_{\mathrm{enc}}C_1^eML_N\,
  E_{\mathrm{apx}}}{\bar c_3^{\,2}\epsilon^{2}},
  \;\frac{96\,\gamma_{\mathrm{enc}}C_1^eML_N}
  {\bar c_3\epsilon}\Bigr\},
\end{equation}
and the
linear term is at most
$\bar c_3\epsilon/8$, and the explicit part of the
square-root term is at most $\bar c_3\epsilon/4$,
so the right-hand side of \eqref{eq:endtoend} is below
$15\bar c_3\epsilon/16
<\bar c_3\epsilon$ up to the equality-slack residual
discussed next. Because
\eqref{eq:budget-cond} constrains only the weights and
errors while \eqref{eq:budget-N} constrains the sample
size, composing the two makes the $\gamma_{\mathrm{enc}}$
scaling explicit: at the minimal enclosure weight admitted
by the second condition of \eqref{eq:budget-cond},
$\gamma_{\mathrm{enc}}\asymp
b^2c_4^2\gamma_{\mathrm{vol}}\log4/
(\bar c_3^{\,2}\epsilon^{2})=\Theta(\epsilon^{-4})$ at
fixed $K$, the first branch of \eqref{eq:budget-N} dominates
and the end-to-end budget closes at an effective sample size
of order $\epsilon^{-6}$ at fixed $K$ (up to $L_N$ and
geometry constants)---polynomially above, though not of the
same order as, the $\epsilon^{-2}$ surrogate scale of
Theorem~\ref{thm:main}. The
remaining equality-slack residual
$b c_4\cdot O_p(\sqrt{\lambda_{\mathrm{eq}}
/(\gamma_{\mathrm{enc}}N)})$ is the one term not governed by
\eqref{eq:budget-cond}--\eqref{eq:budget-N}: it is $o(1)$ as $N$ grows, but at the
minimal $\epsilon^{-2}$ sample scale of
Theorem~\ref{thm:main} it is of order
$\sqrt{\lambda_{\mathrm{eq}}\big/
(\gamma_{\mathrm{enc}}\log(1/\epsilon))}$ rather than
$O(\epsilon)$ when $\lambda_{\mathrm{eq}}>0$. It vanishes
under the noiseless design choice $\lambda_{\mathrm{eq}}=0$,
and in
general it is below $\bar c_3\epsilon/16$ once
$N\gtrsim b^2c_4^2\lambda_{\mathrm{eq}}\big/
(\gamma_{\mathrm{enc}}\bar c_3^2\epsilon^2)$, an
$\epsilon^{-4}$ scale. Under either provision the right-hand
side of \eqref{eq:endtoend} is below $\bar c_3\epsilon$, so
its activated form forces
$\TV(\mathbb{P}_{\widehat{\mathcal{S}}},
\mathbb{P}_{\mathcal{S}_T})\le\epsilon$ with probability at
least $1-3\zeta-2/N$ on
$\{\widehat{\mathcal{S}}\in\mathfrak{S}_T\}$. The second
condition of \eqref{eq:budget-cond} is a weight condition of the
same type as Remark~\ref{rem:design}, its coarser exponent
inherited from the envelope bound
$[\log(V_T/\Vol(\mathcal{S}))]_+\le\log4$. Under
Assumption~\ref{ass:spatial} the first condition is met at the
capacity \eqref{eq:capacity} of Proposition~\ref{prop:apxrate}.
A final scope remark: the theorem analyzes the surrogate
\eqref{eq:surr-risk} alone; the additional components of the
deployed objective \eqref{eq:deep-risk} (the safeguard penalties
and the nonconvex gating branch) enter the budget only through
$\eta_{\mathrm{apx}}$ and $\eta_{\mathrm{opt}}$, and nothing
in \eqref{eq:endtoend} certifies their effect beyond those
residuals.

\begin{proposition}[Lower bounds and necessity of
the gap cap]
  \label{prop:lowerbound}
  We write
  $\mathbb{S}_K(\underline\lambda,\bar\lambda,D_T)$ for the
  class of simplices obeying Assumption~\ref{ass:iso} with
  constants $(\underline\lambda,\bar\lambda)$ and diameter at
  most $D_T$; suprema over $\mathcal{S}_T$ range over this
  class.
  \begin{enumerate}
    \item[(i)] \emph{Noisy companion model.} Let
      $\widetilde X_j$$=X_j+\varepsilon_j$ with
      $X_j\stackrel{\mathrm{i.i.d.}}{\sim}
      \mathbb{P}_{\mathcal{S}_T}$ and
      $\varepsilon_j\stackrel{\mathrm{i.i.d.}}{\sim}
      N(0,\sigma^2I_K)$, $\sigma>0$. If
      $N\ge K^2\sigma^2/(4D_T^2)$, then
      \[
        \inf_{\widehat{\mathcal{S}}}\sup_{\mathcal{S}_T}
        \mathbb{P}\bigl(
        \TV(\mathbb{P}_{\widehat{\mathcal{S}}},
        \mathbb{P}_{\mathcal{S}_T})\ge\epsilon_\sigma\bigr)
        \ge\frac{3}{8},
        \qquad
        \epsilon_\sigma:=\frac{K\sigma}{8D_T\sqrt N},
      \]
      the infimum ranging over all estimators of the
      simplex.
    \item[(ii)] \emph{Noiseless model.} Under
      Assumption~\ref{ass:uniform},
      \[
        \inf_{\widehat{\mathcal{S}}}\sup_{\mathcal{S}_T}
        \mathbb{P}\bigl(
        \TV(\mathbb{P}_{\widehat{\mathcal{S}}},
        \mathbb{P}_{\mathcal{S}_T})\ge\epsilon_0\bigr)
        \ge\frac{3}{8},
        \qquad
        \epsilon_0:=\frac{1}{64N}.
      \]
    \item[(iii)] \emph{Sharpness of the gap.} For every
      $s\in(0,2^{1/K}-1]$ the outward homothety
      $\mathcal{S}^s:=c_T+(1+s)(\mathcal{S}_T-c_T)$ about the
      centroid $c_T$ of $\mathcal{S}_T$ lies in the envelope
      $\mathfrak{S}_T$ and satisfies
      $\TV(\mathbb{P}_{\mathcal{S}^s},
      \mathbb{P}_{\mathcal{S}_T})=1-(1+s)^{-K}$ and
      \[
        R(\mathcal{S}^s)-R(\mathcal{S}_T)
        =\gamma_{\mathrm{vol}}K\log(1+s)
        \;\le\;2\gamma_{\mathrm{vol}}\,
        \TV\bigl(\mathbb{P}_{\mathcal{S}^s},
        \mathbb{P}_{\mathcal{S}_T}\bigr).
      \]
  \end{enumerate}
\end{proposition}
Three conclusions follow. First, in the noisy regime the
$N^{-1/2}$ dependence of Theorem~\ref{thm:main} is
unimprovable in its $N$-exponent: by
part~(i), at any fixed $\sigma>0$ every
estimator, of any
computational cost, incurs a total-variation error of order
$\sigma/\sqrt N$ up to the geometry factor $K/D_T$ as soon as
the observations carry arbitrarily small Gaussian
noise---slower than the $\sim 1/N$ noiseless
benchmark certified by part~(ii)---while
Theorem~\ref{thm:main} attains this exponent whenever the
noise level admits it ($\sigma=O(\epsilon^2/K)$). Part~(i)
certifies only this exponent: at fixed $\sigma$ the
noise-induced floor of Proposition~\ref{prop:noise} is bounded
away from zero while $\epsilon_\sigma\to0$, so the
noisy-model minimax rate as a joint function of $(N,\sigma)$
remains open.
Second, in the strict noiseless model the two-point argument
of part~(ii) certifies a $1/N$ lower bound: with uniform
sampling, information about a small homothetic perturbation
is carried by the boundary mass. Combined with
the maximum-likelihood upper bound, whose error decays as
$\widetilde O(K^2/N)$, part~(ii) sandwiches the noiseless
minimax error between $\Omega(1/N)$ and $\widetilde O(K^2/N)$:
its order in $N$ is thus $1/N$ up to factors polynomial in $K$
and logarithmic in $N$, strictly faster than $N^{-1/2}$.
This $N$-dependence is not new: Saberi et al.~\cite{saberi2025} establish the sharper lower bound $\Omega(K/N)$ in the noiseless model, of the same order in $N$ as part~(ii) and carrying the optimal factor of $K$; their analysis also gives the noisy lower bound $N\ge\Omega(K^3\sigma^2/\epsilon^2+K/\epsilon)$, which at fixed $\sigma$ is consistent with, though looser in $N$ than, the $\sigma/\sqrt{N}$ scaling of part~(i). Relative to \cite{saberi2025}, the contribution of part~(ii) is a fully explicit two-point homothetic construction with numerical constants and no factor of $K$, stated in the total-variation metric under the present uniform-sampling model; the same construction is what part~(iii) reuses to certify the sharp $\gamma_{\mathrm{vol}}$-dependence of the localization gap. Earlier minimax results for set estimation give related rates under different geometries and error metrics, such as the $\log n/n$ rate for polytopes with a fixed number of vertices \cite{brunel2013} and the minimax analysis of \cite{baldin2016}. What remains open is whether any
polynomial-time algorithm can beat the $N^{-1/2}$ benchmark in
the noiseless model, the information-theoretic rate $\Theta(K/N)$ being attained by the NP-hard maximum-likelihood estimator.
Third, part~(iii) shows that the
localization gap of Lemma~S4 of \cite{dmvsa2026supp} is linear in
$\epsilon$ with sharp $\gamma_{\mathrm{vol}}$-dependence: the
homothetic family attains a gap of
$\gamma_{\mathrm{vol}}(1+o(1))\,\epsilon$ as
$\epsilon\downarrow0$, so no localization constant exceeding
$\gamma_{\mathrm{vol}}(1+o(1))$ is valid in general, and the
$\gamma_{\mathrm{vol}}$-dependent cap in
$\bar c_3$ is necessary rather than
conservative; the residual factor
$2/c_{\mathrm g}=4$ between $\gamma_{\mathrm{vol}}(1+o(1))$
and $\gamma_{\mathrm{vol}}c_{\mathrm g}/2$ lies within the
slack of the slab argument of Lemma~S4 of \cite{dmvsa2026supp}. The
$\epsilon^{-2}$ sample scale of Theorem~\ref{thm:main} thus
reflects the linear geometry of the volume--enclosure
trade-off against the class fluctuation (S3) of \cite{dmvsa2026supp},
not looseness of the gap lemma.

\begin{theorem}[Computational complexity]
  \label{thm:complexity}
  Let $P$ denote the total number of learnable parameters
  ($P=O(K^2)$ for the $Q$-flow, the anchor, and the A-Net).
  Each stochastic gradient evaluation on a mini-batch of
  size $B$ requires $O(BK^2+K^3)$
  arithmetic operations, and each full data pass therefore
  requires $O(NK^2+NK^3/B)$, which is $O(NK^2)$ whenever
  $B=\Omega(K)$; the $O(K^3)$ term accounts for
  assembling the triangular factors of $Q$ once per iteration.
  The parameter and optimizer-state
memory---Adam maintains two moment vectors of the
same size as the parameter vector---is $O(K^2)$, plus
$O(BK)$ working memory for a mini-batch of size $B$; the $O(NK)$
input storage is common to every method that reads the
scene, ours included,
and can be streamed in batches. In the typical
  regime $K\ll N$, this is $O(NK^2)$ time per full pass and
  $O(K^2)$ parameter memory. By contrast, the Soft-ML implementation used here requires
$O(NK^3)$ time and $O(NK)$ memory---a count that is
implementation-specific, not intrinsic to the Soft-ML objective.
\end{theorem}
The proof is deferred to the supplement.
The argument is a direct operation count: each
full data pass is dominated by the $O(NK^2)$ matrix--vector
products in the reconstruction and enclosure terms, the $O(K^3)$
assembly of $Q$ from its triangular factors occurs only once per
iteration, and the parameter and optimizer-state memory is a
constant multiple of $P$.

Throughout, ``polynomial time'' qualifies the
per-sample arithmetic of the estimator class, as opposed to the
$O(NK^3)$ per-pass cost of the Soft-ML class of
\cite{najafi2021aos} (Remark~\ref{rem:softml}); it does not
qualify the nonconvex training of \eqref{eq:deep-risk}, for
which no polynomial-time convergence guarantee is claimed
(Section~S1.9 of \cite{dmvsa2026supp}).

\section{Experiments}
\label{sec:exp}

All experiments were conducted on a standard desktop PC
(Intel Core i7-10700 CPU, 32 GB RAM; no distributed or cloud
resources were used); in particular, DeepMVSA
itself, including the training of its coordinate network, runs
CPU-only in MATLAB R2022b.

The experimental design follows the structure of the theory.
Section~\ref{subsec:exp-a} examines the scaling of the
estimation error with $N$ and $K$ on synthetic data with known
ground truth; because this design uses index coordinates, it
operates in the floor-limited regime of
Remark~\ref{rem:apxfloor}, probing the surrogate-level
statistical error, not the approximation-transfer chain of
Proposition~\ref{prop:transfer}. Section~\ref{subsec:exp-bio}
checks the estimator on the biological benchmark of
\cite{shenorr2010,najafi2021aos}---a small-$N$ cross-domain
applicability check that does not exercise the large-$N$
scalability mechanism. Sections~\ref{subsec:exp-c}
and~\ref{subsec:exp-d} assess practical relevance against a
reference library and computational feasibility on real scenes
of increasing size, up to $N\approx 10^7$ pixels.

No existing coordinate/implicit neural simplex estimator
enforces a minimum-volume penalty, so no technically meaningful
implicit baseline is available; the comparisons below establish
computational advantages over classical minimum-volume
formulations, not universal superiority over neural unmixing
methods.

Baseline configurations follow the original papers
\cite{li2015tgrs,najafi2021aos}, with a uniform one-hour
wall-clock budget; implementation and reproducibility details
are collected in the supplement \cite{dmvsa2026supp}.

\subsection{Synthetic simplices: scaling with $N$ and $K$}
\label{subsec:exp-a}

\begin{figure}[tbp]
  \centering
  \begin{subfigure}[b]{0.305\linewidth}
    \centering
    \includegraphics[width=\linewidth]{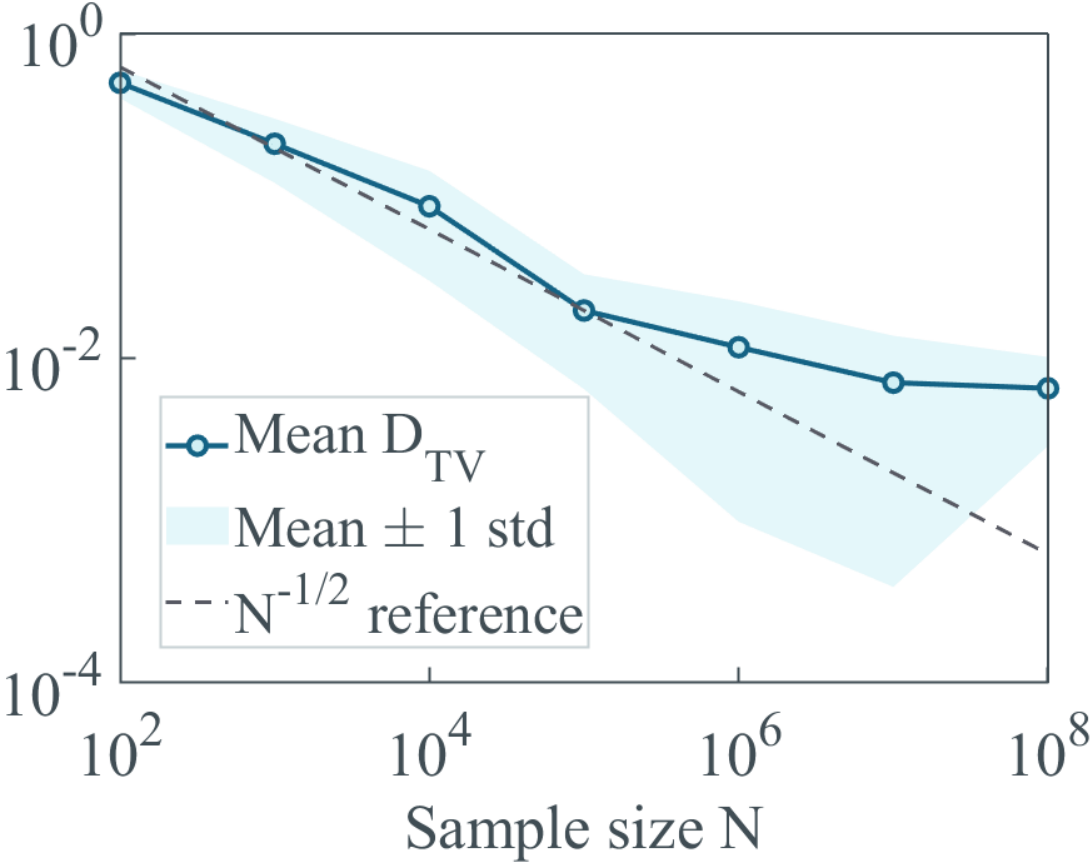}
    \caption{}
    \label{K3}
  \end{subfigure}
  \hfill
  \begin{subfigure}[b]{0.305\linewidth}
    \centering
    \includegraphics[width=\linewidth]{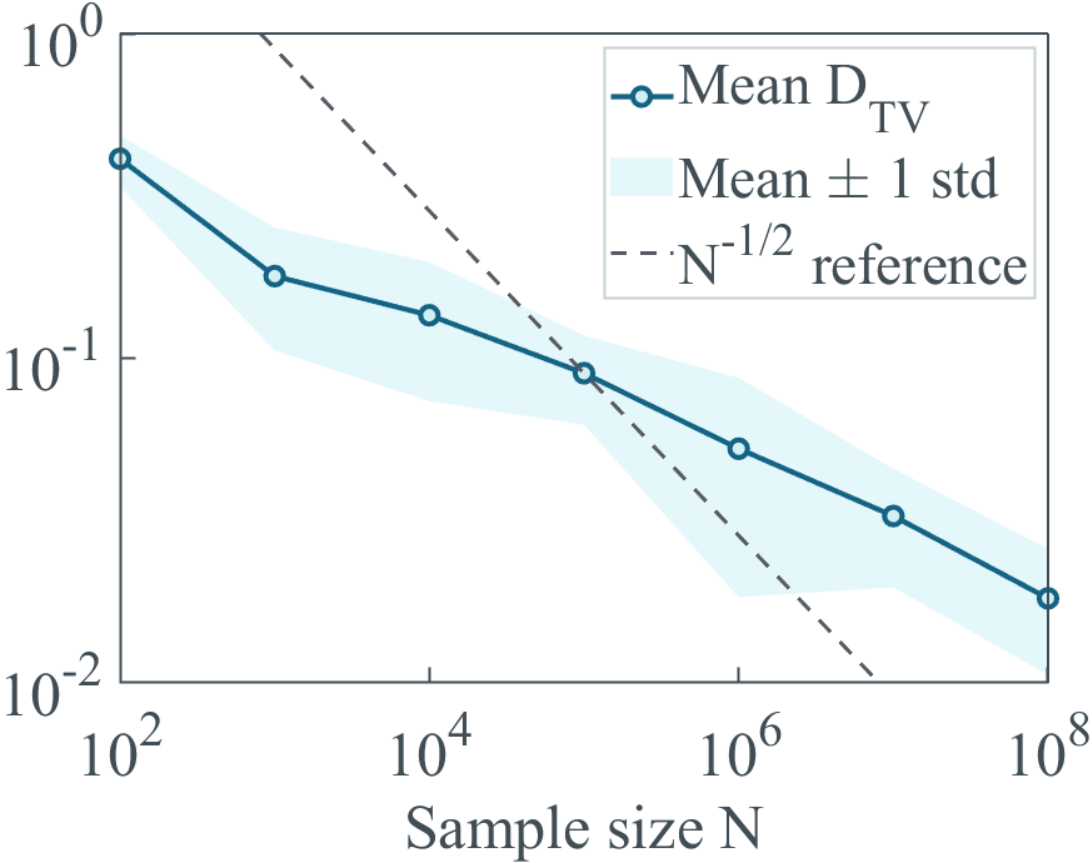}
    \caption{}
    \label{K5}
  \end{subfigure}
  \hfill
  \begin{subfigure}[b]{0.305\linewidth}
    \centering
    \includegraphics[width=\linewidth]{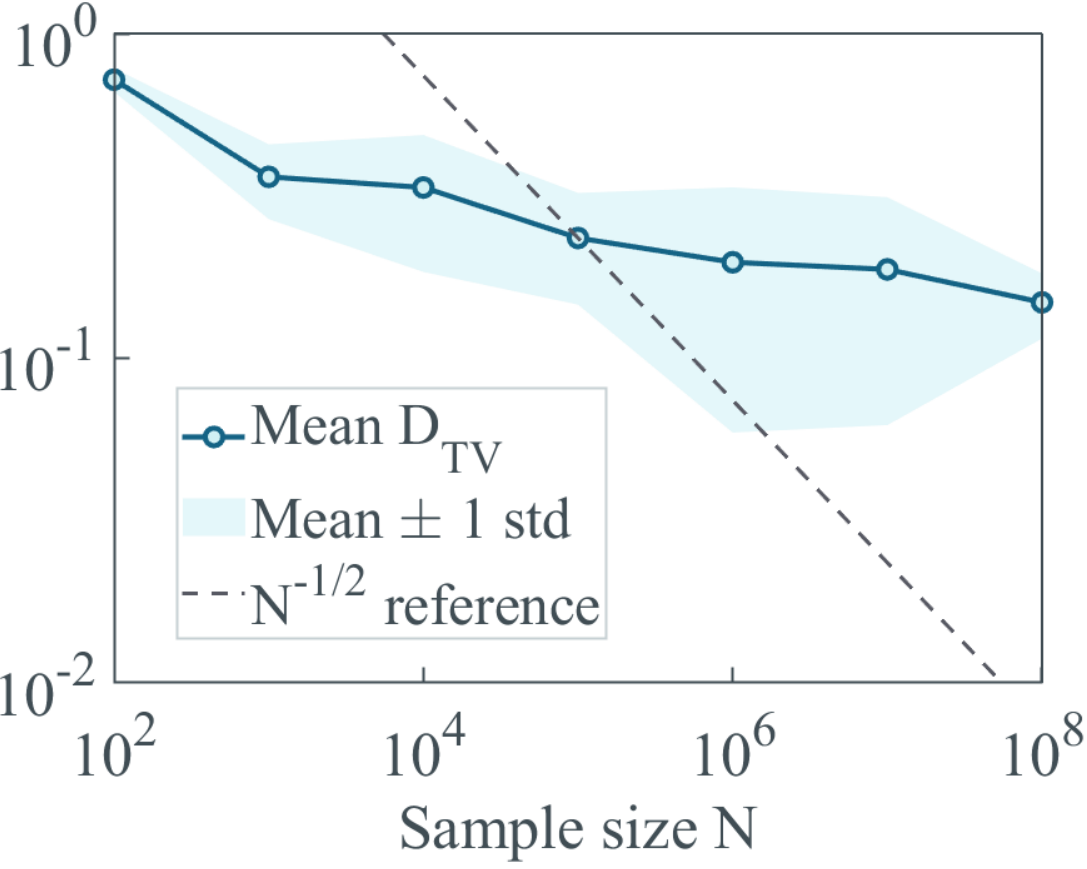}
    \caption{}
    \label{K7}
  \end{subfigure}

  \caption{Total-variation distance
$\TV(\mathbb{P}_{\mathcal{S}_T},\mathbb{P}_{\widehat{\mathcal{S}}})$
versus sample size $N$ on synthetic simplices for $K=3,5,7$
(log--log axes). Solid curves: mean over 20 independent
replications; shaded bands: $\pm1$ standard deviation. Dashed
line: the $N^{-1/2}$ reference rate of Theorem~\ref{thm:main}.}
  \label{D_TV}
\end{figure}

Following Section~4.2 of \cite{najafi2021aos},
we generate $N$ points uniformly from a $K$-simplex, varying
$N$ and $K$ to examine the error rate of
Theorem~\ref{thm:main} and the time--memory scaling of
Theorem~\ref{thm:complexity}; to probe the extreme end of the
linear-scaling regime, we additionally run DeepMVSA with up to
$N=10^8$ observations. The scaling and consistency
experiments of Table~\ref{tab:scaling} and Figure~\ref{D_TV}
use noiseless data, $\rho=0$ in the normalized scale of
\cite{najafi2021aos}; the accuracy comparison of
Table~\ref{tab:mse_sad} is conducted under additive Gaussian
noise. The grid covers moderate $K$ (beyond the $K=2$ of
\cite{najafi2021aos}) and $\mathrm{SNR}\approx20$--$30$~dB, with
the noiseless and large-$N$ cells probing the theoretical boundary. Two baselines are compared: MVSA
\cite{li2015tgrs}, which minimizes the simplex volume under
hard enclosure constraints, and Soft-ML \cite{najafi2021aos}.

Table~\ref{tab:scaling} reports runtime and
peak memory for varying $N$ (upper panel, $K=2$) and varying
$K$ (lower panel, $N=10^6$), corroborating the two predictions
of Theorem~\ref{thm:complexity}. DeepMVSA's peak memory is
essentially independent of $N$, staying within $13.3$~MB as $N$
grows from $10^4$ to $10^8$, whereas the baselines' footprints
grow linearly ($34$--$45$-fold above DeepMVSA's
$13.3$~MB at the same $N$), both exceeding the one-hour
budget at $N=10^8$.
DeepMVSA's runtime carries a fixed training overhead of about
$8$~s, amortized once $N\ge10^7$: at $N=10^7$ it outperforms
MVSA by more than an order of magnitude, and at $N=10^8$ it is
about $5.5$ times faster than Soft-ML with $450$ times less
memory; the mild runtime increase at $N=10^8$ reflects the
$O(NK)$ streaming cost, not growth of the model state.
A scope note: with the fixed iteration count
(supplement), the $N=10^8$ run draws about $6\times10^{6}$
sample presentations, roughly $6\%$ of the data, certifying
throughput and memory feasibility rather than statistical
convergence at $N=10^8$, for which the iteration budget
would have to grow with $N$. In the lower panel, DeepMVSA grows only moderately
with $K$ ($10.50$~s, $20.65$~MB at $K=20$); Soft-ML exceeds
the one-hour budget already at $K=10$. The wall-clock
gap exceeds the operation-count prediction, the Soft-ML
bottleneck being per-sample pseudo-inverse loop overhead rather
than its asymptotic $O(NK^3)$ cost; all comparisons use the published MVSA code and our
reimplementation of Soft-ML, with baselines run to convergence
or the one-hour limit---a protocol favoring the baselines at
small $N$.

\begin{table}[tbp]
  \caption{Runtime and peak memory on
  synthetic data, averaged over $20$ independent replications;
  each cell reports runtime~/~peak memory. Upper panel:
  varying $N$ at fixed $K=2$; lower panel: varying $K$ at
  fixed $N=10^6$. ``$>$1 hour'' marks censored runs exceeding
  the wall-clock budget (no runtime imputed). Memory figures
  report peak workspace storage (MATLAB \texttt{whos}/memory
  profiling), not resident-set size.}
  \label{tab:scaling}
  \footnotesize
  \centering
  \begin{tabular}{lccc}
    \toprule
 & Soft-ML \cite{najafi2021aos} & MVSA \cite{li2015tgrs} & DeepMVSA \\
        \midrule
    \multicolumn{4}{c}{Fixed $K=2$, varying $N$} \\
    \midrule
    $N=10^4$ & 0.10 s / 0.61 MB & 0.13 s / 0.47 MB & 7.91 s / 13.31 MB \\
    $N=10^5$ & 0.16 s / 6.10 MB & 1.23 s / 4.59 MB & 7.87 s / 13.28 MB \\
    $N=10^6$ & 1.14 s / 61.03 MB & 4.09 s / 45.79 MB & 8.14 s / 13.30 MB \\
    $N=10^7$ & 11.07 s / 0.59 GB & 119.60 s / 0.45 GB & 8.56 s / 13.28 MB \\
    $N=10^{8}$ & 196.77 s / 5.96 GB & $>$1 hour & 35.54 s / 13.31 MB \\
    \midrule
    \multicolumn{4}{c}{Fixed $N=10^6$, varying $K$} \\
    \midrule
    $K=5$ & 1.57 s / 0.10 GB & 19.14 s / 91.57 MB & 8.61 s / 14.53 MB \\
    $K=10$ & $>$1 hour & 61.77 s / 0.16 GB & 9.16 s / 16.57 MB \\
    $K=20$ & $>$1 hour & 136.91 s / 0.31 GB & 10.50 s / 20.65 MB \\
    \bottomrule
  \end{tabular}
\end{table}

Figure~\ref{D_TV} displays the
total-variation distance
of Theorem~\ref{thm:main}, estimated by Monte Carlo integration
against fresh draws from both simplices (common point count
across cells, recorded in the released code),
against $N$ on log--log axes for $K=3,5,7$. The error decays
monotonically, at a polynomial rate close to the $N^{-1/2}$
reference line for $K=3$ over the intermediate range; the $K=5$
and $K=7$ panels decay more slowly, consistent with the growth of
the localization and deviation constants with $K$ rather than with
the $N$-exponent itself. At the largest sample
sizes the decay slows and, for $K=3$, flattens into a plateau
slightly below $10^{-2}$; this flattening is computational
rather than statistical---the training budget is fixed across
$N$ (at $N=10^8$, $1500$ iterations of batch size $4096$ cover
about $6\%$ of the data), while Theorem~\ref{thm:main} bounds
the empirical risk minimizer, not a fixed-budget output. At $N=10^4$
(the configuration of Table~\ref{tab:mse_sad}), the noiseless
errors lie below the corresponding noisy entries at both SNR
levels, quantifying the degradation due to noise.
With index coordinates these experiments operate in the
floor-limited regime of Remark~\ref{rem:apxfloor}
(Section~\ref{sec:exp}); Figure~\ref{D_TV} is accordingly
evidence for the soft-enclosure mechanism, to which the
coordinate network supplies a differentiable parameterization.
The wider variability bands at isolated large $N$ reflect
growing relative Monte Carlo error as the true distance
approaches zero, together with a few suboptimal local
solutions, without reversing the decay of the mean.

Table~\ref{tab:mse_sad} evaluates the
estimation accuracy at $N=10^4$ for $K\in\{3,5,7\}$, under
additive white Gaussian noise at $\mathrm{SNR}=30$
and $20$~dB
($\mathrm{SNR}:=10\log_{10}(\|Y\|_F^2/\|E\|_F^2)$, where
$Y$ is the generated noiseless data matrix in the normalized
scale of \cite{najafi2021aos} and $E$ has i.i.d.\ Gaussian
entries whose variance is fixed by this ratio, harsher
than the $50$--$70$~dB regimes of \cite{li2015tgrs}), in terms of
the total-variation distance,
the vertex error \cite{najafi2021aos}, and the abundance
RMSE. At both noise levels DeepMVSA attains the
smallest total-variation distance and vertex error in all six
configurations, its advantage widening with $K$ (at
$\mathrm{SNR}=30$~dB, $K=7$: $\TV=0.64$ against $0.96$ and
$0.99$; vertex error $0.29$ against $0.77$ and $0.89$).
Baseline failure is systematic---Soft-ML already at
$K=5$ at both noise levels ($\TV=0.98$--$0.99$), both baselines
at $K=7$---whereas DeepMVSA stays bounded away from one in all
six configurations (largest value $0.90$).
The volume control of
Proposition~\ref{prop:transfer}(i) precludes the inflation
mode for DeepMVSA, consistent with its vertex errors remaining
below one in every configuration. For the abundance RMSE, MVSA is lowest
at $\mathrm{SNR}=30$~dB, closely followed by DeepMVSA; this does
not indicate superior recovery, since MVSA's failure mode is an
overly large simplex that still encloses the data, on which
constrained least squares refits the abundances almost exactly.
We therefore report total-variation distance and vertex error as
the primary metrics, deliberately \emph{not} the reconstruction
error, which an overly large simplex drives near zero with
vertices arbitrarily far from the truth. Taken together,
Table~\ref{tab:scaling}, Figure~\ref{D_TV}, and
Table~\ref{tab:mse_sad} establish the claims of this
subsection: scalability to $N=10^8$ within constant memory,
error decay with $N$ as predicted by Theorem~\ref{thm:main},
and the highest accuracy under noise, most pronouncedly where
both baselines fail to localize the simplex.

\begin{table}[tbp]
    \footnotesize
  \centering
    \caption{Estimation accuracy on
    synthetic scenes at $N=10^4$ for $K\in\{3,5,7\}$, under
    additive white Gaussian noise at $\mathrm{SNR}=30$ and
    $20$ dB; means over 20 independent replications (standard
    deviations in parentheses). A total-variation distance
    near one indicates failure to localize the
    simplex.
    Entries marked with a dagger $\dagger$
    report total-variation distances
    of $0.96$ or above under the adopted
    display criterion; the threshold $0.96$ is a
    visualization convention for flagging such entries, not
    a statistical criterion.
    Best results in bold.
    A standard deviation displayed as $0.00$
    rounds a value below $0.005$ (the failed runs concentrate
    at the boundary of the simplex class).
    The MVSA abundance RMSE is not directly
    comparable: MVSA's high-dimensional failure mode is an
    overly large simplex that still encloses the data, on
    which the abundance RMSE stays small even when the
    vertices are not recovered; the total-variation distance
    and the vertex error are the comparable metrics
    (Section~\ref{subsec:exp-a}).}
    \label{tab:mse_sad}
    \setlength{\tabcolsep}{1.5pt}
    \begin{tabular}{llcccccc}
        \toprule
        & & \multicolumn{3}{c}{$\mathrm{SNR}=20$ dB}
          & \multicolumn{3}{c}{$\mathrm{SNR}=30$ dB} \\
        \cmidrule(lr){3-5} \cmidrule(lr){6-8}
        $K$ & Metric & MVSA & Soft-ML & DeepMVSA & MVSA & Soft-ML & DeepMVSA \\
        \midrule
        \multirow{3}{*}{3} & $\TV$ distance
          & 0.65(0.16) & 0.86(0.26) & \textbf{0.36(0.19)}
          & 0.34(0.12) & 0.71(0.42) & \textbf{0.19(0.10)} \\
          & Vertex error
          & 0.67(0.27) & 1.01(0.48) & \textbf{0.26(0.11)}
          & 0.28(0.17) & 1.01(0.67) & \textbf{0.20(0.14)} \\
          & Abundance RMSE
          & 0.07(0.02) & 0.34(0.15) & \textbf{0.05(0.02)}
          & \textbf{0.03(0.02)} & 0.30(0.21) & 0.04(0.01) \\
        \midrule
        \multirow{3}{*}{5} & $\TV$ distance
          & 0.96(0.03)$^{\dagger}$ & 0.99(0.01)$^{\dagger}$ & \textbf{0.74(0.13)}
          & 0.76(0.12) & 0.98(0.01)$^{\dagger}$ & \textbf{0.43(0.13)} \\
          & Vertex error
          & 1.19(0.38) & 0.95(0.21) & \textbf{0.45(0.26)}
          & 0.53(0.47) & 1.01(0.30) & \textbf{0.25(0.10)} \\
          & Abundance RMSE
          & 0.09(0.01) & 0.26(0.06) & \textbf{0.08(0.03)}
          & \textbf{0.04(0.01)} & 0.27(0.07) & 0.04(0.02) \\
        \midrule
        \multirow{3}{*}{7} & $\TV$ distance
          & 0.99(0.00)$^{\dagger}$ & 0.99(0.00)$^{\dagger}$ & \textbf{0.90(0.06)}
          & 0.96(0.04)$^{\dagger}$ & 0.99(0.00)$^{\dagger}$ & \textbf{0.64(0.13)} \\
          & Vertex error
          & 1.50(0.25) & 0.91(0.24) & \textbf{0.53(0.19)}
          & 0.77(0.31) & 0.89(0.11) & \textbf{0.29(0.13)} \\
          & Abundance RMSE
          & \textbf{0.08(0.01)} & 0.24(0.04) & 0.11(0.01)
          & \textbf{0.05(0.01)} & 0.19(0.05) & 0.05(0.02) \\
        \bottomrule
    \end{tabular}
\end{table}

\subsection{Computational biology: cell-type deconvolution}
\label{subsec:exp-bio}

The purpose of this experiment is to assess accuracy and
stability on a real biological deconvolution task. Following
the biological deconvolution protocol of \cite{shenorr2010,najafi2021aos},
we deconvolve cell-type-specific gene-expression profiles from
bulk tissue samples. The dataset consists of $N=42$ micro-array
samples in which $K+1=3$ tissues (brain, liver, and kidney) were
mixed in known proportions \cite{shenorr2010}. The initial
gene-expression dimension is $K_{\mathrm{init}}=31\,100$; because
$N\le K_{\mathrm{init}}$, the data are linearly projected onto
the leading $41$ principal components via PCA, an
exact reduction here since $42$ centered
samples span a subspace of dimension at most $41$.

We compare DeepMVSA with Soft-ML \cite{najafi2021aos}, the
convex-hull-based initialization strategy analyzed in
Section~\ref{sec:prelim}. Accuracy is
assessed by the vertex error against the ground-truth tissue
profiles; following \cite{najafi2021aos}, the Pearson correlation
and mean squared error between the estimated and real mixture
coefficients are reported as benchmark-specific metrics.
All reported metrics are averaged over $50$
independent replications of each algorithm, with standard
deviations reported alongside. Since the $N=42$
samples are fixed, these replications measure optimization
variability (random initialization and mini-batch order), not
statistical sampling variability; the latter is assessed in
Section~\ref{subsec:exp-a}.

\begin{figure}[tbp]
  \centering
  \begin{subfigure}[b]{0.45\linewidth}
    \centering
    \includegraphics[height=4.05cm]{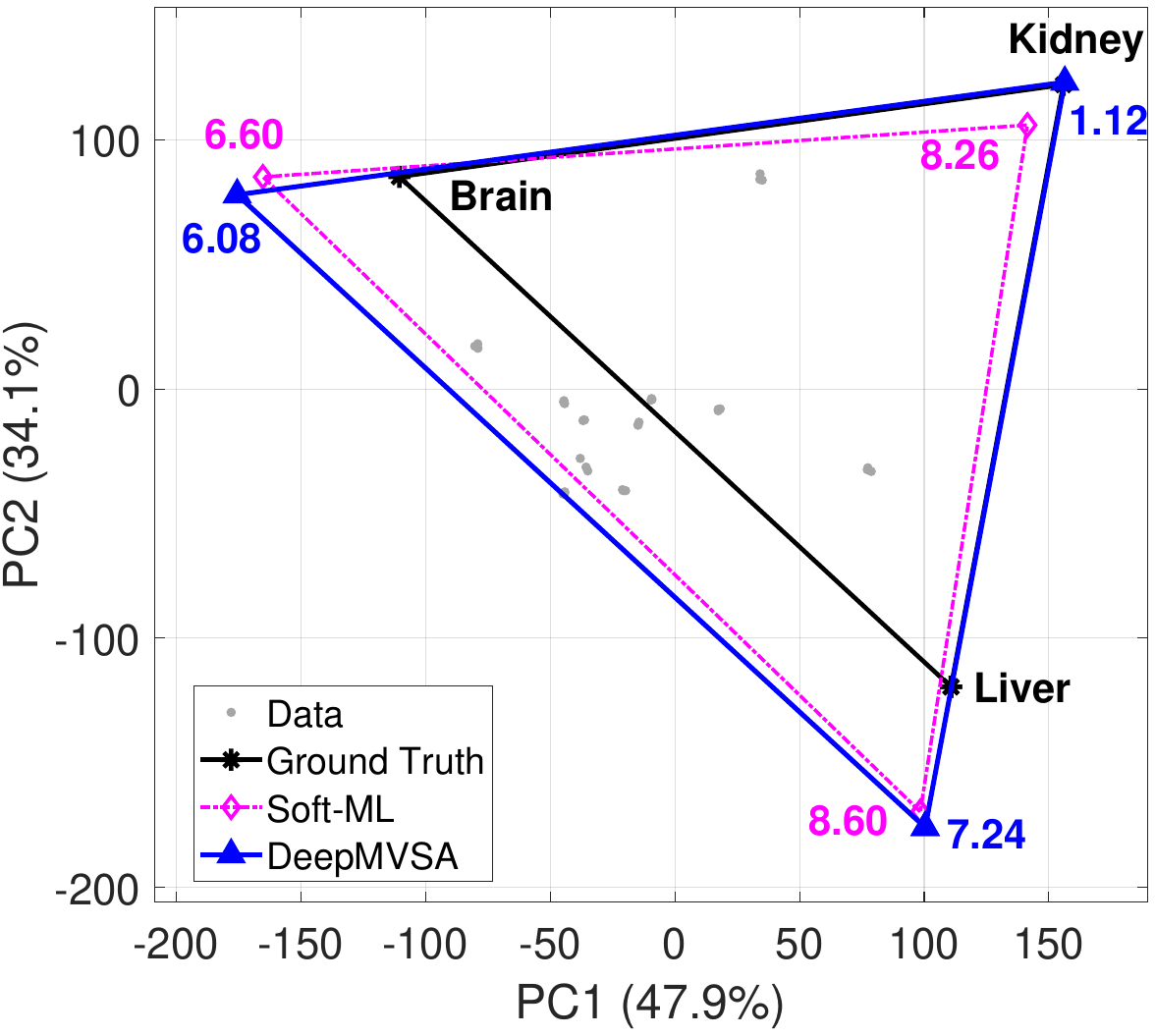}
    \caption{Projection of the micro-array data onto the first two PCs, which explain $47.9\%$ and $34.1\%$ of the total variance.}
    \label{fig:bio-a}
  \end{subfigure}
  \hfill
  \begin{subfigure}[b]{0.40\linewidth}
    \centering
    \includegraphics[height=4.05cm]{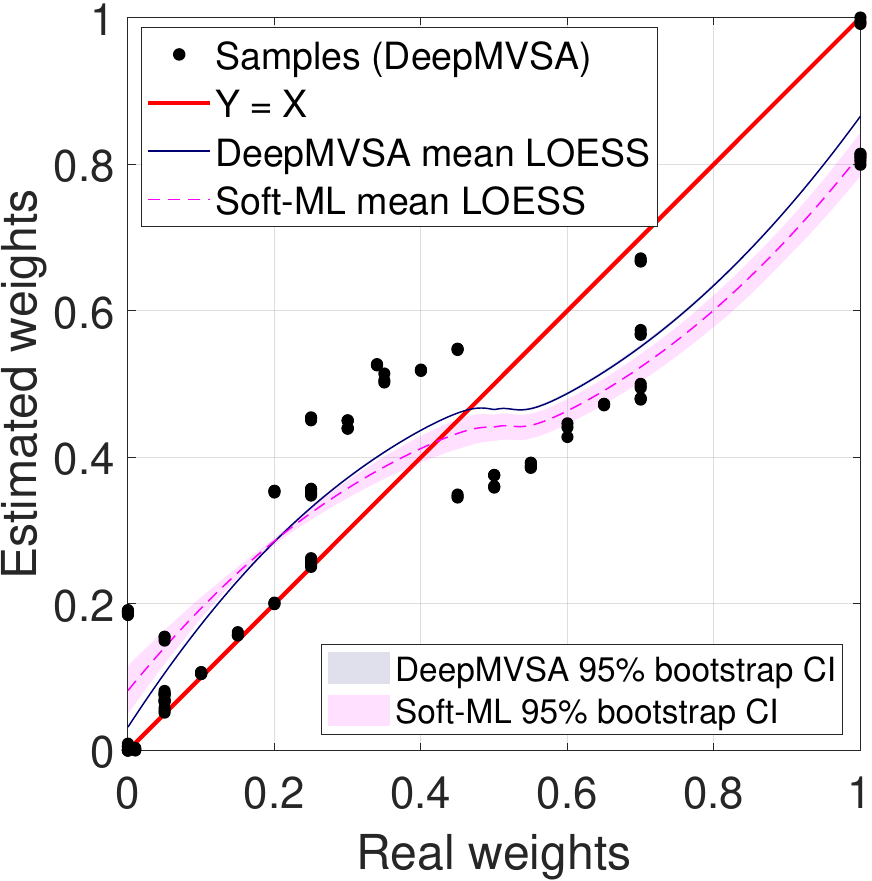}
    \caption{Scatter plot of inferred mixture coefficients against the ground-truth proportions.}
    \label{fig:bio-b}
  \end{subfigure}

  \caption{Cell-type identification from micro-array data given in \cite{shenorr2010}.
  (a) Projection onto the first two PCs, with the ground-truth
  tissue simplex and the Soft-ML and DeepMVSA estimates;
  vertex labels report the vertex error (the
  Euclidean distance between the estimated and the ground-truth
  tissue vertex, in the measurement units of the gene-expression
  data) of each method, averaged over $50$ independent
  replications.
  (b) Inferred mixture coefficients against ground-truth
  proportions, with the identity reference line, the mean
  LOESS trends, and 95\% bootstrap confidence bands.}
  \label{fig:bio}
\end{figure}

In Figure~\ref{fig:bio-a},
the simplex estimated by DeepMVSA closely tracks the
ground-truth tissue simplex in the first two principal-component
directions: all three tissue vertices are located
more accurately than Soft-ML's ($6.08$, $7.24$, $1.12$ versus
$6.60$, $8.60$, $8.26$ for brain, liver, kidney), the mean vertex
error reduced from $7.82$ to $4.81$, approximately $38\%$. The DeepMVSA estimates are also markedly more stable across
replications: the standard deviations of the vertex errors are
$1.18$, $1.34$ and $0.005$ for the liver, brain and kidney
vertices, against $6.41$, $4.29$ and $16.99$ for Soft-ML,
whose spread reflects occasional severe vertex misplacements.
Figure~\ref{fig:bio-b} compares the inferred
mixture weights with the known mixture proportions: DeepMVSA attains Pearson
correlation $0.912(0.011)$ and mean squared error
$0.015(0.001)$, clearly improving on Soft-ML ($0.797(0.282)$
and $0.033(0.043)$, approximately $54\%$ lower); since the
Soft-ML mean is inflated by occasional failed replications, this
is primarily a stability gain; the per-replication values are
released with the code.
The mean LOESS trends of both methods track the point
cloud closely, DeepMVSA's closer to the identity line, with a
barely discernible $95\%$ bootstrap confidence band; both
attenuate toward the center of the simplex (small proportions
overestimated, large underestimated), a bias typical of
minimum-volume estimates under measurement noise, visibly milder
for DeepMVSA. These results
indicate optimization stability and applicability
at small $N$; as noted in Section~\ref{sec:exp}, this experiment
does not exercise the large-$N$ scalability mechanism.

\subsection{Real hyperspectral data with a reference library: AVIRIS Cuprite}
\label{subsec:exp-c}

This experiment validates DeepMVSA on a real
hyperspectral scene collected by NASA's Jet Propulsion Laboratory against a reference library. The
benchmark is the full $250\times 191$ Airborne Visible Infra-Red Imaging Spectrometer (AVIRIS) Cuprite subscene
($N=47\,500$ pixels, $183$ spectral bands after removing
water-absorption channels); in contrast to the $10\,000$-pixel
subset used in \cite{najafi2021aos}, we process the full
subscene. Following \cite{li2015tgrs}, the data are projected onto a
$K=13$-dimensional subspace, so that the model has $K+1=14$ vertices.

We compare DeepMVSA with MVSA \cite{li2015tgrs} and Soft-ML
\cite{najafi2021aos} on runtime, peak memory, vertex matching
error, and reconstruction RMSE. The first two are as in
Section~\ref{subsec:exp-a}. Since no ground-truth vertices exist
for this scene, each estimated vertex $\widehat m_k$ is matched
against the USGS spectral library \cite{clark2007usgs}: the
\emph{vertex matching error} is $\mathrm{VME}:=(K+1)^{-1}\sum_{k=1}^{K+1}
\min_{m\in\mathcal{R}}\|\widehat m_k-m\|_2/\|m\|_2$,
with $\mathcal{R}$ the library's reference spectra. The
reconstruction RMSE is reported as a goodness-of-fit measure only,
being insensitive to vertex displacement
(Section~\ref{subsec:exp-a}).

Table~\ref{tab:cuprite} reports the results
(means over $20$ independent replications).
The number of recovered vertices is the primary
indicator on this benchmark: Soft-ML reliably estimates only
five of the fourteen vertices---the inward-collapse failure
mode of the saturating loss identified in
Remark~\ref{rem:softml}, observed here on real data---and its
vertex matching error of $0.16$ is computed conditionally on
those five, as the table note states, while the DeepMVSA and
MVSA figures average over all fourteen vertices.
With this conditioning made explicit, DeepMVSA attains
the smallest vertex matching error ($0.14$, against $0.18$ for
MVSA and $0.16$ for Soft-ML) and the smallest reconstruction
RMSE ($0.01$), while completing the full subscene in $9.10$~s
with $17.76$~MB of peak memory. At this moderate sample size
Soft-ML is the fastest and lightest of the three methods, as
expected from Table~\ref{tab:scaling}. The
VME comparison is asymmetric---conditional on successful
recovery, five vertices against fourteen---does not reflect
Soft-ML's failure to return nine of the fourteen vertices, and
should not be read as an accuracy advantage for Soft-ML; the
recovered-vertex count---fourteen for MVSA and DeepMVSA, five
for Soft-ML---is therefore the primary indicator for this scene.

\begin{table}[tbp]
  \caption{Comparison on the AVIRIS Cuprite scene
  (full $47\,500$-pixel subscene, $K+1=14$ vertices):
  runtime, peak memory, vertex matching error (VME)
  against the USGS spectral library, and reconstruction RMSE;
  means over $20$ independent replications (standard
  deviations in parentheses; peak memory is deterministic
  across replications and reported without one).
  Soft-ML is limited to reliably estimating only five vertices.
  Best results in bold.
  The Soft-ML VME is computed for its five
  reliable vertices and matched to the nearest library spectra;
  it is therefore conditioned on the subset of vertices the
  method successfully returns, rather than on all $14$
  requested vertices, and is not directly comparable to the
  full-set errors of the other two columns.}
  \label{tab:cuprite}
  \footnotesize
  \centering
  \begin{tabular}{lccc}
    \toprule
    Metric
    & MVSA \cite{li2015tgrs}
    & Soft-ML \cite{najafi2021aos}
    & DeepMVSA \\
    \midrule
    Runtime
    & 21.25(5.41) s & \textbf{0.82(0.01) s} & 9.10(0.16) s \\
    Peak memory
    & 10.54 MB & \textbf{4.37} MB & 17.76 MB \\
    Vertex matching error (VME)
    & 0.18(0.01) & 0.16(0.03) & \textbf{0.14(0.02)} \\
    Reconstruction RMSE
    & 0.06(0.001) & 0.02(0.008) & \textbf{0.01(0.001)} \\
    \bottomrule
  \end{tabular}
\end{table}

\subsection{Large-scale real hyperspectral data without a reference library}
\label{subsec:exp-d}

We process a Geology-1
Hyperspectral Satellite scene of $N=9\times 10^{6}$ pixels
($3000\times 3000$) and $L=16$ spectral
bands---nearly two hundred times the Cuprite
scene of Section~\ref{subsec:exp-c}. Following the HySime estimate
\cite{bioucas2008}, the data are projected onto a
$K=2$-dimensional subspace ($K+1=3$ vertices), so the cloud and
fitted simplex display exactly in the plane.
Evaluation is qualitative and computational: the tier
certifies that the $O(NK^2)$ time and $O(K^2)$ memory of
Theorem~\ref{thm:complexity} yield a completed, stable run at
$N\approx10^7$---a feasibility demonstration, not accuracy
evidence. Competing estimators maintain per-pixel variables growing
linearly in $N$ or require $N\times N$ attention matrices
infeasible at $N\approx 10^{7}$
\cite{hong2022egunet,ghosh2022deeptrans}; we therefore report
DeepMVSA only.

\begin{figure}[tbp]
    \centering
    \begin{subfigure}[b]{0.32\linewidth}
        \centering
        \includegraphics[height=2.55cm]{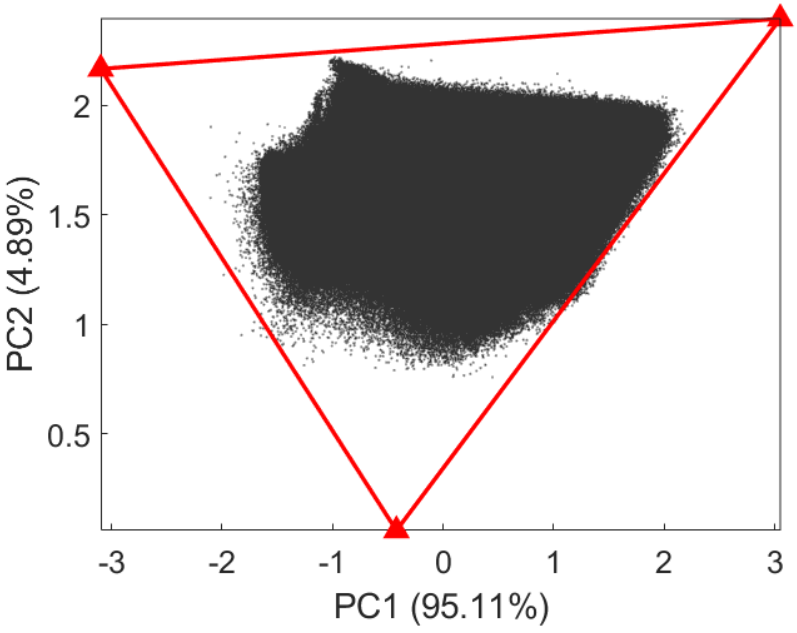}
        \caption{}
        \label{la}
    \end{subfigure}
    \hfill
    \begin{subfigure}[b]{0.32\linewidth}
        \centering
        \includegraphics[height=2.55cm]{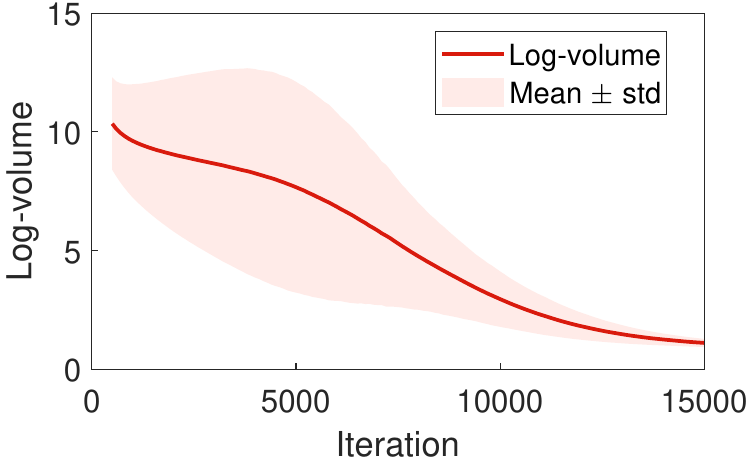}
        \caption{}
        \label{lb}
    \end{subfigure}
    \hfill
    \begin{subfigure}[b]{0.32\linewidth}
        \centering
        \includegraphics[height=2.55cm]{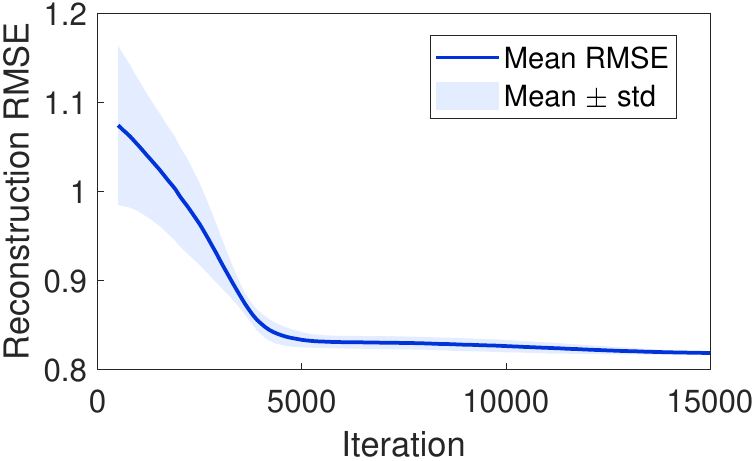}
        \caption{}
        \label{lc}
    \end{subfigure}
    \caption{Estimation results for the Geology-1 scene,
    averaged over $20$
    independent replications. (a) Projected data cloud (first
    two PCs, $95.11\%$ and $4.89\%$ of variance) and the
    simplex of the mean estimated vertices. (b)
    Log-volume and (c) reconstruction RMSE versus training
    iteration (mean $\pm1$ standard deviation).}
    \label{large}
\end{figure}

Figure~\ref{large} summarizes the geometry and training
dynamics, averaged over $20$ independent
replications. The log-volume
(Figure~\ref{lb}) decreases rapidly during the first
$5{,}000$ iterations and slowly
thereafter; the reconstruction RMSE (Figure~\ref{lc}),
a goodness-of-fit measure rather than an accuracy metric,
plateaus at $\approx 0.83$.
The residual volume decay reflects progressive
tightening of the facets against the extremal pixels: under
the soft-enclosure objective, residual slack along a facet can
always be traded for a small log-volume gain at negligible
reconstruction cost. The mean estimated simplex (Figure~\ref{la})
encloses essentially the entire projected cloud, its vertices
near the extremal directions of the pixel distribution---the
signature of a minimum-volume enclosing simplex. DeepMVSA
completes the full scene in
$86.59$~s with $14.12$~MB peak memory---below the
$17.76$~MB for Cuprite despite nearly two hundred times as many
pixels---directly illustrating the $N$-independent $O(K^2)$
memory of Theorem~\ref{thm:complexity}.

\section{Conclusion}
\label{sec:conclusion}

We have introduced DeepMVSA, a neural implicit estimator
re-ex\-press\-ing the minimum-volume principle in a scalable form,
with trainable-state memory $O(K^2)$, independent of $N$, and cost
$O(NK^2)$ per data pass. The surrogate estimator retains the best
sample complexity known to be attainable in polynomial time, and the
analysis is rounded out by a conditional end-to-end error budget
(Theorem~\ref{prop:endtoend}), a quantitative approximation
rate with an $N$-independent capacity prescription
(Proposition~\ref{prop:apxrate}), and two-point lower bounds
certifying the noisy rate exponent and the necessity of the
$\gamma_{\mathrm{vol}}$-depen\-dent gap cap
(Proposition~\ref{prop:lowerbound}). Experiments, including a
$10^7$-pixel satellite scene, are consistent with the
predicted accuracy and scaling; the largest runs probe the
computational scaling, not the statistical rate, and the
biological benchmark of Section~\ref{subsec:exp-bio} is a
small-$N$ applicability check, not scaling evidence. The theory
concerns the surrogate minimizer and its oracle transfer to
global minimizers of the practical objective; the Adam-trained
implementation's statistical behavior is not covered by the
conditional budget.



\begin{funding}
This work was supported by the National Natural
Science Foundation of China under Grants T2225019 and 62631026.
\end{funding}

\begin{supplement}
\stitle{Supplement to ``Scalable minimum-volume simplex
estimation with non-asymptotic analysis''}
\sdescription{All proofs and deferred quantitative discussions (Section~S1), the sensitivity study (Section~S2), reproducibility details (Section~S3), and a notation table. DOI to be supplied upon assignment.}
\end{supplement}
\begin{supplement}
\stitle{MATLAB code for DeepMVSA}
\sdescription{MATLAB implementation of the estimator, the synthetic-data demo, per-replication logs, and baseline configurations, publicly
available at \\   }
\end{supplement}
\begin{supplement}
\stitle{Data availability}
\sdescription{Synthetic data are generated as in Section~\ref{sec:exp} and \cite{dmvsa2026supp}; the micro-array dataset is that of \cite{shenorr2010}; the Cuprite scene is distributed by the U.S.\ Geological Survey; the Geology-1 scene was provided by the satellite operator. }
\end{supplement}

\begingroup\makeatletter\def\bibfont{\scriptsize}\makeatother
\setlength{\itemsep}{0.2pt}\setlength{\parsep}{0pt}\setlength{\parskip}{0pt}

\endgroup

\end{document}


\begin{frontmatter}
\title{Supplementary material for
``Scalable Minimum-Volume Simplex Estimation with
Non-asymptotic Analysis''}
\runtitle{Supplement: Scalable Simplex Estimation}

\begin{aug}
\author[A]{\fnms{Jun}~\snm{Li}\ead[label=e1]{lijuncug@cug.edu.cn}}
\author[A]{\fnms{Yanlong}~\snm{Guo}\ead[label=e2]{guoyanlong@cug.edu.cn}}
\author[A]{\fnms{Zhaozhao}~\snm{Zeng}\ead[label=e3]{zengzhaozhao@cug.edu.cn}}

\address[A]{{School of Computer Science, China University of Geosciences; corresponding author: Jun Li}\printead[presep={,\ }]{e1}\printead[presep={,\\ }]{e2}\printead[presep={,\ }]{e3}}
\end{aug}

\begin{abstract}
This supplement contains a self-contained geometric dichotomy
underpinning the localization arguments, complete proofs of all
theoretical results stated in the main text, the
hyperparameter-sensitivity protocol, and implementation and
reproducibility details.
\end{abstract}

\end{frontmatter}

\section{Proofs}\label{app:proofs}

Throughout this supplement, references to displays numbered
$(1)$--$(33)$ and to theorem-like environments numbered by the
main text (e.g.\ Theorem~3.1, Assumption~2.2) point to the main
paper; lemmas, equations, and tables numbered with an S prefix
are local to this supplement. The subsections of
Section~\ref{app:proofs} are ordered by logical dependence rather
than by statement number: the oracle transfer (Proposition~3.2)
and the operation count (Theorem~3.3) precede the noise stability
(Proposition~3.1) and approximation-rate (Proposition~3.3) proofs
that rely on the notation they fix.

The table below collects the notation used throughout the paper
and this supplement; constants $c_1,c_3,c_4,c_5,c_g,C'$ are
explicit, and their roles and values are recalled in the third
column.

\begin{center}
\textbf{Table of principal notation.}\\[4pt]
\footnotesize
\setlength{\tabcolsep}{2pt}
\begin{tabular}{lll}
\toprule
Symbol & Meaning & Value/role \\
\midrule
$N$, $K$, $L$ & sample size; simplex dimension; ambient dimension & $K+1$ vertices \\
$d$ & coordinate dimension of the A-Net input & $d=2$ imagery, $d=1$ vectorized \\
$\mathcal{S}_T$, $M_T$, $\theta_k$ & true simplex, vertex matrix, vertices & $k=0,\dots,K$ \\
$\Delta_K$, $a_j$ & probability simplex; abundance vector & $a_j\in\Delta_K$ \\
$x_j$, $X_j$ & raw observation; projected/whitened observation & $\mathbb{R}^{L}$; $\mathbb{R}^{K}$ \\
$c_j$ & normalized coordinate input & $c_j\in[0,1]^d$ \\
$\psi=(Q,\theta_0)$, $\xi$, $f_\xi$ & dual simplex parameters; A-Net parameters/map & \\
$\gamma_{\mathrm{vol}},\gamma_{\mathrm{enc}}$ & volume and enclosure weights & $\gamma_{\mathrm{vol}}\le C'$ \\
$\gamma_{\mathrm{so}},\lambda_{\mathrm{nn}},\lambda_{\mathrm{eq}},\kappa_0$
  & calibration/safeguard/gating weights & \\
$D_T$, $h_{\min}$, $V_T$ & diameter, minimum facet height, volume of $\mathcal{S}_T$ & \\
$\epsilon$, $\zeta$ & target accuracy; confidence level & \\
$R_N$, $\mathrm{enc}_N$ & empirical surrogate risk; empirical enclosure violation & \\
$\eta_{\mathrm{apx}}$, $\eta_{\mathrm{opt}}$ & approximation and optimization errors & \\
$b$ & slope scale of the surrogate risk & $(K/\epsilon)\max\{1,2/\kappa_T\}$ \\
$\kappa_T$ & geometry constant of Lemma~S4 & $KV_T/(4(K+1)(2D_T)^{K-1})$ \\
$c_1$ & localization constant of Lemma~S4 & $1/16$ \\
$c_3$ & population gap constant of Lemma~S4 & $1/32$ \\
$c_g$ & dichotomy constant of Lemma~S1 & $1/2$ \\
$\bar c_3$ & effective gap constant & $\min\{c_3,\gamma_{\mathrm{vol}}c_g/2\}$ \\
$c_4$ & enclosure-deviation constant & $2D_T$ \\
$c_5$ & facet-partition constant & depends on $K$ and facet geometry \\
$C'$ & admissible cap on $\gamma_{\mathrm{vol}}$ & $\min\{1/32,1/(32\log4)\}\approx0.0225$ \\
$C_1$, $C_2$ & uniform-deviation constant; transfer constant & (S3); $C_2=1$ \\
$N_0(K)$ & window-event threshold of Theorem~3.1 & $O((4(K+1))^{K}K\log K)$ \\
$v_0$ & lower volume fraction of the envelope & $1/4$ \\
\bottomrule
\end{tabular}
\end{center}

A few letters carry two context-separated roles, disambiguated
as follows. The surrogate slope $b$ of (21) never shares a
display with the network biases $b_1,b_2,b_3,b_{g1},b_{g2}$ of
(10)--(13) or the facet biases $b_k$ of Section~2. The A-Net
hidden width $h$ is distinct from the facet altitudes
$h_k,h_{\min}$ and from the homogeneous-coordinate superscript
in $X_j^{h},Q_h$ (Section~S3). The ambient dimension $L$, the
Lipschitz constant of Assumption~3.2, and the triangular factor
$L(\psi_L)$ of (5) never occur in the same display; the spectral
library of Section~4.3 is denoted $\mathcal{R}$. The coordinate
dimension $d$ is distinct from the planar distance
$d_{\mathcal{S}}$, which is always subscripted. The gating
scalar $\kappa$ of (14) is distinct from the geometry constant
$\kappa_T$ of (21). Among script letters, $\mathcal{S}$ denotes
a generic simplex, $\mathbb{S}_K$ the simplex class,
$\mathcal{S}_T$ the truth, $\mathfrak{S}_T$ the envelope, and
$\mathcal{S}^s,\mathcal{S}_s$ homothetic copies; the noisy
observations of Propositions~3.1 and~3.4(i) are denoted
$\widetilde X_j$, reserving $Y$ for the data matrix
$[x_1,\dots,x_N]$.

\subsection{A geometric dichotomy}
\label{app:dichotomy}

The localization arguments below rest on a set-theoretic
dichotomy between \emph{missing mass} and \emph{volume
inflation}. We state and prove it here in full; the statement
plays the role of Claim~1 of \cite{najafi2021aos} in earlier
analyses, but is self-contained and carries the explicit
constant $1/2$.

\begin{lemma}[Missing mass or volume inflation]
  \label{lem:dichotomy}
  Let $\mathcal{S},\mathcal{S}_T\in\mathbb{S}_K$ and write
  $t:=\TV(\mathbb{P}_{\mathcal{S}},\mathbb{P}_{\mathcal{S}_T})$
  and $V_T:=\Vol(\mathcal{S}_T)$. Then at least one of
  \[
    \textup{(i)}\ \
    \Vol(\mathcal{S}_T\setminus\mathcal{S})
    \;\ge\;\frac{t}{2}\,V_T,
    \qquad
    \textup{(ii)}\ \
    \Vol(\mathcal{S})
    \;\ge\;\Bigl(1+\frac{t}{2}\Bigr)V_T
  \]
  holds.
\end{lemma}

\begin{proof}
  For uniform distributions on sets of finite positive volume,
  \[
    t \;=\; 1-\frac{\Vol(\mathcal{S}\cap\mathcal{S}_T)}
    {\max\{\Vol(\mathcal{S}),V_T\}}.
  \]
  If $\Vol(\mathcal{S})\le V_T$, then
  $\Vol(\mathcal{S}_T\setminus\mathcal{S})
  =V_T-\Vol(\mathcal{S}\cap\mathcal{S}_T)=tV_T$, so (i) holds.
  If $\Vol(\mathcal{S})>V_T$ and (i) fails, then
  $\Vol(\mathcal{S}\cap\mathcal{S}_T)
  =V_T-\Vol(\mathcal{S}_T\setminus\mathcal{S})
  >(1-t/2)V_T$, while
  $\Vol(\mathcal{S}\cap\mathcal{S}_T)
  =(1-t)\Vol(\mathcal{S})$; combining,
  \[
    \Vol(\mathcal{S})
    \;>\;\frac{1-t/2}{1-t}\,V_T
    \;\ge\;\Bigl(1+\frac{t}{2}\Bigr)V_T,
  \]
  because $(1-t)(1+t/2)=1-t/2-t^2/2\le1-t/2$.
\end{proof}

\subsection{Proof of Theorem~3.1}

The proof follows the strategy of \cite{najafi2021aos},
Theorem~3.2 and Corollary~3.1, adapted to the windowed
log-volume surrogate (21). Throughout, we work
in the whitened coordinate system, write
$V_T=\Vol(\mathcal{S}_T)$, $D_T=\diam(\mathcal{S}_T)$, and take
the loss steepness $b$ as in (21), agreeing
with Corollary~3.1 of \cite{najafi2021aos} up to the
geometry-dependent factor $\max\{1,2/\kappa_T\}$. The population risk is
\[
  R(\mathcal{S}) :=
  \gamma_{\mathrm{vol}}\log\Vol(\mathcal{S})
  + \mathbb{E}\,\ell_b\bigl(d_{\mathcal{S}}(X)\bigr)
  + \gamma_{\mathrm{enc}}\,
    \mathbb{E}\bigl\|\min\bigl(a(X;\mathcal{S}),0\bigr)\bigr\|_2^2,
  \qquad X\sim\mathbb{P}_{\mathcal{S}_T}.
\]
Since $X_j\in\mathcal{S}_T$ almost surely,
$d_{\mathcal{S}_T}(X_j)=0$ and $a(X_j;\mathcal{S}_T)\ge 0$, so
both loss terms vanish at the true simplex and
\begin{equation}
  R_N(\mathcal{S}_T)=R(\mathcal{S}_T)
  =\gamma_{\mathrm{vol}}\log V_T .
  \label{eq:oracle}
\end{equation}

\begin{lemma}[Data-driven window]
  \label{lem:window}
  There exists $N_0=N_0(K)$, of order $K^{O(K)}$
  (quantified at the end of the proof), such that, for all $N\ge N_0$, the
  event $E_N$ of (18) has probability at
  least $1-2/N$. On $E_N$, every parameter pair $(\psi,\theta_0)$
  with $\mathcal{S}(Q(\psi)^{-1};\theta_0)=\mathcal{S}_T$
  satisfies $(\psi,\theta_0)\in\widehat\Psi_N$, and the class
  $\mathcal{C}_N$ of (22) is contained in the
  deterministic envelope $\mathfrak{S}_T$ of
  (19).
\end{lemma}

\begin{proof}
  The upper bounds $\widehat V_N\le V_T$ and
  $\widehat D_N\le D_T$ hold almost surely, since
  $X_j\in\mathcal{S}_T$. For the lower bound on
  $\widehat D_N$, let $u,v\in\mathcal{S}_T$ attain the
  diameter, $\|u-v\|_2=D_T$. For any $r\le D_T$ and any
  $z\in\mathcal{S}_T$, convexity gives
  $u+t(z-u)\in\mathcal{S}_T\cap B(u,r)$ for
  $t\le r/D_T$, so
  \begin{equation}
    \label{eq:cap}
    \Vol\bigl(\mathcal{S}_T\cap B(u,r)\bigr)
    \;\ge\;\Bigl(\frac{r}{D_T}\Bigr)^{K}V_T ,
  \end{equation}
  and the same bound holds at $v$. With $r=D_T/4$,
  \[
    \mathbb{P}\bigl(\{X_j\}_{j=1}^N\cap B(u,r)=\varnothing\bigr)
    \le \bigl(1-4^{-K}\bigr)^{N}\le e^{-N4^{-K}},
  \]
  and likewise at $v$; on the complement of these two events,
  $\widehat D_N\ge D_T-2r=D_T/2$. For the lower bound on
  $\widehat V_N$, write $a(x)\in\mathbb{R}^{K+1}$ for the
  barycentric coordinates of $x$ relative to $\mathcal{S}_T$
  and consider the $K+1$ vertex caps
  $C_k:=\{x\in\mathcal{S}_T:a_k(x)\ge1-\eta\}$ with
  $\eta:=1/(4(K+1))$; each cap is a scaled copy of
  $\mathcal{S}_T$ with $\Vol(C_k)=\eta^K V_T$ (for
  $K=1$ a cap is an interval adjacent to an endpoint, the facet
  hyperplanes being points). If every cap
  contains a sample point $x_l\in C_l$, the barycentric matrix
  $A_{kl}:=a_k(x_l)$ has diagonal entries at least $1-\eta$ and
  unit column sums, so each column is strictly
  diagonally dominant:
  $a_{ll}-\sum_{k\ne l}a_{kl}\ge(1-\eta)-\eta=1-2\eta$.
  The Ostrowski--Hadamard theorem, applied to the
  (row diagonally dominant) transpose $A^{\top}$, therefore
  gives
  \[
    \frac{\Vol(\mathrm{conv}\{x_0,\dots,x_K\})}{V_T}
    =|\det A|
    \;\ge\;
    \prod_{l=0}^{K}\Bigl(a_{ll}-\sum_{k\ne l}a_{kl}\Bigr)
    \ge (1-2\eta)^{K+1}
    \ge\tfrac12 ,
  \]
  the last bound holding since
  $1-2\eta=1-1/(2(K+1))$ and $(1-1/(2m))^{m}$ is increasing in
  $m$ with limit $e^{-1/2}$, with value $9/16$ at $K=1$. Hence
  \[
    \mathbb{P}\bigl(\widehat V_N<V_T/2\bigr)
    \le \sum_{k=0}^{K}\mathbb{P}\bigl(\{X_j\}\cap C_k=\varnothing\bigr)
    \le (K+1)\,e^{-N\eta^K}.
  \]
  Choosing $N_0=N_0(K)$ so that
  $(K+1)e^{-N\eta^K}+2e^{-N4^{-K}}\le 2/N$ for all $N\ge N_0$
  yields $\mathbb{P}(E_N)\ge1-2/N$. Solving the two
  inequalities with $\eta=1/(4(K+1))$ gives $N_0$ of order
  $K\,\eta^{-K}\log(1/\eta)+K\,4^{K}$, that is,
  $N_0=O\bigl((4(K+1))^{K}\,K\log K\bigr)=K^{O(K)}$; the
  diagonal-dominance argument above is what improves the cap
  mass from $(2(K+1)!)^{-K}$---which a permutation expansion of
  $\det A$ would require---to $(4(K+1))^{-K}$. The threshold
  $N_0$ nevertheless grows exponentially in $K\log K$, so
  Theorem~3.1 is a fixed-$K$ guarantee: it does not cover
  regimes in which $K$ grows with $N$, and the probability
  guarantee $1-\zeta-2/N$ is informative only for
  $N\ge\max\{N_0,4\}$. This threshold enters
  the results only through the probability $2/N$ of the window
  event and not through the rate of Theorem~3.1. On $E_N$,
  $\Vol(\mathcal{S}_T)=V_T\in[\widehat V_N/2,2\widehat V_N]$
  and $\diam(\mathcal{S}_T)=D_T\le2\widehat D_N$
  ; moreover, the sample point in the cap $C_k$ lies
  within $\eta D_T\le D_T\le 2\widehat D_N$ of vertex $k$ of
  $\mathcal{S}_T$, so $\mathcal{S}_T$ also satisfies the
  positional constraint of (17), so any
  parameterization of $\mathcal{S}_T$ lies in
  $\widehat\Psi_N$; and for every
  $(\psi,\theta_0)\in\widehat\Psi_N$,
  $\Vol\ge\widehat V_N/2\ge V_T/4$,
  $\Vol\le2\widehat V_N\le2V_T$,
  $\diam\le2\widehat D_N\le2D_T$, and every
  vertex lies within $2\widehat D_N\le 2D_T$ of the data, hence
  of $\mathcal{S}_T$, which is
  $\mathcal{C}_N\subseteq\mathfrak{S}_T$.
\end{proof}

\begin{lemma}[Localization]
  \label{lem:bounded}
  On the event $E_N$ of Lemma~\ref{lem:window}, any minimizer
  $\widehat{\mathcal{S}}$ of (21)
  over $\mathcal{C}_N$ satisfies
  $V_T/4\le\Vol(\widehat{\mathcal{S}})\le V_T$,
  $\diam(\widehat{\mathcal{S}})\le 2D_T$, and
  $\max_k\mathrm{dist}(v_k(\widehat{\mathcal{S}}),\mathcal{S}_T)
  \le 2D_T$. In particular,
  $\widehat{\mathcal{S}}$ is a bounded-degree simplex: all of
  its facet angles are bounded below by a constant
  $c(D_T,K)>0$.
\end{lemma}

\begin{proof}
  The bounds $\Vol(\widehat{\mathcal{S}})\ge\widehat V_N/2
  \ge V_T/4$ and $\diam(\widehat{\mathcal{S}})\le 2\widehat D_N
  \le 2D_T$ are imposed directly by the data-driven window
  (17) on $E_N$. For the volume upper bound,
  the planar-distance and enclosure terms of
  (21) are nonnegative, so
  $R_N(\widehat{\mathcal{S}})\allowbreak
   \ge \gamma_{\mathrm{vol}}\log\Vol(\widehat{\mathcal{S}})$, while
  $\mathcal{S}_T\in\mathcal{C}_N$ by
  Assumption~3.1(i) and Lemma~\ref{lem:window}, and hence
  $R_N(\widehat{\mathcal{S}})\le R_N(\mathcal{S}_T)
   =\gamma_{\mathrm{vol}}\log V_T$ by \eqref{eq:oracle}.
  Combining the two inequalities gives
  $\Vol(\widehat{\mathcal{S}})\le V_T$.
  Finally, a simplex with $\Vol\ge V_T/4$ and
  $\diam\le 2D_T$ has every altitude bounded below by a constant
  multiple of $V_T/(4(2D_T)^{K-1})$, because
  $\Vol=\frac{1}{K}\cdot(\text{facet area})\cdot(\text{altitude})$
  and all facet areas are at most $(2D_T)^{K-1}$; the facet angles
  are therefore bounded below by a constant $c(D_T,K)>0$.
\end{proof}

Recall the deterministic envelope
\[
  \mathfrak{S}_T
  = \bigl\{\mathcal{S}\in\mathbb{S}_K:\,
  \tfrac14 V_T\le\Vol(\mathcal{S})\le 2V_T,\
  \diam(\mathcal{S})\le 2D_T,\
  \max_k\,\mathrm{dist}(v_k(\mathcal{S}),\mathcal{S}_T)
  \le 2D_T\bigr\}
\]
of (19), which contains both
$\widehat{\mathcal{S}}_{\mathrm{DEEP}}$
(Lemma~\ref{lem:bounded}) and $\mathcal{S}_T$ on $E_N$; the
remaining arguments are uniform over $\mathfrak{S}_T$ and
therefore do not depend on the realized sample through the
window.

\begin{lemma}[Population gap]
  \label{lem:gap}
  Let $b$ be as in (21),
  $c_1:=1/16$, and
  $c_3:=c_1/2=1/32$. Whenever
  $\gamma_{\mathrm{vol}}\le C':=
  \min\{c_1/2,\,1/(32\log(1/v_0))\}$ with $v_0=1/4$,
  \[
    R(\mathcal{S})-R(\mathcal{S}_T)\;\ge\;c_3\,\epsilon
  \]
  for every $\mathcal{S}\in\mathfrak{S}_T$ with
  $\Vol(\mathcal{S})\le V_T$ and
  $\TV(\mathbb{P}_{\mathcal{S}},\mathbb{P}_{\mathcal{S}_T})
  \ge\epsilon$.
\end{lemma}

\begin{proof}
  Fix such an $\mathcal{S}$ and write
  $t_0:=\TV(\mathbb{P}_{\mathcal{S}},
  \mathbb{P}_{\mathcal{S}_T})\ge\epsilon$. Since
  $\Vol(\mathcal{S})\le V_T$, Lemma~\ref{lem:dichotomy} places
  us in alternative~(i): with
  $\mathcal{A}:=\mathcal{S}_T\setminus\mathcal{S}$ and
  $\delta:=\Vol(\mathcal{A})/V_T$,
  \[
    \delta\;\ge\;\frac{t_0}{2}\;\ge\;\frac{\epsilon}{2}.
  \]
  (The restriction $\Vol(\mathcal{S})\le V_T$ is legitimate
  because the lemma is applied only to the minimizer
  $\widehat{\mathcal{S}}$, which satisfies it by
  Lemma~\ref{lem:bounded}; it also excludes
  alternative~(ii), so no lower bound on $\gamma_{\mathrm{vol}}$
  is required.)

  The points of $\mathcal{A}$ within distance $t$ of
  $\mathcal{S}$ are covered by the $K+1$ facet slabs
  $\{x\in\mathcal{S}_T:
  0\le \boldsymbol{w}_k^\top x+b_k\le t\}$ of
  $\mathcal{S}$, since for $x\notin\mathcal{S}$ the planar
  distance $d_{\mathcal{S}}(x)=\max_k(\boldsymbol{w}_k^\top
  x+b_k)$ is attained at some facet $k$. Each slab has volume at
  most $(2D_T)^{K-1}t$, because every hyperplane slice of
  $\mathcal{S}_T$ has $(K-1)$-area at most $(2D_T)^{K-1}$:
  the slice is a $(K-1)$-dimensional set of diameter at
  most $D_T$, so by the isodiametric inequality in
  $\mathbb R^{K-1}$ its $(K-1)$-area is at most
  $\omega_{K-1}(D_T/2)^{K-1}
  \le(2D_T)^{K-1}$ (with the convention $\omega_0:=1$,
  a $0$-area being a counting measure, so the bound reads
  $1\le1$ at $K=1$). Taking
  \[
    t\;:=\;\frac{\Vol(\mathcal{A})}
    {2(K+1)(2D_T)^{K-1}}
  \]
  therefore places at least half of $\mathcal{A}$ at distance
  at least $t$ from $\mathcal{S}$. By
  $b=(K/\epsilon)\max\{1,2/\kappa_T\}$ with
  $\kappa_T:=KV_T\big/\bigl(4(K+1)(2D_T)^{K-1}\bigr)$,
  \[
    bt\;\ge\;\frac{K}{\epsilon}\cdot
    \frac{\delta V_T}{2(K+1)(2D_T)^{K-1}}
    \cdot\max\{1,2/\kappa_T\}
    \;\ge\;\kappa_T\max\{1,2/\kappa_T\}
    \;=\;\max\{\kappa_T,2\}\;\ge\;2,
  \]
  so $1-e^{-bt}\ge 1-e^{-2}\ge 1/2$ and, integrating over the
  far half of $\mathcal{A}$,
  \[
    \mathbb{E}\,\ell_b\bigl(d_{\mathcal{S}}(X)\bigr)
    \;\ge\;\frac{1}{V_T}
    \int_{\mathcal{A}\cap\{d_{\mathcal{S}}\ge t\}}
    \bigl(1-e^{-b d_{\mathcal{S}}(x)}\bigr)\,dx
    \;\ge\;\frac{\Vol(\mathcal{A})}{4V_T}\,(1-e^{-2})
    \;\ge\;\frac{\delta}{8}
    \;=:\;2c_1\,\delta .
  \]
  Meanwhile, the volume reward of shrinking is
  controlled by the missing mass itself:
  since $\Vol(\mathcal{S})\ge\Vol(\mathcal{S}\cap\mathcal{S}_T)
   =(1-\delta)V_T$,
  \[
    \gamma_{\mathrm{vol}}\log\frac{V_T}{\Vol(\mathcal{S})}
    \;\le\;\gamma_{\mathrm{vol}}\,
           \log\frac{V_T}{\Vol(\mathcal{S}\cap\mathcal{S}_T)}
    \;=\;\gamma_{\mathrm{vol}}\log\frac{1}{1-\delta}
    \;\le\;2\,\gamma_{\mathrm{vol}}\,\delta
  \]
  for $\delta\le 1/2$, and it suffices to treat
  $\epsilon\le 1$ since the claim is monotone in
  $\epsilon$; for $\delta>1/2$ the loss bound
  $\mathbb{E}\,\ell_b\bigl(d_{\mathcal{S}}(X)\bigr)\ge\delta/8
  \ge 1/16$ is already of constant order, while the reward is at
  most $\gamma_{\mathrm{vol}}\log(1/v_0)$ on the envelope, so
  $\gamma_{\mathrm{vol}}\le 1/(32\log(1/v_0))$ yields a gap
  of at least $1/16-1/32=1/32\ge c_3\epsilon$, because
  $\epsilon\le 1$ and $c_3=1/32$. The enclosure term is
  nonnegative. For $\delta\le1/2$,
  $\gamma_{\mathrm{vol}}\le c_1/2$ yields
  \[
    R(\mathcal{S})-R(\mathcal{S}_T)
    \;\ge\;2c_1\delta-2\gamma_{\mathrm{vol}}\delta
    \;\ge\;c_1\delta\;\ge\;c_1\,\frac{\epsilon}{2}
    \;=\;c_3\,\epsilon .
  \]
  The enclosure term is nonnegative
  throughout and only enlarges the gap; in the noisy model it is
  the term that controls simplices whose facets cut through
  regions of positive data mass.
\end{proof}

\paragraph{Uniform deviation}
All three empirical terms of (21) are indexed
by $\mathcal{S}\in\mathfrak{S}_T$ and are uniformly bounded
on the support of $\mathbb{P}_{\mathcal{S}_T}$: the
planar-distance loss takes values in $[0,1]$; the log-volume
term in $[\log(V_T/4),\log(2V_T)]$; and the enclosure
integrand satisfies
$\|\min(a(x;\mathcal{S}),0)\|_2^2\le
(K+1)\bigl(1+3D_T/h_{\min}\bigr)^2$ for
$x\in\mathcal{S}_T$, because
$a_k(\cdot;\mathcal{S})$ is affine, equals one at the
$k$-th vertex $v_k$ of $\mathcal{S}$, and has gradient norm
$1/h_k$, so $|a_k(x;\mathcal{S})|\le 1+\|x-v_k\|_2/h_k$,
and the positional clause of $\mathfrak{S}_T$ gives
$\|x-v_k\|_2\le D_T+2D_T=3D_T$ for $x\in\mathcal{S}_T$;
every altitude $h_k$ of $\mathcal{S}\in\mathfrak{S}_T$ obeys
$h_k\ge h_{\min}:=c(K)\,V_T\big/(2D_T)^{K-1}>0$ by the
altitude argument of Lemma~\ref{lem:bounded}. Each class has
pseudo-dimension $O(K^2\log K)$. For the
planar-distance class: affine functionals on $\mathbb R^K$
form a VC-subgraph class whose growth function obeys the
Sauer bound $\Pi_{\mathrm{aff}}(N)\le(Ne/(K+1))^{K+1}$; the
pointwise maximum of $K+1$ such functionals has at most
$\Pi_{\mathrm{aff}}(N)^{K+1}\le(Ne/(K+1))^{(K+1)^2}$
restrictions to any $N$-point set, since each restriction is
determined by those of its $K+1$ affine components; composing
with a fixed increasing univariate function preserves the
growth function; and the growth-function--pseudo-dimension
relation converts $(Ne/(K+1))^{(K+1)^2}$ to pseudo-dimension
$O(K^2\log K)$ \cite{anthony1999}. The enclosure class consists of a fixed
quadratic composed with the $K+1$ truncated affine maps
$x\mapsto\min(a_k(x;\mathcal{S}),0)$, each of
pseudo-dimension $O(K)$, so the standard composition rules for
pseudo-dimension \cite{anthony1999} give $O(K^2\log K)$; and
the log-volume term is non-random and identical
in $R_N$ and $R$, so it cancels exactly in the deviation. Applying the standard
symmetrization-and-chaining uniform-deviation bound for
bounded function classes of finite pseudo-dimension
\cite{anthony1999} to each of the three classes, followed by
a union bound, with probability at least $1-\zeta$,
\begin{equation}
  \sup_{\mathcal{S}\in\mathfrak{S}_T}
  \bigl|R_N(\mathcal{S})-R(\mathcal{S})\bigr|
  \;\le\;
  C_1\sqrt{\frac{K^2\log(N/K)+\log(1/\zeta)}{N}},
  \label{eq:uniform}
\end{equation}
where $C_1=C_1(K,\gamma_{\mathrm{enc}},D_T,h_{\min})$
does not depend on
$(N,\epsilon,\zeta)$. Explicitly, the
bound scales with the ranges of the two non-degenerate
empirical terms---the planar-distance class takes values in
$[0,1]$ and the weighted enclosure class in
$[0,\gamma_{\mathrm{enc}}(K+1)a_{\max}^2]$ with
$a_{\max}:=1+3D_T/h_{\min}$---so
$C_1=c\,\bigl(1+\gamma_{\mathrm{enc}}(K+1)a_{\max}^2\bigr)$
for a universal constant $c$ absorbing the pseudo-dimension
constants of both classes; in particular $C_1$ grows
linearly with the enclosure weight $\gamma_{\mathrm{enc}}$,
a dependence that the end-to-end analysis of
Theorem~3.2 removes through the variance-adaptive bound
of Lemma~\ref{lem:relative}. Note that
$\mathcal{S}_T\in\mathfrak{S}_T$, so the deviation bound
applies at $\mathcal{S}_T$ as well.

\paragraph{Conclusion}
We work on the intersection of $E_N$
(Lemma~\ref{lem:window}) and the deviation event of
\eqref{eq:uniform}, which has probability at least
$1-\zeta-2/N$. Suppose
$\TV(\mathbb{P}_{\widehat{\mathcal{S}}_{\mathrm{DEEP}}},
\mathbb{P}_{\mathcal{S}_T})\ge\epsilon$. Since
$\Vol(\widehat{\mathcal{S}}_{\mathrm{DEEP}})\le V_T$ by
Lemma~\ref{lem:bounded}, Lemma~\ref{lem:gap} applies;
combining it with the comparison
$R_N(\widehat{\mathcal{S}}_{\mathrm{DEEP}})\le
R_N(\mathcal{S}_T)$ established in Lemma~\ref{lem:bounded} and
\eqref{eq:uniform} applied to
$\widehat{\mathcal{S}}_{\mathrm{DEEP}}$ and $\mathcal{S}_T$,
\[
  c_3\epsilon
  \;\le\;R(\widehat{\mathcal{S}}_{\mathrm{DEEP}})-R(\mathcal{S}_T)
  \;\le\;2C_1\sqrt{\frac{K^2\log(N/K)+\log(1/\zeta)}{N}} .
\]
For
$N\ge C(K)\,
\epsilon^{-2}\bigl[K^2\log(K/\epsilon)+\log(1/\zeta)\bigr]$
the right-hand side is strictly smaller than $c_3\epsilon$
(the $\log(N/K)$ factor satisfies
$\log(N/K)\asymp\log(K/\epsilon)$ at the stated rate, so it is
absorbed into the constant $C$), which is a contradiction.
Hence $\TV(\mathbb{P}_{\mathcal{S}_T},
\mathbb{P}_{\widehat{\mathcal{S}}_{\mathrm{DEEP}}})\le\epsilon$
with probability at least $1-\zeta-2/N$.

\medskip
\noindent
\emph{Remark.} The reconstruction accuracy
$\eta_{\mathrm{apx}}$ of Assumption~3.1(ii) does not
enter the above chain: the comparison step uses only the
representability clause~(i), since the surrogate risk of the
true simplex equals $\gamma_{\mathrm{vol}}\log V_T$ exactly by
\eqref{eq:oracle}. Clause~(ii) instead calibrates the practical
objective (15) at the oracle parameter, as
discussed in the remark following Theorem~3.1.
\qed

\subsection{Proof of Proposition~3.2}

\paragraph{Oracle value of the practical objective}
Evaluate the practical objective at the
oracle parameter
$(\psi^*,\xi^*)$ of Assumption~3.1, with
anchor $\theta_0^*$. Since
$X_j\in\mathcal{S}_T$, the completed enclosure term
(16) vanishes. Take the oracle A-Net parameters $\xi^*$ of
Assumption~3.1(ii) in the \emph{base branch}, that
is, with gating scalar $\kappa^*=0$: the output
$a_j=f_{\xi^*}(c_j)$ is then the softmax vector
(12), which satisfies $\mathbf{1}^{\top}a_j=1$
and $a_j\ge0$ exactly. The sum-to-one and nonnegativity
safeguard penalties of (15) therefore vanish at
the oracle parameter, irrespective of the weights
$(\gamma_{\mathrm{so}},\lambda_{\mathrm{nn}})$. The reconstruction term of
(15) is evaluated in the whitened coordinates
of Assumption~3.1(ii), so at the oracle it is at
most $\eta_{\mathrm{apx}}$; no whitening-conditioning factor
enters. For the centroid-calibration
penalty, the barycentric coordinates of $X_j$ relative to
$\mathcal{S}_T$ satisfy
$\mathbf{1}^{\top}Q_T(X_j-\theta_0^*)=
\sum_{k=1}^{K}a_k^*(X_j)=1-a_0^*(X_j)$, so
\[
  \mathbf{1}^{\top}Q_T(\bar X-\theta_0^*)-\frac{K}{K+1}
  =\frac{1}{N}\sum_{j=1}^{N}\Bigl(\tfrac{1}{K+1}-a_0^*(X_j)\Bigr),
\]
an average of centered bounded random variables; its square is
$O_p(1/N)$, and the penalty contributes
$O_p(\lambda_{\mathrm{eq}}/N)$, negligible at the
$\epsilon^{-2}$ sample scale of Theorem~3.1. Finally, the volume term
$-\sum_k\psi_{r,k}$ differs from
$\log\Vol(\mathcal{S}(Q(\psi)^{-1};\theta_0))$ by a
parameter-independent constant (the initialization offset);
replacing it by $\log\Vol$ shifts the objective by that
constant, leaves the minimizer unchanged, and cancels in the
comparison below. Collecting terms and using
\eqref{eq:oracle},
\begin{equation}
  \label{eq:oracle-prac}
  \widehat R_{\mathrm{DEEP}}(\psi^*,\xi^*;Y)
  \;\le\; R_N(\mathcal{S}_T) + C_2\,\eta_{\mathrm{apx}}
  \,+\,O_p(\lambda_{\mathrm{eq}}/N),
\end{equation}
with
$C_2:=1$ in these whitened coordinates; the
  safeguard weights $(\gamma_{\mathrm{so}},\lambda_{\mathrm{nn}})$
  do not enter the comparison, since the corresponding penalties vanish
  at the base-branch oracle.

\paragraph{Comparison at the global minimizer}
All terms of (15) except the volume term are
nonnegative, so global optimality
$\widehat R_{\mathrm{DEEP}}(\widehat\psi,\widehat\xi;Y)
\le\widehat R_{\mathrm{DEEP}}(\psi^*,\xi^*;Y)$ together with
\eqref{eq:oracle-prac} yields
\[
  \begin{aligned}
    \gamma_{\mathrm{vol}}\log\frac{\Vol(\widehat{\mathcal{S}})}{V_T}
    &\;\le\;C_2\eta_{\mathrm{apx}}
    \,+\,O_p(\lambda_{\mathrm{eq}}/N),\\
    \gamma_{\mathrm{enc}}\,\mathrm{enc}_N(\widehat{\mathcal{S}})
    &\;\le\;C_2\eta_{\mathrm{apx}}
    +\gamma_{\mathrm{vol}}\Bigl[\log\frac{V_T}
    {\Vol(\widehat{\mathcal{S}})}\Bigr]_+
    \,+\,O_p(\lambda_{\mathrm{eq}}/N).
  \end{aligned}
\]
which is consequence (i) and (25); no
property of $\widehat{\mathcal{S}}$ beyond global optimality
is used.

\paragraph{Enclosure controls the planar distance}
For any simplex $\mathcal{S}$ and any $x$, let
$a(x)\in\mathbb{R}^{K+1}$ denote the barycentric coordinates of
$x$ with respect to $\mathcal{S}$, so that
$x=M_{\mathcal{S}}a(x)$ and $\mathbf{1}^\top a(x)=1$; in the
anchored parameterization (7), (8)
gives the identity
\[
  \|\min(a(x),0)\|_2^2
  \;=\;\bigl\|\min\bigl(Q_{\mathcal{S}}(x-\theta_0),0\bigr)\bigr\|_2^2
  +\min\bigl(1-\mathbf{1}^{\top}Q_{\mathcal{S}}(x-\theta_0),0\bigr)^{2},
\]
which is exactly the completed enclosure term of
(21); both barycentric violations, including the
one opposite the anchor vertex, are penalized.
Let
$w_k^{\top}x+\beta_k=0$ be the hyperplane containing the facet
opposite vertex $k$ of $\mathcal{S}$, with $w_k$ the outward
unit normal, and let $h_k\le\diam(\mathcal{S})$ be the
altitude from vertex $k$ to this facet. Since $a_k$ is affine,
equals one at vertex $k$, and vanishes on the opposite facet,
$a_k(x)=-(w_k^{\top}x+\beta_k)/h_k$. Definition~2.1
therefore gives
\[
  d_{\mathcal{S}}(x)
  \;=\;\max\bigl\{0,\max_k\,(w_k^{\top}x+\beta_k)\bigr\}
  \;=\;\max_k\,h_k\,\bigl(-a_k(x)\bigr)_+
  \;\le\;\diam(\mathcal{S})\,
  \|\min(a(x),0)\|_2 .
\]
On the envelope $\mathfrak{S}_T$,
$\diam(\mathcal{S})\le2D_T$ (Lemma~\ref{lem:bounded}). Since
$\ell_b(u)=1-e^{-bu}\le bu$,
\[
  \frac1N\sum_{j=1}^N
  \ell_b\bigl(d_{\widehat{\mathcal{S}}}(X_j)\bigr)
  \;\le\;b\,\frac1N\sum_{j=1}^N d_{\widehat{\mathcal{S}}}(X_j)
  \;\le\;b\,c_4\sqrt{\mathrm{enc}_N(\widehat{\mathcal{S}})},
  \qquad c_4:=2\,D_T,
\]
by Jensen's inequality. Adding the volume and enclosure terms
back yields (26).

\paragraph{Shrinkage bias and the scaling guideline}
Let $s:=\log(V_T/\Vol(\mathcal{S}))\in[0,1]$,
so that the log-volume reward of $\mathcal{S}$ relative to
$\mathcal{S}_T$ is exactly $\gamma_{\mathrm{vol}}s$. Then
$\Vol(\mathcal{S}_T\setminus\mathcal{S})\ge(1-e^{-s})V_T
\ge (s/2)V_T$. Partition the missed region
$\mathcal{A}=\mathcal{S}_T\setminus\mathcal{S}$ according to a
violated facet; on the part violating facet $k$,
$\|\min(a(x),0)\|_2\ge |a_k(x)|=u_k(x)/h_k$, where $u_k(x)$
is the distance to the corresponding facet hyperplane and
$h_k\le\diam(\mathcal{S})\le2D_T$ is the altitude. Among sets
of prescribed volume outside a fixed facet, the second moment
of $u_k$ is minimized by a slab adjacent to that facet, whose
$(K-1)$-area is at most $(2D_T)^{K-1}$; hence, by the
power-mean inequality applied to the facet partition,
\[
  \mathbb{E}\,\|\min(a(X),0)\|_2^2
  \;\ge\;\frac{\Vol(\mathcal{A})^3}
  {3(K+1)^2(2D_T)^{2K}\,V_T}
  \;\ge\;c_5\,s^3,
  \qquad
  c_5=\frac{V_T^2}{24\,(K+1)^2(2D_T)^{2K}} .
\]
At the population level (ignoring the deviation
\eqref{eq:uniform}), comparison with the oracle value
therefore gives
$\gamma_{\mathrm{enc}}c_5s^3\le
\gamma_{\mathrm{vol}}s+C_2\eta_{\mathrm{apx}}$ for the
shrinkage $s$ of any global minimizer, and solving this
inequality yields
\[
  s\;\le\;\max\Bigl\{
  \sqrt{\tfrac{2\gamma_{\mathrm{vol}}}
  {c_5\gamma_{\mathrm{enc}}}},\;
  \Bigl(\tfrac{2C_2\eta_{\mathrm{apx}}}
  {c_5\gamma_{\mathrm{enc}}}\Bigr)^{1/3}\Bigr\}.
\]
Under the scalings
\begin{equation}
  \label{eq:guideline}
  \gamma_{\mathrm{enc}}\;\ge\;
  2\,\gamma_{\mathrm{vol}}\,c_5^{-1/3}\,
  K^{4/3}\epsilon^{-8/3}
  \qquad\text{and}\qquad
  \eta_{\mathrm{apx}}\;\le\;
  \frac{\gamma_{\mathrm{enc}}\,\epsilon^4}{2C_2K^2},
\end{equation}
both terms on the right-hand side of (25) are
at most $\epsilon^4/(2K^2)$ up to a constant factor; hence
$\mathrm{enc}_N(\widehat{\mathcal{S}})\le\epsilon^4/K^2$ up
to the same factor, which is the first condition of
(27) with $b$ as in (21), since
$\bar c_3=\gamma_{\mathrm{vol}}/4$ and $c_4=2D_T$ are constant
at fixed $(\gamma_{\mathrm{vol}},D_T)$. The second
condition of (27),
$\eta_{\mathrm{apx}}\le \bar c_3\epsilon/(8C_2)$, is independent
of the penalty weights and must be supplied by the
approximation quality of the A-Net
(Remark~3.5).
\qed

\subsection{Proof of Theorem~3.3}

The $Q$-flow contributes $O(K^2)$ parameters, the anchor
contributes $K$, and the A-Net with
hidden width $h=O(K)$ contributes $O(h^2+hK+dh)=O(K^2)$
parameters, so $P=O(K^2)$; Adam maintains two auxiliary buffers
of the same size, so the parameter and optimizer-state memory is
$3P=O(K^2)$. In each stochastic gradient evaluation on a
mini-batch of size $B$, the forward pass
through the A-Net for the $B$ sampled pixels costs
$O(B(h^2+hK))=O(BK^2)$; the reconstruction term is evaluated in
the whitened $K$-dimensional space and costs
$O(BK(K+1))=O(BK^2)$; the enclosure term requires the
matrix--vector products $Q(X_j-\theta_0)$ and the
completed form (16), costing $O(BK^2)$; and
the volume term costs $O(K)$. Assembling
$Q=L(\psi_L)Q_0R(\psi_R,\psi_r)$ from its triangular factors
costs $O(K^3)$ once per iteration, as does forming the
vertex matrix from $Q^{-1}$, and back-propagation is a
constant multiple of the forward cost. The total is therefore
$O(BK^2+K^3)$ per stochastic step and
$O(NK^2+NK^3/B)=O(NK^2)$ per full data pass when
$B=\Omega(K)$ and $K\ll N$.

Peak memory decomposes as follows: parameters and optimizer
states $O(K^2)$; mini-batch activations and the batch slice of
$X$, $O(BK)$ for batch size $B$; and input storage
$O(NK)$, which is common to every method that reads the full
scene and can be streamed in batches. The $N$-independent
working memory is therefore $O(K^2)$ whenever $B=O(K)$. By
contrast, the Soft-ML implementation used here requires $O(NK^3)$
time and $O(NK)$ memory (an implementation-specific count),
and the MVSA family instantiates constraint matrices with
$O(NK)$ rows.
\qed

\subsection{Proof of Proposition~3.1}

Fix $\mathcal{S}\in\mathfrak{S}_T$. The planar-distance
map $x\mapsto d_{\mathcal{S}}(x)$ is $1$-Lipschitz,
since it is the positive part of a maximum of
affine functions with unit gradients, and $\ell_b$ is
$b$-Lipschitz; hence
\[
  \bigl|\mathbb{E}\,\ell_b(d_{\mathcal{S}}(\widetilde X)
  -\mathbb{E}\,\ell_b(d_{\mathcal{S}}(X))\bigr|
  \;\le\; b\,\mathbb{E}\|\varepsilon\|_2\le b\sigma .
\]
For the enclosure term, the barycentric coordinates
$a_k(\cdot;\mathcal{S})$ are affine with gradient norm
$1/h_k\le 1/h_{\min}$, so
$|a_k(X+\varepsilon;\mathcal{S})-a_k(X;\mathcal{S})|
\le\sigma/h_{\min}$, and on $\mathcal{S}_T$ they obey
$|a_k|\le a_{\max}$, since $a_k$ equals one at the
$k$-th vertex $v_k$ of $\mathcal{S}$ and the positional clause
of $\mathfrak{S}_T$ gives $\|X-v_k\|_2\le D_T+2D_T=3D_T$,
hence $|a_k(X)|\le 1+\|X-v_k\|_2/h_k\le 1+3D_T/h_{\min}$. Using
$|\min(u+\delta,0)^2-\min(u,0)^2|\le 2|u||\delta|+\delta^2$
coordinatewise and summing over $k=0,\dots,K$,
\[
  \bigl|\|\min(a(\widetilde X;\mathcal{S}),0)\|_2^2
  -\|\min(a(X;\mathcal{S}),0)\|_2^2\bigr|
  \;\le\;(K+1)\Bigl(\frac{2a_{\max}\sigma}{h_{\min}}
  +\frac{\sigma^2}{h_{\min}^2}\Bigr).
\]
Both bounds are uniform in $\mathcal{S}\in\mathfrak{S}_T$, and
the volume term is noise-free, which gives
(24). For the second claim, repeat the
concluding argument of Theorem~3.1 with $R$
replaced by $R^\sigma$: the comparison
$R_N(\widehat{\mathcal{S}})\le R_N(\mathcal{S}_T)$ is
unaltered; the window event $E_N$ persists under
noise with inflated constants---for
\(\sigma\le\min\{D_T/8,\eta h_{\min}\}\), absorbed into the
smallness of $c$---where $h_{\min}$ in this
bound is the minimal altitude of $\mathcal{S}_T$, governing
the cap geometry of Lemma~S2, and is not the envelope
altitude floor of Lemma~S3 entering $a_{\max}$---, the noisy data diameter and hull volume obey
$\widehat D_N^{\sigma}\in[D_T/4,2D_T]$ and
$\widehat V_N^{\sigma}\le(1+\sigma/r_T)^KV_T$ with $r_T$ the
inradius of $\mathcal{S}_T$, since the noisy points lie in
$\mathcal{S}_T+\sigma B$. The volume bound is a
determinant perturbation made precise: every hull vertex is a
noisy point, so $\mathrm{conv}\{X_j+\varepsilon_j\}\subseteq
\mathcal{S}_T+\sigma B$, and, writing $c_T$ for an incenter of
$\mathcal{S}_T$, the inclusion $B(c_T,r_T)\subseteq\mathcal{S}_T$
implies $\mathcal{S}_T+\sigma B\subseteq
c_T+(1+\sigma/r_T)(\mathcal{S}_T-c_T)$: for $x+u\in
\mathcal{S}_T+\sigma B$, the point
$(1+\sigma/r_T)^{-1}(x-c_T)+(1+\sigma/r_T)^{-1}u$ is a convex
combination of $x-c_T\in\mathcal{S}_T-c_T$ and
$(r_T/\sigma)u\in B(0,r_T)\subseteq\mathcal{S}_T-c_T$ with weights
$(1+\sigma/r_T)^{-1}$ and $(\sigma/r_T)(1+\sigma/r_T)^{-1}$ summing
to one, hence lies in $\mathcal{S}_T-c_T$. Since
$\Vol(\mathcal{S})=|\det(v_1-v_0,\dots,v_K-v_0)|/K!$ scales by the
factor $(1+\sigma/r_T)^K$ under the homothety,
$\widehat V_N^{\sigma}\le(1+\sigma/r_T)^KV_T$, and the cap events of
Lemma~\ref{lem:window} apply with $\eta$ replaced by
$2\eta$---so the localization of Lemmas~\ref{lem:window} and
\ref{lem:bounded} is retained on an event of the same
probability; the population gap of Lemma~\ref{lem:gap} applies to
$R$; under noise $\Vol(\widehat{\mathcal{S}})\le V_T$ is
unavailable, so the volume-inflation alternative of
Lemma~\ref{lem:dichotomy} is no longer excluded, and in that
branch $\Vol(\mathcal{S})\ge(1+c_{\mathrm g}\epsilon)V_T$, so the
volume term alone contributes
$\gamma_{\mathrm{vol}}\log(1+c_{\mathrm g}\epsilon)
\ge\gamma_{\mathrm{vol}}c_{\mathrm g}\epsilon/2$ (using
$\log(1+u)\ge u/2$ on $[0,1]$)---the same computation as in the
proof of \eqref{eq:gap-tv} in Section~\ref{subsec:endtoend}, recorded here so that this proof is self-contained; the effective gap constant is therefore
$\min\{c_3,\gamma_{\mathrm{vol}}c_{\mathrm g}/2\}=\bar c_3$,
and the two replacements cost at most $2\Delta_{\sigma}
\le \bar c_3\epsilon/2$, leaving a residual gap of
$\bar c_3\epsilon/2$ against the uniform deviation; the same
sample-size condition then yields the contradiction. Sub-Gaussian
noise with parameter $\sigma_0$ reduces to the bounded case at
level $\sigma=\sigma_0\sqrt{2K\log (2KN^2)}$ on an event of
probability $1-1/N$, by a union bound over the $NK$
coordinates: $2NK\,e^{-t^2/(2\sigma_0^2)}\le 1/N$ at
$t=\sigma_0\sqrt{2\log(2KN^2)}$ and
$\|\varepsilon_j\|_2\le\sqrt{K}\,t$, and the threshold condition becomes
$\sigma_0=O\bigl(\epsilon^2/(K^{3/2}\sqrt{\log N})\bigr)$
, of unchanged order since $\log(2KN^2)=O(\log N)$
for $K\le N$.
This is a perturbation analysis around the
noiseless model of \cite{najafi2021aos}, whose guarantees
require exact containment $X_j\in\mathcal{S}_T$; the
information-theoretic noisy-regime sample complexity of
\cite{saberi2023} is complementary, as discussed after
Proposition~3.1.
\qed

\subsection{Proof of Proposition~3.3}

Set $\tau\in(0,1]$ to be chosen below and
write $z_k(c):=\log(g_k(c)+\tau)$, $k=0,\dots,K$. Since
$0\le g_k\le1$ and $g_k$ is $L$-Lipschitz, $z_k$ is
$(L/\tau)$-Lipschitz and $\log\tau\le z_k\le\log(1+\tau)$.
By the quantitative sup-norm approximation theorem for
sigmoidal networks \cite{mhaskar1996}---whose smoothness and
nonpolynomiality requirements are met by $\tanh$,
and whose one-hidden-layer construction embeds into the
two-hidden-layer base branch by letting the first hidden layer
realize the cited construction and the second act as an
approximate identity, $\tanh(\lambda u)/\lambda=u+O(\lambda^2)$
uniformly on compacta as $\lambda\downarrow0$---applied to a mollification of $z_k$ at the scale
$h^{-1/d}$ (a standard reduction: the mollified field is
$C^1$ with gradient norm at most $L/\tau$, to which the
$C^1$ case of the theorem applies, and the mollification
error $(L/\tau)h^{-1/d}$ is of the same order), each $z_k$
admits an
approximation $\widehat z_k$ of the form
(10)--(12) with hidden width
$h$ such that
\[
  \sup_{c\in[0,1]^d}\bigl|\widehat z_k(c)-z_k(c)\bigr|
  \;\le\;C_a\,\frac{L}{\tau}\,h^{-1/d},
\]
with $C_a$ depending only on $d$; the construction localizes
to a regular partition of $[0,1]^d$ into $\Theta(h)$ cells,
so the hidden features are shared across the $K+1$
components---the partition is a function-independent
regular grid, and since every $z_k$ is $(L/\tau)$-Lipschitz the
same partition certifies all $K+1$ approximations
simultaneously, so the simultaneous claim does follow from the
scalar theorem---and only the readout layer $W_3$ of
(12) is component-specific, in keeping with
the base-branch architecture. The weights of the
cited construction grow polynomially with $h^{1/d}$; the result
is an existence statement at the stated width and makes no
claim about the norm of the trained weights, which
the theory does not constrain. Take
$f_\xi(c)=\softmax(\widehat z(c))$ (base branch,
$\kappa=0$). The softmax map satisfies
$|\softmax(u)_k-\softmax(v)_k|\le\tfrac12\|u-v\|_\infty$:
its Jacobian $\diag(p)-pp^{\top}$ has row absolute sums
$2p_k(1-p_k)\le1/2$, so its $\ell_\infty\to\ell_\infty$
operator norm is at most $1/2$; and a direct expansion gives
$|\softmax(z(c))_k-g_k(c)|\le(K+2)\tau$, because
$\softmax(z(c))_k=(g_k(c)+\tau)\big/(1+(K+1)\tau)$. Hence
\[
  \bigl|f_\xi(c)_k-g_k(c)\bigr|
  \;\le\;\frac{C_aL}{2\tau}\,h^{-1/d}+(K+2)\tau
  \;\le\;\sqrt{2C_a(K+2)L}\;h^{-1/(2d)},
\]
where the second inequality is AM--GM applied at the
optimizer $\tau_*$---the balance point displayed below---at which it
holds with equality; it is therefore valid whenever $\tau_*\le1$,
independently of the size of $2C_a(K+2)L$.
At the balance
$\tau=\sqrt{C_aL/(2(K+2))}\,h^{-1/(2d)}$; if this balance
point exceeds $1$, i.e.\ $h<\bigl(C_aL/(2(K+2))\bigr)^{d}$,
the right-hand side of (28) exceeds
$4(K+1)(K+2)^2\ge2$, while
$\sup_c\|f_\xi(c)-g(c)\|_2^2\le2$ for simplex-valued $f_\xi$
and $g$, so the claim holds trivially and the restriction
$\tau\le1$ is legitimate. Squaring and
summing over $k=0,\dots,K$ gives (28) with
$C_0:=2C_a$, and (29) follows from
\[
\begin{aligned}
  \bigl\|X_j-M_Tf_\xi(c_j)\bigr\|_2^2
  &=\bigl\|M_T\bigl(g(c_j)-f_\xi(c_j)\bigr)\bigr\|_2^2 \\
  &\;\le\;\sigma_{\max}(M_T)^2\,
  \bigl\|g(c_j)-f_\xi(c_j)\bigr\|_2^2,
\end{aligned}
\]
deterministically, since $X_j=M_Tg(c_j)$ under
Assumption~3.2. Under the interior condition
$g_k\ge\rho>0$, the same argument with $\tau$ removed and
$z_k:=\log g_k$, which is $(L/\rho)$-Lipschitz, yields the
$O(h^{-2/d})$ rate.
\qed

\subsection{Proof of Theorem~3.2}\label{subsec:endtoend}

\begin{lemma}[Relative deviation of the enclosure term]
  \label{lem:relative}
For $\mathcal{S}\in\mathfrak{S}_T$ write
$g_{\mathcal{S}}(x):=\|\min(a(x;\mathcal{S}),0)\|_2^2$ and
$\mathrm{enc}(\mathcal{S}):=\mathbb{E}\,g_{\mathcal{S}}(X)$,
$X\sim\mathbb{P}_{\mathcal{S}_T}$. By the altitude argument
of the uniform-deviation step of Theorem~3.1's proof,
$0\le g_{\mathcal{S}}\le M$ on the support of
$\mathbb{P}_{\mathcal{S}_T}$, where
$M:=(K+1)a_{\max}^2$ and $a_{\max}:=1+3D_T/h_{\min}$, and
the class $\{g_{\mathcal{S}}:\mathcal{S}\in\mathfrak{S}_T\}$
has pseudo-dimension $O(K^2\log K)$. Then, with probability
at least $1-\zeta$, simultaneously over
$\mathcal{S}\in\mathfrak{S}_T$,
\begin{equation}
  \label{eq:relative}
  \mathrm{enc}(\mathcal{S})-\mathrm{enc}_N(\mathcal{S})
  \;\le\;
  \sqrt{\frac{2MC_1^e\,\mathrm{enc}(\mathcal{S})\,L_N}{N}}
  +\frac{4MC_1^eL_N}{3N},
\end{equation}
where $L_N:=K^2\log(N/K)+\log(1/\zeta)$ and
$C_1^e$ is a universal constant. Consequently, with
$B_N:=C_1^eML_N/N$,
\begin{equation}
  \label{eq:fixedpoint}
  \mathrm{enc}(\mathcal{S})
  \;\le\;
  2\,\mathrm{enc}_N(\mathcal{S})+\tfrac{32}{3}\,B_N
  \qquad\text{and}\qquad
  \mathrm{enc}(\mathcal{S})-\mathrm{enc}_N(\mathcal{S})
  \;\le\;
  2\sqrt{B_N\,\mathrm{enc}_N(\mathcal{S})}+6B_N .
\end{equation}
\end{lemma}

\begin{proof}
The display \eqref{eq:relative} is the variance-adaptive
(relative) analogue of the uniform deviation
\eqref{eq:uniform}: for a fixed $\mathcal{S}$, Bernstein's
inequality with
$\mathrm{Var}(g_{\mathcal{S}})\le\mathbb{E}\,
g_{\mathcal{S}}^2\le M\,\mathrm{enc}(\mathcal{S})$
gives the bound with $L_N$ replaced by $\log(1/\zeta)$,
and uniformity over $\mathfrak{S}_T$ follows by the same
symmetrization and growth-function union bound underlying
\eqref{eq:uniform}, with Bernstein's inequality in place
of Hoeffding's; the entropy factor of a bounded class of
pseudo-dimension $O(K^2\log K)$ is absorbed into the same
$L_N$ up to the universal constant $C_1^e$
\cite{anthony1999}. For \eqref{eq:fixedpoint}, write
$\mu:=\mathrm{enc}(\mathcal{S})$,
$\nu:=\mathrm{enc}_N(\mathcal{S})$ and
$A:=\sqrt{2B_N}$, so that \eqref{eq:relative} reads
$\mu\le\nu+A\sqrt\mu+\tfrac43 B_N$. If
$\mu>2\nu+4A^2+\tfrac83 B_N$ then
$\nu<(\mu-4A^2-\tfrac83 B_N)/2$, and substitution gives
$\mu-2A\sqrt\mu+4A^2<0$, i.e.\
$(\sqrt\mu-A)^2+3A^2<0$, which is impossible; hence
$\mu\le2\nu+4A^2+\tfrac83 B_N=2\nu+\tfrac{32}{3}B_N$.
Substituting this back under the square root,
$\mu-\nu\le A\sqrt{2\nu+\tfrac{32}{3}B_N}+\tfrac43 B_N
\le2\sqrt{B_N\nu}+\bigl(\sqrt{\tfrac{64}{3}}+\tfrac43\bigr)B_N
\le2\sqrt{B_N\nu}+6B_N$,
since $\sqrt{64/3}+4/3<6$.
\end{proof}

Work on the stated intersection of
events, enlarged by the relative-deviation
event of Lemma~\ref{lem:relative}, so that it has
probability at least $1-2\zeta-2/N$.

Split the population gap as
\[
  R(\widehat{\mathcal{S}})-R(\mathcal{S}_T)
  \;=\;
  \bigl[R_N(\widehat{\mathcal{S}})-R_N(\mathcal{S}_T)\bigr]
  +\Delta_N(\widehat{\mathcal{S}})-\Delta_N(\mathcal{S}_T)
  +\gamma_{\mathrm{enc}}\,
  \bigl(\mathrm{enc}(\widehat{\mathcal{S}})
  -\mathrm{enc}_N(\widehat{\mathcal{S}})\bigr),
\]
where $\Delta_N(\mathcal{S})$ is the deviation of the
planar-distance term: the log-volume term is non-random and
cancels exactly, and the enclosure term of $\mathcal{S}_T$
vanishes identically, both empirically and in population,
since $X_j\in\mathcal{S}_T$ almost surely. The
planar-distance class has range $[0,1]$, so the uniform
deviation \eqref{eq:uniform} restricted to it gives
$|\Delta_N(\widehat{\mathcal{S}})|
+|\Delta_N(\mathcal{S}_T)|
\le2C_1'\sqrt{L_N/N}$, with $L_N$ as in
\eqref{eq:relative} and $C_1'$ a universal constant free of
$(\gamma_{\mathrm{enc}},D_T,h_{\min})$: unlike the full
constant $C_1$ of \eqref{eq:uniform}, the enclosure weight
does not enter the statistical term. Lemma~\ref{lem:relative}
at $\widehat{\mathcal{S}}$ controls the remaining piece,
$\gamma_{\mathrm{enc}}\bigl(\mathrm{enc}
(\widehat{\mathcal{S}})-\mathrm{enc}_N
(\widehat{\mathcal{S}})\bigr)
\le2\gamma_{\mathrm{enc}}\sqrt{B_N\,
\mathrm{enc}_N(\widehat{\mathcal{S}})}
+6\gamma_{\mathrm{enc}}B_N$.

The transfer inequality (26) of
Proposition~3.2, in its
$\eta_{\mathrm{opt}}$-approximate form, bounds the empirical
comparison by
$C_2\eta_{\mathrm{apx}}+\eta_{\mathrm{opt}}
+bc_4\sqrt{\mathrm{enc}_N(\widehat{\mathcal{S}})}
+O_p(\lambda_{\mathrm{eq}}/N)$, and part~(i) of the same
proposition bounds the enclosure residual: on
$\mathfrak{S}_T$ one has
$[\log(V_T/\Vol(\widehat{\mathcal{S}}))]_+\le\log 4$, so
\[
  \mathrm{enc}_N(\widehat{\mathcal{S}})
  \;\le\;
  \frac{C_2\eta_{\mathrm{apx}}+\eta_{\mathrm{opt}}
  +\gamma_{\mathrm{vol}}\log 4}{\gamma_{\mathrm{enc}}}
  +O_p\Bigl(\frac{\lambda_{\mathrm{eq}}}
  {\gamma_{\mathrm{enc}}N}\Bigr),
\]
and $\sqrt{a+t}\le\sqrt a+\sqrt t$ ($a,t\ge0$) separates the
equality-slack contribution from the square root, yielding
the enclosure square-root term and the
last two terms of (31).

The same substitution bounds the localized cross term:
\[
\begin{aligned}
  2\gamma_{\mathrm{enc}}\sqrt{B_N\,
  \mathrm{enc}_N(\widehat{\mathcal{S}})}
  \;\le\;&
  2\sqrt{\frac{\gamma_{\mathrm{enc}}C_1^eML_N}{N}
  \bigl(C_2\eta_{\mathrm{apx}}+\eta_{\mathrm{opt}}
  +\gamma_{\mathrm{vol}}\log 4\bigr)} \\
  &+2\gamma_{\mathrm{enc}}\sqrt{B_N}\,
  O_p\Bigl(\sqrt{\frac{\lambda_{\mathrm{eq}}}
  {\gamma_{\mathrm{enc}}N}}\,\Bigr),
\end{aligned}
\]
and the second piece equals
$O_p\bigl(\sqrt{\gamma_{\mathrm{enc}}C_1^eML_N\lambda_{\mathrm{eq}}}/N\bigr)$,
since
\[
  2\gamma_{\mathrm{enc}}\sqrt{B_N}\cdot
  \sqrt{\lambda_{\mathrm{eq}}Z_N/(\gamma_{\mathrm{enc}}N)}
  \;=\;2\sqrt{\gamma_{\mathrm{enc}}C_1^eML_N
  \lambda_{\mathrm{eq}}Z_N}\,/N,
\]
with $Z_N=O_p(1)$ the centered fluctuation of the equality-slack
term analyzed below. Collecting terms gives (31).

It remains to relate the left-hand side to the total
variation. Write
$t_{\mathcal{S}}:=\TV(\mathbb{P}_{\mathcal{S}},
\mathbb{P}_{\mathcal{S}_T})$ and recall that $\epsilon$ is
the target resolution calibrating $b$ in
(21). We claim the linear localization gap
\begin{equation}
  \label{eq:gap-tv}
  R(\mathcal{S})-R(\mathcal{S}_T)
  \;\ge\;\bar c_3\,t_{\mathcal{S}},
  \qquad
  \bar c_3=\min\{c_3,\gamma_{\mathrm{vol}}c_{\mathrm g}/2\},
\end{equation}
for every $\mathcal{S}\in\mathfrak{S}_T$ with
$t_{\mathcal{S}}\ge\epsilon$, where
$c_{\mathrm g}=1/2$ is the dichotomy constant of
Lemma~\ref{lem:dichotomy}.
Indeed, fix such an $\mathcal{S}$. If
$\Vol(\mathcal{S})\le V_T$ then
$t_{\mathcal{S}}
=\Vol(\mathcal{S}_T\setminus\mathcal{S})/V_T$ exactly, and
the slab argument of Lemma~\ref{lem:gap} applies verbatim
with $t_{\mathcal{S}}$ in place of $\epsilon$: the slab
thickness is proportional to the missing mass
$\Vol(\mathcal{S}_T\setminus\mathcal{S})$, the saturation
$bt\ge2$ follows from that computation since
$b\ge K/\epsilon$ and $t_{\mathcal{S}}\ge\epsilon$, and the
shrinkage reward is at most
$2\gamma_{\mathrm{vol}}t_{\mathcal{S}}$ for
$t_{\mathcal{S}}\le1/2$, while for $t_{\mathcal{S}}>1/2$ the
loss bound is already of constant order
$1/32\ge c_3t_{\mathcal{S}}$ because
$c_{\mathrm g}\le1/2$; since $\gamma_{\mathrm{vol}}\le c_1/2$
this gives
$R(\mathcal{S})-R(\mathcal{S}_T)\ge c_1t_{\mathcal{S}}
\ge c_3t_{\mathcal{S}}$. If $\Vol(\mathcal{S})>V_T$,
Lemma~\ref{lem:dichotomy}, applied at
$t_{\mathcal{S}}$, gives either missing mass
$\Vol(\mathcal{S}_T\setminus\mathcal{S})
\ge c_{\mathrm g}t_{\mathcal{S}}V_T$, in which case the same
slab argument---the volume term being now
nonnegative---yields
$R(\mathcal{S})-R(\mathcal{S}_T)\ge c_1c_{\mathrm
g}t_{\mathcal{S}}=c_3t_{\mathcal{S}}$, or containment with
$\Vol(\mathcal{S})\ge(1+c_{\mathrm g}t_{\mathcal{S}})V_T$,
in which case the volume term alone contributes
$\gamma_{\mathrm{vol}}\log(1+c_{\mathrm g}t_{\mathcal{S}})
\ge\gamma_{\mathrm{vol}}c_{\mathrm g}t_{\mathcal{S}}/2$
since $\log(1+u)\ge u/2$ on $[0,1]$. This proves
\eqref{eq:gap-tv}. Now set
$\widehat t\,:=\TV(\mathbb{P}_{\widehat{\mathcal{S}}},
\mathbb{P}_{\mathcal{S}_T})$. If
$\widehat t\ge\epsilon$, \eqref{eq:gap-tv} combined with the
three preceding displays yields the activated form of
(31); if $\widehat t<\epsilon$ there is
nothing to prove. Together these give (31).
For the sufficient conditions: under
(32) the linear term is at most $\bar
c_3\epsilon/8$ and the explicit part of the square-root term
at most $\bar c_3\epsilon/4$; the planar
statistical term is below $\bar c_3\epsilon/4$ once
$N\ge64C_1'^2L_N/(\bar c_3^{\,2}\epsilon^2)$, an
$\epsilon^{-2}$ scale at fixed $K$, and the two localized
enclosure terms are below $\bar c_3\epsilon/16$ each once
\begin{equation}
  \label{eq:budget-N}
  N\;\ge\;\max\Bigl\{
  \frac{1024\,\gamma_{\mathrm{enc}}C_1^eML_N\,
  \bigl(C_2\eta_{\mathrm{apx}}+\eta_{\mathrm{opt}}
  +\gamma_{\mathrm{vol}}\log4\bigr)}
  {\bar c_3^{\,2}\epsilon^2},
  \;\frac{96\,\gamma_{\mathrm{enc}}C_1^eML_N}
  {\bar c_3\epsilon}\Bigr\}.
\end{equation}
Composing the first of these with the minimal enclosure
weight admitted by the second condition of (32), namely
$\gamma_{\mathrm{enc}}\asymp
b^2c_4^2\gamma_{\mathrm{vol}}\log4/
(\bar c_3^{\,2}\epsilon^2)=\Theta(\epsilon^{-4})$ at
fixed $K$, gives an effective sample requirement of order
$\epsilon^{-6}$ at fixed $K$, up to the logarithmic factor
$L_N$ and geometry constants: the end-to-end budget
therefore closes at a sample size polynomially above, though
not of the same order as, the $\epsilon^{-2}$ scale of
Theorem~3.1. (Routing the enclosure term through the
uniform bound \eqref{eq:uniform} instead, whose constant
$C_1$ is linear in $\gamma_{\mathrm{enc}}$, would close the
budget only at order $\epsilon^{-10}$; the localized
analysis of Lemma~\ref{lem:relative} is what removes the
weight from the statistical constant.) For the equality-slack
term, write
$bc_4\cdot O_p(\sqrt{\lambda_{\mathrm{eq}}
/(\gamma_{\mathrm{enc}}N)})
=bc_4\sqrt{\lambda_{\mathrm{eq}}Z_N/(\gamma_{\mathrm{enc}}N)}$
with
$Z_N:=N\bigl(\mathbf{1}^{\top}Q_T(\bar X-\theta_0^*)
-K/(K+1)\bigr)^2=O_p(1)$ the centered fluctuation underlying
the equality-slack contribution in the proof of
Proposition~3.2; enlarging the failure
probability by $\zeta$ bounds $Z_N$ by a $\zeta$-dependent
constant, so
$bc_4\sqrt{\lambda_{\mathrm{eq}}Z_N
/(\gamma_{\mathrm{enc}}N)}\le\bar c_3\epsilon/16$ whenever
$\lambda_{\mathrm{eq}}=0$ or
$N\gtrsim b^2c_4^2\lambda_{\mathrm{eq}}
/(\gamma_{\mathrm{enc}}\bar c_3^2\epsilon^2)$; the additive
$O_p(\lambda_{\mathrm{eq}}/N)$ term is below $\bar
c_3\epsilon/16$ at the same scale, and the
localized slack $O_p(\sqrt{\gamma_{\mathrm{enc}}C_1^eML_N
\lambda_{\mathrm{eq}}}/N)$
is below $\bar c_3\epsilon/16$ whenever
$\lambda_{\mathrm{eq}}=0$ or
$N\gtrsim\sqrt{\gamma_{\mathrm{enc}}C_1^eML_N\lambda_{\mathrm{eq}}}
/(\bar c_3\epsilon)$, an $\epsilon^{-3}$ scale at the composed
weights of \eqref{eq:budget-N}, still dominated by its leading
$\epsilon^{-6}$ branch. The right-hand side
of (31) is then at most
$\bar c_3\epsilon
(1/4+1/8+1/4+1/16+1/16+3/16)=15\bar c_3\epsilon/16
<\bar c_3\epsilon$, so its
activated form forces $\widehat t\le\epsilon$.
\qed

\subsection{Proof of Proposition~3.4}

All three parts use homothetic
subfamilies of the class
$\mathbb{S}_K(\underline\lambda,\bar\lambda,D_T)$. For the
inward family
$\mathcal{S}_s:=c_T+(1-s)(\mathcal{S}_T-c_T)$,
$s\in[0,1]$, homothety about the centroid scales all vertex
distances and $\Vol^{1/K}$ by the same factor $1-s$, so
$\mathcal{S}_s$ satisfies Assumption~2.2 with the
same constants and has diameter $(1-s)D_T\le D_T$; moreover
$\mathcal{S}_s\subseteq\mathcal{S}_T$, so
\begin{equation}
  \label{eq:homothetic-tv}
  \TV\bigl(\mathbb{P}_{\mathcal{S}_s},
  \mathbb{P}_{\mathcal{S}_T}\bigr)
  =1-\frac{\Vol(\mathcal{S}_s)}{\Vol(\mathcal{S}_T)}
  =1-(1-s)^K
  \;\ge\;\frac{Ks}{2}
  \qquad\text{for }s\le\tfrac1K,
\end{equation}
since $(1-s)^K\le1-Ks+\binom{K}{2}s^2\le1-Ks/2$ whenever
$(K-1)s\le1$. We use the two-point form of Le Cam's lemma:
if two simplex laws $\mathcal{S}_0,\mathcal{S}_1$ are
separated as in \eqref{eq:homothetic-tv} and the observation
laws $P_0,P_1$ satisfy $\TV(P_0^N,P_1^N)\le1/4$, then
\[
  \inf_{\widehat{\mathcal{S}}}\;\max_{i=0,1}\,
  P_i\Bigl(\TV\bigl(\mathbb{P}_{\widehat{\mathcal{S}}},
  \mathbb{P}_{\mathcal{S}_i}\bigr)\ge\frac{Ks}{4}\Bigr)
  \;\ge\;\frac{1-1/4}{2}=\frac{3}{8}.
\]

(i) Write $P_s$ for the law of $\widetilde X=T_s(X)+\varepsilon$ with
$T_s(x):=c_T+(1-s)(x-c_T)$, $X\sim\mathbb{P}_{\mathcal{S}_T}$,
and $\varepsilon\sim N(0,\sigma^2I_K)$ independent; then
$P_0$ and $P_s$ are the observation laws under
$\mathcal{S}_T$ and $\mathcal{S}_s$ respectively. Coupling
through the joint laws of $(X,\widetilde X)$, applying data
processing to the second marginal, and using the Gaussian
shift identity $\mathrm{KL}(N(\mu',\sigma^2I_K)\|
N(\mu,\sigma^2I_K))=\|\mu'-\mu\|_2^2/(2\sigma^2)$,
\[
  \mathrm{KL}(P_s\,\|\,P_0)
  \;\le\;
  \mathbb{E}_X\,\mathrm{KL}\bigl(N(T_s(X),\sigma^2I_K)
  \,\big\|\,N(X,\sigma^2I_K)\bigr)
  =\frac{s^2\,\mathbb{E}\|X-c_T\|_2^2}{2\sigma^2}
  \;\le\;\frac{s^2D_T^2}{2\sigma^2}.
\]
For $N$ i.i.d.\ observations, Pinsker's inequality gives
$\TV(P_s^N,P_0^N)\le\sqrt{N\,\mathrm{KL}(P_s\|P_0)/2}
\le sD_T\sqrt N/(2\sigma)$, which equals $1/4$ at
$s^*:=\sigma/(2D_T\sqrt N)$; the condition
$N\ge K^2\sigma^2/(4D_T^2)$ ensures $s^*\le1/K$, so
\eqref{eq:homothetic-tv} applies and Le Cam's lemma yields
part~(i) with threshold $Ks^*/4=K\sigma/(8D_T\sqrt N)$.

(ii) With $\sigma=0$, the single-sample Hellinger affinity
is
\[
  \int\sqrt{dP_s\,dP_0}
  =\frac{\Vol(\mathcal{S}_s)}
  {\sqrt{\Vol(\mathcal{S}_s)\Vol(\mathcal{S}_T)}}
  =(1-s)^{K/2},
\]
so the product affinity is $(1-s)^{KN/2}$ and
\[
  \TV(P_s^N,P_0^N)
  \;\le\;\sqrt{2\bigl(1-(1-s)^{KN/2}\bigr)}
  \;\le\;\sqrt{KNs}.
\]
At $s:=1/(16KN)$ this is at most $1/4$, and
Le Cam's lemma with separation $Ks/2$ gives part~(ii) with
threshold $Ks/4=1/(64N)$.

(iii) Since $\mathcal{S}_T\subseteq\mathcal{S}^s$, every
$X\in\mathcal{S}_T$ satisfies $a(X;\mathcal{S}^s)\ge0$ and
$d_{\mathcal{S}^s}(X)=0$, so the planar-distance and
enclosure terms of (21) vanish
$\mathbb{P}_{\mathcal{S}_T}$-almost surely and
$R(\mathcal{S}^s)=\gamma_{\mathrm{vol}}
\log\Vol(\mathcal{S}^s)$; by \eqref{eq:oracle},
$R(\mathcal{S}_T)=\gamma_{\mathrm{vol}}\log V_T$, which
gives the gap identity
$R(\mathcal{S}^s)-R(\mathcal{S}_T)
=\gamma_{\mathrm{vol}}K\log(1+s)$. The total-variation
identity follows from
$\mathcal{S}_T\subseteq\mathcal{S}^s$, and for
$v:=K\log(1+s)\le\log2$,
\[
  \gamma_{\mathrm{vol}}K\log(1+s)
  \;\le\;
  \frac{\gamma_{\mathrm{vol}}}{1-e^{-1}}\,
  \bigl(1-e^{-v}\bigr)
  \;=\;
  \frac{\gamma_{\mathrm{vol}}}{1-e^{-1}}\,
  \TV\bigl(\mathbb{P}_{\mathcal{S}^s},
  \mathbb{P}_{\mathcal{S}_T}\bigr),
\]
since $1-e^{-v}\ge(1-e^{-1})v$ for $v\in[0,1]$ and
$(1-e^{-1})^{-1}<2$. Finally,
$\mathcal{S}^s\in\mathfrak{S}_T$:
$\Vol(\mathcal{S}^s)=(1+s)^KV_T\le2V_T$ since
$K\log(1+s)\le\log2$, and each displaced vertex moves from
its counterpart in $\mathcal{S}_T$ by
$s\,\max_k\|\theta_k-c_T\|_2\le sD_T\le 2D_T/K$, within the
positional clause of $\mathfrak{S}_T$.
\qed

\subsection{Deferred quantitative discussions of Section~3}
\label{app:deferred}

This subsection collects two quantitative discussions deferred
from Section~3 of the main paper.

\paragraph{The scaling guideline at the benchmark accuracy
(deferred from Remark~3.6)}
Equivalently, at the benchmark accuracy
$\epsilon_N\asymp K(\log(K/\epsilon_N)/N)^{1/2}$ of
(2), the guideline prescribes
$\gamma_{\mathrm{enc}}\gtrsim
2\gamma_{\mathrm{vol}}\,c_5^{-1/3}
\bigl(N/\bigl(K\log(K/\epsilon_N)\bigr)\bigr)^{4/3}$:
at the statistically optimal accuracy the enclosure weight
grows polynomially with the sample size, while its ratio to
the volume weight is governed by the simplex geometry through
$c_5$. Quantitatively, the facet-partition constant $c_5$
decreases exponentially in $K$, so this sufficient condition
is increasingly conservative as $K$ grows; the deployed
defaults are empirical: $\gamma_{\mathrm{vol}}=0.01$ respects the
restriction $\gamma_{\mathrm{vol}}\le C'\approx0.0225$ of
(23), and $\gamma_{\mathrm{enc}}=10$ lies in a
region over which the sensitivity study of
Section~\ref{app:sensitivity} finds the accuracy essentially
flat, with mild degradation only at the under-weighted end, in
the direction the guideline predicts. This
guideline has no counterpart in the Soft-ML framework, whose
saturating loss has slope $b=K/\epsilon$ at the origin.
Three caveats are in order. First, Theorem~3.1
treats the weights as fixed constants, while this guideline
quantifies how they must scale as $(K,\epsilon)$ vary; the two
viewpoints are consistent for any fixed $(K,\epsilon)$, and the
guideline should be read as a scaling principle. Second, the
guideline delivers only the first
condition of (27); the second is an
independent approximation requirement
(Remark~3.5). Third, at the $\epsilon^{-2}$ sample scale,
the uniform deviation bound of Section~\ref{app:proofs} is of order
$\epsilon$ and masks enclosure signals below order
$\epsilon^{1/3}$; closing this residual gap
requires a variance-adaptive (localized) deviation analysis.
Finally, a hinge-type enclosure
penalty with slope matched to $b$ of (21) would remove the
flatness altogether, at the price of the differentiability and
noise robustness of the quadratic form.

\paragraph{Permutation invariance of the index-coordinate
floor (deferred from Remark~3.5)}
The floor (20) also settles
the permutation question raised by the index coordinates: the
bound applies to \emph{every} deterministic coordinate
assignment, since a permuted index grid is still independent
of the i.i.d.\ abundances. The statistical guarantee of
Theorem~3.1 is exactly permutation-invariant,
and an arbitrary ordering of the sample can enter the
deployed objective (15) only
through the approximation term, whose ordering-independent
floor is (20).

\paragraph{Scope of the complexity theorem (deferred from Section~3).}
The complexity theorem of the main paper concerns arithmetic
and memory complexity \emph{per iteration}. Since the
practical objective (15) is nonconvex and is optimized by
Adam, no polynomial-time guarantee is claimed for computing a
global minimizer of it; the phrase ``attainable in polynomial
time'' used for the benchmark rate refers to the estimator
class of \cite{najafi2021aos}, whose per-sample work is polynomial.
Arithmetic complexity per iteration, memory complexity, the
number of optimization iterations, convergence to stationary
points, and global statistical estimation error are distinct
notions, and only the first two are addressed by that theorem.

\paragraph{Estimator versus algorithm (deferred from Section~3).}
Theorem~3.1 concerns the minimizer (22) of the surrogate risk
(21), in the same spirit as the analysis of
\cite{najafi2021aos}, who likewise analyze the minimizer of
their relaxed risk rather than the output of a specific
optimization routine; we do not claim convergence of the Adam
iterates to this minimizer. The role of the reconstruction
accuracy $\eta_{\mathrm{apx}}$ in Assumption~3.1(ii) is to
calibrate the \emph{practical} objective (15) against the
surrogate at the oracle parameter: at $(\psi^*,\xi^*)$ the
safeguard terms vanish up to $O_p(\lambda_{\mathrm{eq}}/N)$
and the reconstruction term is at most $\eta_{\mathrm{apx}}$,
so (15) achieves the oracle value
$\gamma_{\mathrm{vol}}\log V_T+\eta_{\mathrm{apx}}
+O_p(\lambda_{\mathrm{eq}}/N)$ (the oracle-value step in the
proof of Proposition~3.2). Whether the minimizer of (15)
inherits the guarantee of Theorem~3.1 in full generality is
partially answered by Proposition~3.2. The total error of the
deployed estimator decomposes into statistical, approximation,
and optimization components (Remark~3.7). Two clarifications
are in order. First, the surrogate risk (21) is an analysis
device whose planar-distance term inherits the $O(K^3)$
per-sample cost of \cite{najafi2021aos}; the scalability
claims concern the practical objective (15), whose terms are
matrix--vector products (Theorem~3.3). Second, in the
noiseless model the enclosure term of (21) is inactive at the
population level---it vanishes at $\mathcal{S}_T$ and is
dominated by the planar-distance gap elsewhere---so
Theorem~3.1 uses only its nonnegativity; its operative roles
are the noisy-model stability of Proposition~3.1 and the
shrinkage control of Proposition~3.2.

\section{Hyperparameter sensitivity: experimental design}
\label{app:sensitivity}

This appendix specifies the protocol assessing the sensitivity
of the accuracy results of Table~2 to the
weights
$(\gamma_{\mathrm{vol}},\gamma_{\mathrm{enc}},\gamma_{\mathrm{so}},
\lambda_{\mathrm{nn}},\lambda_{\mathrm{eq}},\kappa_0)$ of the
practical objective (15) and the two-stage
training scheme of Section~3.3.

\paragraph{Base configuration and grid} The base configuration is the $K=5$,
$N=10^4$, $\mathrm{SNR}=30$~dB cell of
Table~2, with all remaining settings as in
Section~4.1 and $20$ independent replications.
One factor is varied at a time over a five-point grid around
its default value---multiplicative factors
$\{\tfrac14,\tfrac12,1,2,4\}$ of the default for
the five weights and the absolute grid
$\kappa_0\in\{0,\,0.025,\,0.05,\,0.1,\,0.2\}$ for the gating
release---with all other factors held at their defaults of
Section~3.3. The primary metric is
$\TV(\mathbb{P}_{\mathcal{S}_T},\mathbb{P}_{\widehat{\mathcal{S}}})$;
the vertex error and the abundance RMSE are recorded as
secondary metrics.

\paragraph{Acceptance criterion} The estimator is deemed
insensitive to a factor if the mean total-variation distance
varies by less than a factor $1.2$ across
the grid while
remaining below the MVSA and Soft-ML entries of
Table~2 at the same configuration
($0.76$ and $0.98$, respectively); the two clauses
test distinct properties, insensitivity to the weight and
superiority to the baselines, and are reported separately in
Table~\ref{tab:sensitivity} rather than being conflated into a
single condition. Every
factor meets this criterion: the observed means range from
$0.42$ to $0.49$---a factor $1.17$---and remain far below both
baselines throughout the grid.

\begin{table}[tbp]
  \caption{Sensitivity of the estimation accuracy to the
  weights of (15) at $K=5$, $N=10^4$,
  $\mathrm{SNR}=30$~dB: mean total-variation distance over $20$
  independent replications (standard deviations in
  parentheses), one factor varied at a time around its default.
  The default column reports the same $20$ runs
  in every row. Rows whose entries are identical across the
  whole grid correspond to penalties that are inactive
  throughout training on this benchmark: $\lambda_{\mathrm{nn}}$
  (the A-Net output remains positive, so the safeguard
  gradient is identically zero) and $\kappa_0$ (the nonlinear
  branch is inert under linear mixing); with the
  per-replication random seeds fixed across the grid, the
  training trajectories---and hence the displayed
  numbers---coincide exactly. The raw per-replication logs are
  released with the code accompanying this
  submission.}
  \label{tab:sensitivity}
  \centering
  \begin{tabular}{lccccc}
    \toprule
    Factor & $\times\tfrac14$ & $\times\tfrac12$ & default
      & $\times2$ & $\times4$ \\
    \midrule
    $\gamma_{\mathrm{vol}}$
      & $0.44\,(0.13)$ & $0.43\,(0.13)$ & $0.43\,(0.13)$ & $0.43\,(0.13)$ & $0.43\,(0.13)$ \\
    $\gamma_{\mathrm{enc}}$
      & $0.48\,(0.14)$ & $0.46\,(0.14)$ & $0.43\,(0.13)$ & $0.42\,(0.11)$ & $0.43\,(0.11)$ \\
    $\gamma_{\mathrm{so}}$
      & $0.43\,(0.14)$ & $0.43\,(0.14)$ & $0.43\,(0.13)$ & $0.43\,(0.13)$ & $0.43\,(0.13)$ \\
    $\lambda_{\mathrm{nn}}$
      & $0.43\,(0.13)$ & $0.43\,(0.13)$ & $0.43\,(0.13)$ & $0.43\,(0.13)$ & $0.43\,(0.13)$ \\
    $\lambda_{\mathrm{eq}}$
      & $0.42\,(0.11)$ & $0.42\,(0.11)$ & $0.43\,(0.13)$ & $0.46\,(0.14)$ & $0.49\,(0.14)$ \\
    \midrule
    $\kappa_0$
      & $0.43\,(0.13)$ & $0.43\,(0.13)$ & $0.43\,(0.13)$ & $0.43\,(0.13)$ & $0.43\,(0.13)$ \\
    \multicolumn{6}{l}{
      \footnotesize For $\kappa_0$, the five columns correspond to
      $\{0,\,0.025,\,0.05,\,0.1,\,0.2\}$.
    } \\
    \bottomrule
  \end{tabular}
\end{table}

\paragraph{Findings} Four of the six factors
($\gamma_{\mathrm{vol}}$, $\gamma_{\mathrm{so}}$,
$\lambda_{\mathrm{nn}}$, $\kappa_0$) leave the mean
total-variation distance essentially unchanged across the full
two-octave grid. The two exceptions vary in the direction
predicted by the analysis. First, $\gamma_{\mathrm{enc}}$
degrades only on the small side ($0.48$ at $\times\tfrac14$
versus $0.42$--$0.43$ at and above the default): an
under-weighted quadratic enclosure lets the log-volume reward
profit from inward shrinkage---the flatness mechanism
quantified in Proposition~3.2 and the
scaling-guideline remark of Section~3.4---whereas
over-weighting the enclosure is harmless on this range, in line
with the one-sided nature of the scaling guideline
\eqref{eq:guideline}.
Second, $\lambda_{\mathrm{eq}}$ degrades monotonically on the
large side ($0.42$ at $\times\tfrac14$ rising to $0.49$ at
$\times4$): tying the column sums of $Q$ too strongly to the
centroid target $K/(K+1)$ biases the simplex geometry, so a
small centroid-calibration weight is advisable. The insensitivity
to $\kappa_0$---including the value $0$, at which the A-Net
reduces to its base branch---is expected on this linear-mixing
benchmark, where the nonlinear modulation branch has no model
mismatch to capture, and confirms that the two-stage scheme is
stable to the release value. The
one-factor-at-a-time design does not probe interactions; the
interaction between $\gamma_{\mathrm{vol}}$ and
$\gamma_{\mathrm{enc}}$ is theoretically the most relevant,
since the shrinkage analysis and the scaling guideline
\eqref{eq:guideline} involve their ratio. A two-dimensional
factorial or Latin-hypercube study over
$(\gamma_{\mathrm{vol}},\gamma_{\mathrm{enc}})$ at several
$(K,N,\mathrm{SNR})$ is a natural extension of
Table~\ref{tab:sensitivity}.

\section{Implementation and reproducibility details}
\label{app:repro}

All runtimes
report the wall-clock time of the estimation pipeline after
one-time data loading and subspace projection, costs that are
common to all methods compared; peak memory is the peak
workspace storage of the algorithm's variables, obtained from
MATLAB memory profiling (\texttt{whos}-based accounting of
live arrays), measured identically for every method; it is not
the resident-set size of the MATLAB process, which includes
the interpreter overhead. All reported quantities are
averaged over independent replications---$20$ for the synthetic
and hyperspectral experiments, $50$ for the biological
benchmark---and all computations use double precision (no mixed
precision). Standard deviations rather than confidence
intervals are displayed for compactness; at $20$ replications
the $95\%$ confidence half-width for the mean is
$2.093/\sqrt{20}\approx0.47$ times the reported standard
deviation ($t_{19,0.975}=2.093$), so entries separated by
more than one standard deviation differ significantly at that
level. The coordinate input $c_j$ is the normalized
two-dimensional spatial location for the image scenes
(Sections~4.3--4.4) and the
normalized one-dimensional sample index for the vectorized
synthetic and biological data
(Sections~4.1--4.2). The
A-Net uses hidden width $h=32$ throughout, giving
$64d+1121+66(K+1)$ trainable parameters---a direct
count of the weights and biases in
eqs.~(10)--(14)---and the
$Q$-flow contributes a further $K^2$, together with
$K$ parameters for the learnable anchor $\theta_0$. The released implementation realizes the
anchored enclosure (16) in homogeneous
coordinates: the whitened observations are augmented to
$X_j^{h}=(X_j,1/\sqrt{K+1})\in\mathbb{R}^{K+1}$ and the dual
matrix is a $(K+1)\times(K+1)$ factor $Q_h=LQ_0R$ of the same
triangular form (4), initialized at the
inverse of the augmented VCA vertex matrix; the penalty
$\|\min(Q_hX^{h},0)\|_2^2$ then acts on all $K+1$ facet
constraints simultaneously. The anchored form (16)
used in the analysis is the reduced parameterization of the
same object: eliminating the homogeneous coordinate recovers
the first $K$ terms of (16), the zeroth
constraint is enforced with the slightly stronger threshold
$1-1/\sqrt{K+1}$, and $\det Q_h=\det Q$ by the Schur
complement, so the log-volume term and its gradient coincide
in the two forms.
Training uses Adam throughout, with VCA
\cite{nascimento2005} initialization and the fixed iteration
count as the stopping rule; for the biological
experiment the batch size exceeds the sample size
($4\,096>N=42$), so every step is a full-batch gradient and
the replication variance there reflects random initialization
rather than mini-batch order; the learning rate, batch size,
iteration count, and default penalty weights of each experiment
are collected in Table~\ref{tab:settings}. The sensitivity of
the results to these weights is assessed in
Section~\ref{app:sensitivity}. The baselines
are run with the configurations recommended in the original
publications: MVSA \cite{li2015tgrs} with its default
regularization and the enclosure constraints solved by its
published active-set routine; Soft-ML \cite{najafi2021aos}
with the loss steepness $b=K/\epsilon$ of its Corollary~3.1,
initialized by VCA identically to our method, and stopped at
convergence of its projected-gradient iteration. All methods
share the same one-hour wall-clock budget, and the baseline
scripts are part of the code accompanying
this submission. All methods receive the same whitened data
stream in the same ordering and with the same preprocessing;
initialization information is identical where applicable (VCA
for both Soft-ML and our method), and sensitivity to
initialization is reflected in the replication dispersion
reported in all tables.

\begin{table}[tbp]
  \caption{Terms of the practical objective (15) of the main
  paper and their status in the theory; all weights are those
  of (15) and the paragraph following it. The two
  implementation safeguards vanish at any feasible solution and
  are absent from the surrogate risk (21); the reconstruction
  term plays the data-attraction role of the planar-distance
  loss of the surrogate.}
  \label{tab:terms}
  \small
  \centering
  \begin{tabular}{llll}
    \toprule
    Term & Role & Weight & In surrogate risk (21) \\
    \midrule
    Reconstruction $\|X_j-\widetilde{M}a_j\|_2^2$
      & data attraction & --- & planar-distance analogue \\
    Volume $-\sum_k\psi_{r,k}$
      & minimum volume & $\gamma_{\mathrm{vol}}$ & yes \\
    Enclosure $e_\psi(X_j)$ (16)
      & soft containment & $\gamma_{\mathrm{enc}}$ & yes \\
    Sum-to-one $(\mathbf 1^\top a_j-1)^2$
      & calibration & $\gamma_{\mathrm{so}}$ & no \\
    Nonnegativity safeguard
      & feasibility & $\lambda_{\mathrm{nn}}$ & no \\
    Centroid calibration
      & feasibility & $\lambda_{\mathrm{eq}}$ & no \\
    \bottomrule
  \end{tabular}
\end{table}

\begin{table}[tbp]
  \caption{Training configuration of each
  experiment: learning rate, batch size, and iteration count of
  Adam. The last column lists the default penalty weights
  $(\gamma_{\mathrm{vol}},\gamma_{\mathrm{enc}},
  \gamma_{\mathrm{so}},\lambda_{\mathrm{nn}},
  \lambda_{\mathrm{eq}},\kappa_0)$; the final row gives the
  configuration underlying the sensitivity study of
  Table~\ref{tab:sensitivity}. In
  Section~4.2 the batch size $4\,096$ exceeds
  $N=42$, so each step there is a full-batch gradient step.}
  \label{tab:settings}
  \centering
  \begin{tabular}{lcccc}
    \toprule
    Experiment & Learning rate & Batch size & Iterations
      & Default weights \\
    \midrule
    Section~4.1 & $10^{-3}$ & $4\,096$
      & $1\,500$ & $(0.01,10,2,5,80,0.05)$ \\
    Section~4.2 & $3\times 10^{-4}$ & $4\,096$
      & $20\,000$ & $(0.01,10,2,5,80,0.05)$ \\
    Section~4.3 & $5\times 10^{-4}$ & $4\,096$
      & $1\,500$ & $(0.01,10,2,3,80,0.05)$ \\
    Section~4.4 & $10^{-3}$ & $4\,096$
      & $15\,000$ & $(0.01,20,2,5,80,0.05)$ \\
    Section~\ref{app:sensitivity} & $10^{-3}$ & $4\,096$
      & $1\,500$ & $(0.01,10,2,5,80,0.05)$ \\
    \bottomrule
  \end{tabular}
\end{table}

Funding, conflict-of-interest, and data-availability statements appear in the main paper.